\documentclass[10pt,journal,compsoc]{IEEEtran}

\ifCLASSOPTIONcompsoc
  \usepackage[nocompress]{cite}
\else
  \usepackage{cite}
\fi

\usepackage{diagbox}

\usepackage{tabularx}
\usepackage{multirow} 
\usepackage{multicol}

\usepackage{enumitem}

\usepackage{subfig}
\usepackage{graphicx}
\usepackage{amsmath}
\usepackage{amssymb}

\usepackage{xcolor}
\usepackage{booktabs}

\usepackage{xspace}
\usepackage{xcolor}
\usepackage{enumitem}
\usepackage{enumitem}
\usepackage[pagebackref=true,breaklinks=true,letterpaper=true,colorlinks,bookmarks=false]{hyperref}

\usepackage[normalem]{ulem}
\useunder{\uline}{\ul}{}
\usepackage[font=footnotesize]{caption}%,labelfont=bf

\usepackage{url}  

\usepackage{xspace}
\makeatletter
\DeclareRobustCommand\onedot{\futurelet\@let@token\@onedot}
\def\@onedot{\ifx\@let@token.\else.\null\fi\xspace}
\def\eg{\emph{e.g}\onedot} 
\def\ie{\emph{i.e}\onedot}

\def\etal{\emph{et al}\onedot}
\makeatother

\usepackage{pifont}
\newcommand{\cmark}{\ding{51}}%
\newcommand{\xmark}{\ding{55}}%

\begin{document}
% \title{Bare Demo of IEEEtran.cls for\\ IEEE Computer Society Journals}
\title{Accurate 3-DoF Camera Geo-Localization via Ground-to-Satellite Image Matching}

\author{Yujiao~Shi*,
        Xin~Yu*,
        Liu~Liu,
        Dylan~Campbell,
        Piotr~Koniusz,
        and~Hongdong~Li% <-this % stops a space
% \author{Michael~Shell,~\IEEEmembership{Member,~IEEE,}
%         John~Doe,~\IEEEmembership{Fellow,~OSA,}
%         and~Jane~Doe,~\IEEEmembership{Life~Fellow,~IEEE}% <-this % stops a space
\IEEEcompsocitemizethanks{
\IEEEcompsocthanksitem *Equal contribution. 
\IEEEcompsocthanksitem 
Y. Shi and H. Li are with ANU, Australian National University. \protect\\%
E-mail:{firstname.lastname}@anu.edu.au.
\IEEEcompsocthanksitem X. Yu is with Australian Artificial Intelligence Institute, the University of Technology Sydney.\protect\\
E-mail: xin.yu@uts.edu.au.
\IEEEcompsocthanksitem L. Liu is with the Cyberverse Lab, Huawei.\protect\\
E-mail: NWPUliuliu@gmail.com.
\IEEEcompsocthanksitem D. Campbell is with the University of Oxford.\protect\\
E-mail: dylan@robots.ox.ac.uk.
\IEEEcompsocthanksitem P. Koniusz is with the Machine Learning Research Group (MLRG), Data61/CSIRO (NICTA), Canberra, and ANU.
E-mail: piotr.koniusz@data61.csiro.au.}% <-this % stops an unwanted space
% \thanks{Manuscript received April 19, 2005; revised August 26, 2015.}
}


% \markboth{Journal of \LaTeX\ Class Files,~Vol.~14, No.~8, August~2015}%
% {Shell \MakeLowercase{\textit{et al.}}: Bare Demo of IEEEtran.cls for Computer Society Journals}
\markboth{ }%
{ }




\IEEEtitleabstractindextext{%

\begin{abstract}
We address the problem of ground-to-satellite image geo-localization, that is, estimating the camera latitude, longitude and orientation (azimuth angle) by matching a query image captured at the ground level against a large-scale database with geotagged satellite images.
Our prior arts treat the above task as pure image retrieval by selecting the most similar satellite reference image matching the ground-level query image.
However, such an approach often produces coarse location estimates because the geotag of the retrieved satellite image only corresponds to the image center while the ground camera can be located at any point within the image. 
To further consolidate our prior research finding, we present a novel geometry-aware geo-localization method. Our new method is able to achieve the fine-grained location of a query image, up to pixel size precision of the satellite image, once its coarse location and orientation have been determined.
Moreover, we propose a new geometry-aware image retrieval pipeline to improve the coarse localization accuracy. 
Apart from a polar transform in our conference work, this new pipeline also maps satellite image pixels to the ground-level plane in the ground-view via a geometry-constrained projective transform to emphasize informative regions, such as road structures, for cross-view geo-localization.
Extensive quantitative and qualitative experiments demonstrate the effectiveness of our newly proposed framework.
We also significantly improve the performance of coarse localization results compared to the state-of-the-art in terms of location recalls.
\end{abstract}

}

\maketitle

\IEEEdisplaynontitleabstractindextext

% \fi
\IEEEpeerreviewmaketitle

\IEEEraisesectionheading{\section{Introduction}\label{sec:introduction}}

\begin{figure}
    \centering
    \subfloat[Coarse camera geo-localization]{
    \includegraphics[width=\linewidth]{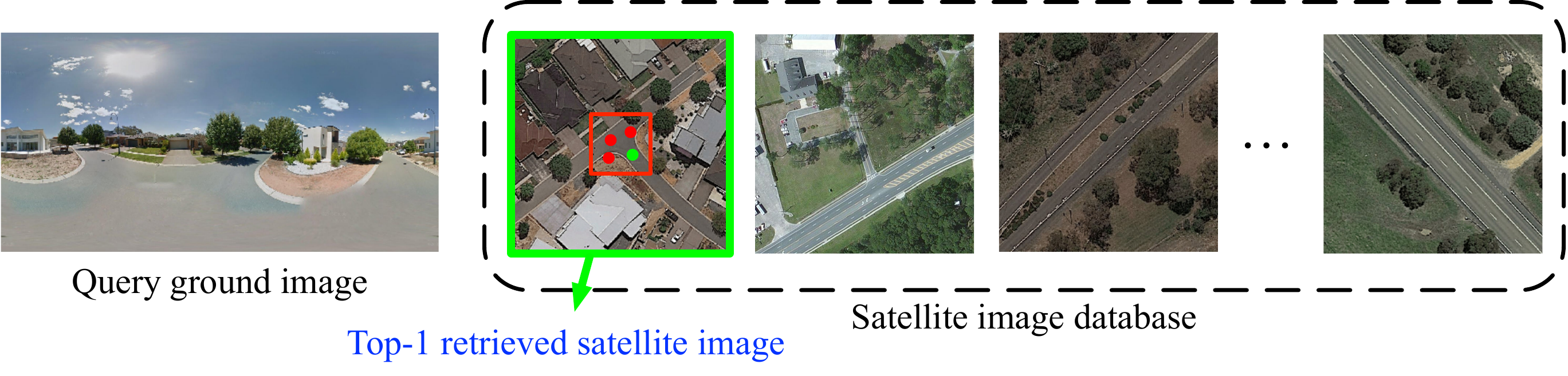}
    \label{fig:intro_coarse_localization}
    }\\
    \subfloat[Fine-grained camera geo-localization]{
    \parbox[][][c]{0.66\linewidth}{
    \includegraphics[width=\linewidth]{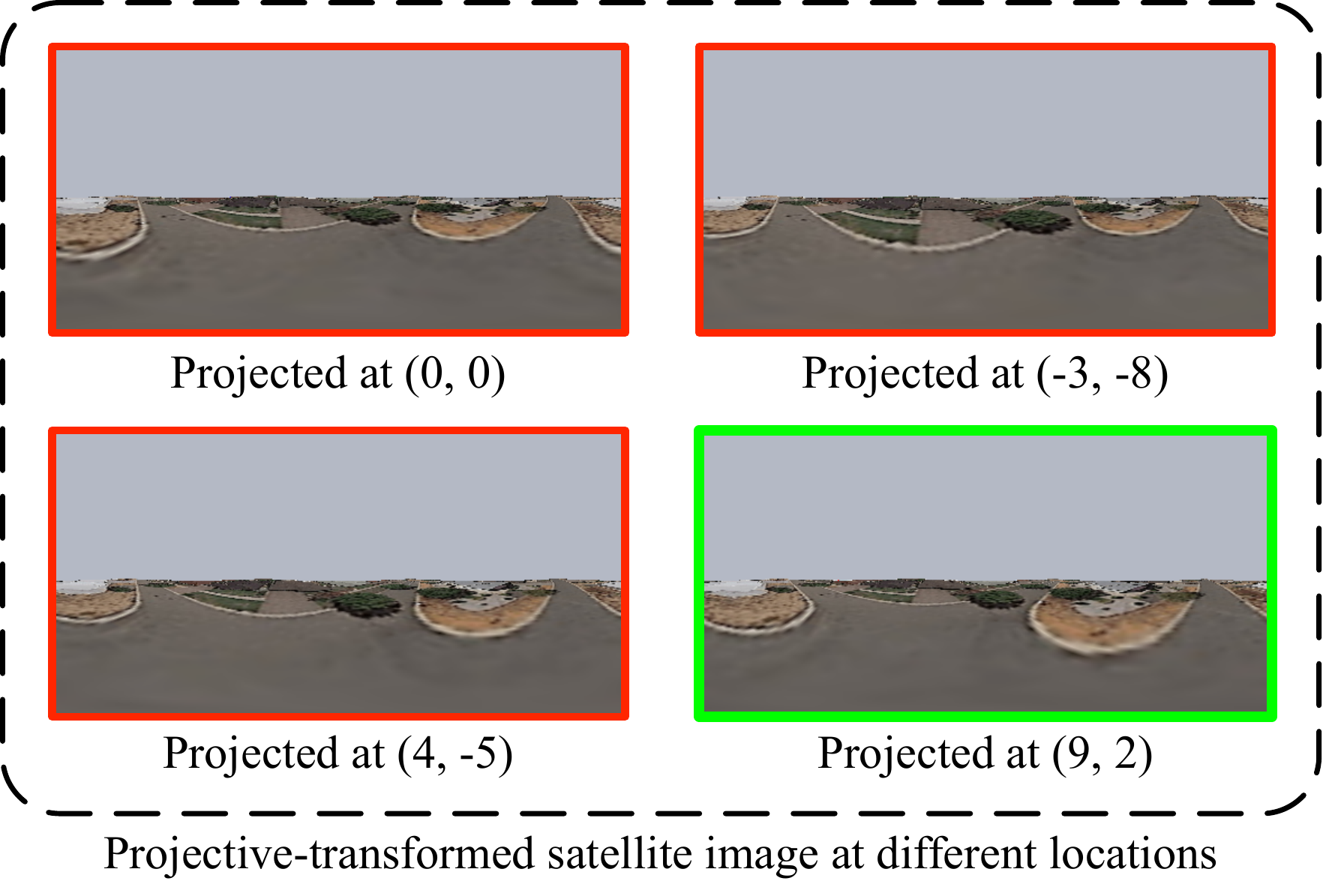}}
    \label{fig: intro_accurate_localization}
    }
    \subfloat[Region selected for fine-grained camera geo-localization]{
    \centering
    \parbox[][][c]{0.3\linewidth}{
    \includegraphics[width=0.9\linewidth]{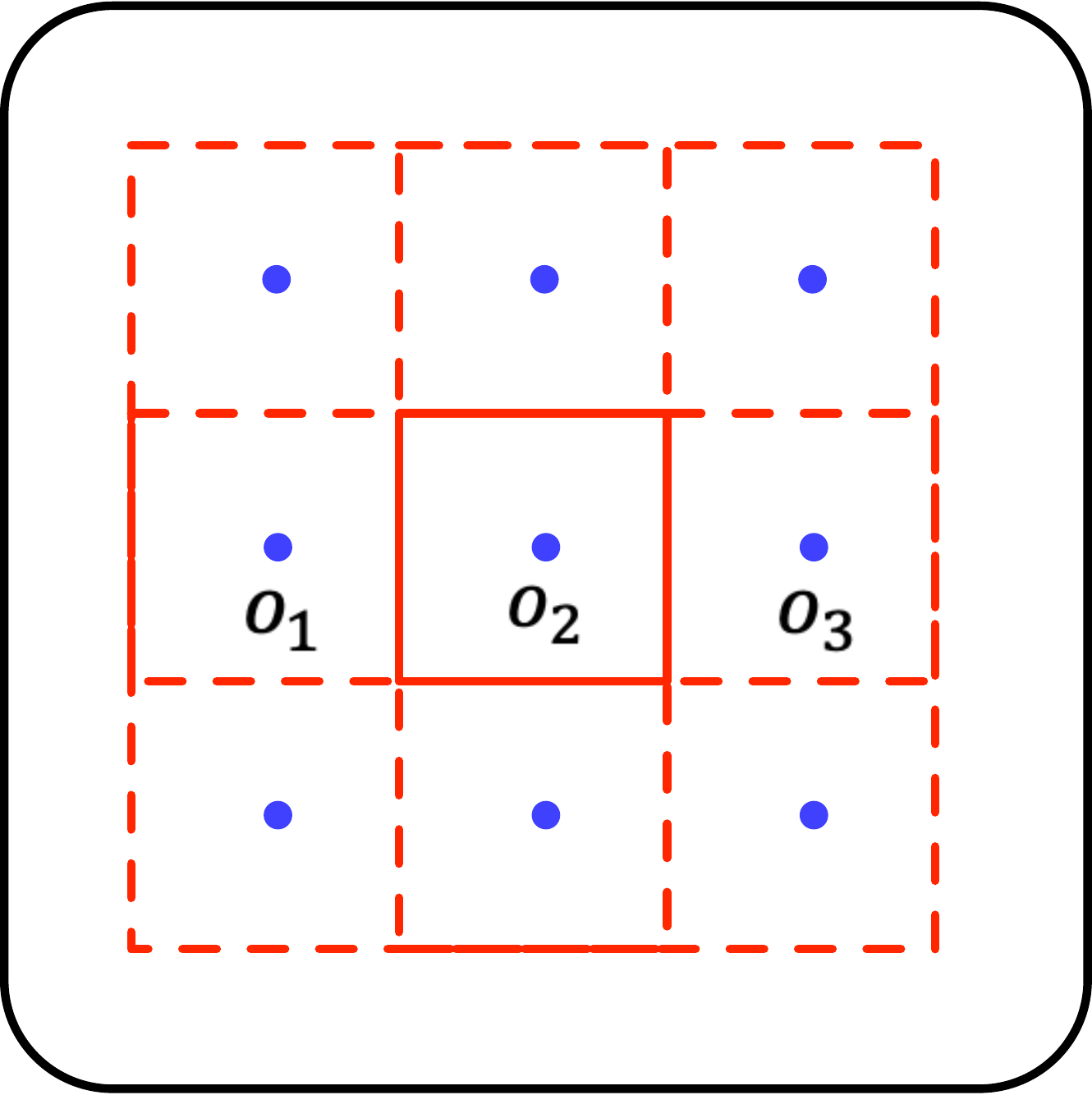}}
    \label{fig:central_region_selection}
    }
    \caption{\footnotesize (a) Given a query ground image, we first retrieve its most similar satellite image from the database. 
    (b) Then, we use a set of candidate locations in the satellite image as projection centers.
    The fine-grained location of the query ground image is then achieved from the projected satellite image that is most similar to the query image.
    The estimated orientation is obtained by comparing the selected projected image and the query image.
    (c) {\color{black}
    The black box represents a large satellite map covering the whole region, from which the small satellite images in the database (shown in (a)) are cropped for coarse camera localization. 
    The blue dots denote the centers of those cropped images.
    The red boxes indicate the regions selected for fine-grained camera localization, which cover nearly the entire satellite map. }}
    \label{fig:open_figure}
\end{figure}

\begin{figure*}[t]
\setlength{\abovecaptionskip}{0pt}
\setlength{\belowcaptionskip}{0pt}
    \centering
    \subfloat[]{
    \includegraphics[width=0.1\linewidth]{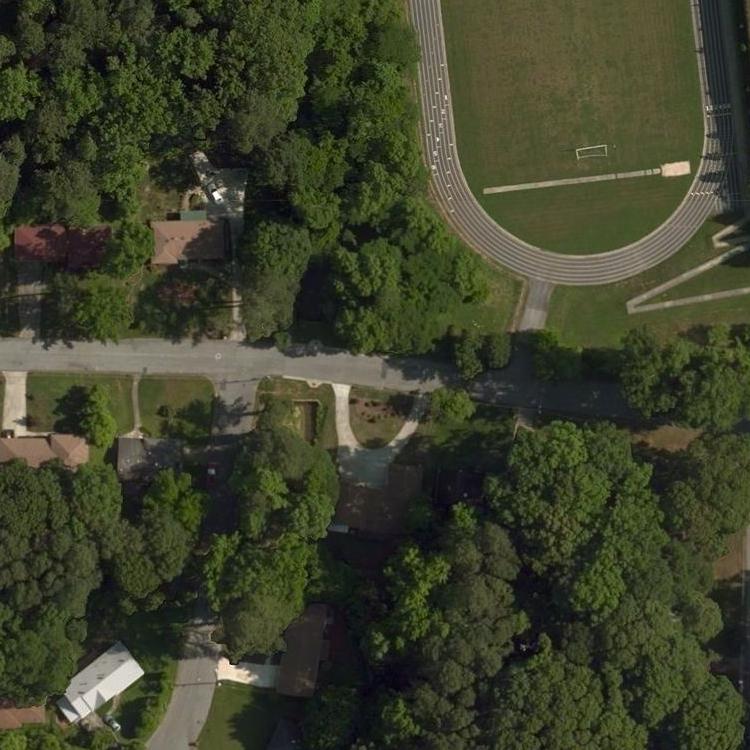}
    \label{fig: intro_satellite}
    }
    \subfloat[]{
    \includegraphics[width=0.25\linewidth, height=0.1\linewidth]{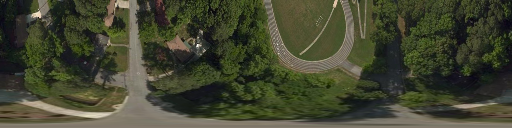}
    \label{fig: intro_Polar-transformed}
    }
    \subfloat[]{
    \includegraphics[width=0.25\linewidth, height=0.1\linewidth]{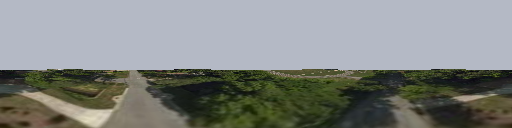}
    \label{fig: intro_Geometry-transformed}
    }
    \subfloat[]{
    \includegraphics[width=0.25\linewidth, height=0.1\linewidth]{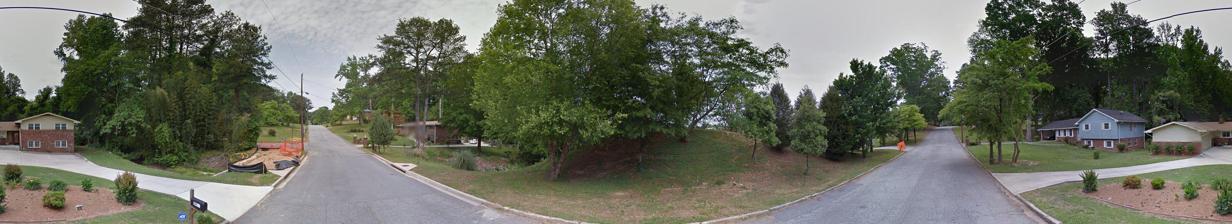}
    \label{fig: intro_Panorama}
    }
    \label{fig: intro_PT_HT}
    \caption{\footnotesize Given a satellite image (a), we explore two transforms, \ie, polar transform (b) and the projective transform (c), to align it to its corresponding ground-view panorama (d).} % up to an unknown azimuth angle
\end{figure*}

\IEEEPARstart{G}{iven} an image captured by a camera at the ground level in some large open space, estimating the camera position and the direction it faces is a useful but also challenging problem. 
%
%it is natural to ask: where is the camera and which direction is it facing?
This paper addresses the problem of ground-to-satellite image geo-localization which aims to determine the geographical location and azimuth angle of a query image by matching it against a large geo-tagged satellite map covering the region.
Due to the accessibility and extensive coverage of satellite imagery, the problem of ground-to-satellite image alignment has been recently noted by researchers as it can feature in a number of computer vision applications \eg, autonomous driving, robot navigation, and way-finding in augmented/virtual reality. %for AR/VR. 

Conventional methods for solving this task often formulate the problem as {\em image retrieval} ({\em e.g.},~\cite{workman2015wide, zhai2017predicting, vo2016localizing, Hu_2018_CVPR, Liu_2019_CVPR, Regmi_2019_ICCV, Cai_2019_ICCV, shi2020optimal, shi2019spatial, shi2020looking}).  They do so by first partitioning a large satellite map to smaller satellite images, to form the reference database.  Once a ground-view query image is provided, they compare the query image with the database images to retrieve the most similar one, as shown in Figure~\ref{fig:intro_coarse_localization}. 
The gps-tag of the matched satellite image is then used to approximate the location of the query image location. 
However, oftentimes the geo-tag of the retrieved satellite image  corresponds to the centre of the retrieved satellite image from the database, while the true ground camera location can be rather off. Therefore, camera locations estimated by these methods are rather coarse and inaccurate.

To address this problem, we introduce a new two-stage mechanism for \emph{accurate 3-DoF (latitude, longitude, and azimuth angle) camera geo-localization} in this paper.
Firstly, we estimate a coarse camera location by searching the most similar satellite image from the database (Figure~\ref{fig:intro_coarse_localization}). Subsequently, we compute the displacement between the center of the retrieved satellite image and the inquired camera location, achieving fine-grained localization results (Figure~\ref{fig: intro_accurate_localization}). 
The orientation alignment (azimuth angle) between the ground and satellite images is estimated in both steps.

At the coarse camera localization stage, we extend the method proposed in our previous work~\cite{shi2020looking}.
Apart from a polar transform (Figure~\ref{fig: intro_Polar-transformed}) that roughly bridges the cross-view domain gap, we develop a projective transform in this work to establish geometrically constrained correspondences (Figure~\ref{fig: intro_Geometry-transformed}) between the satellite (Figure~\ref{fig: intro_satellite}) and ground-level (Figure~\ref{fig: intro_Panorama}) images for scene objects on the ground plane.
Both the polar-transformed and projective-transformed satellite images are used in our coarse localization pipeline. 
The former preserves all details from the original satellite image while the latter exhibits better visual similarity to the captured ground-level scenes. %particularly for close-to-planar scenes. 
Then, we employ CNNs to learn feature correspondences between ground-level and transformed satellite images. 
After satellite images are projected to the corresponding pixels in the ground-level coordinate system, the spatial layout gap between ground and satellite images is significantly reduced. 
Following our prior works~\cite{shi2020looking}, we opt to extract feature volumes as our global descriptors to encode discriminative spatial information. 

\begin{figure*}[t!]
\setlength{\abovecaptionskip}{0pt}
\setlength{\belowcaptionskip}{0pt}
    \centering
    \includegraphics[width=0.85\linewidth]{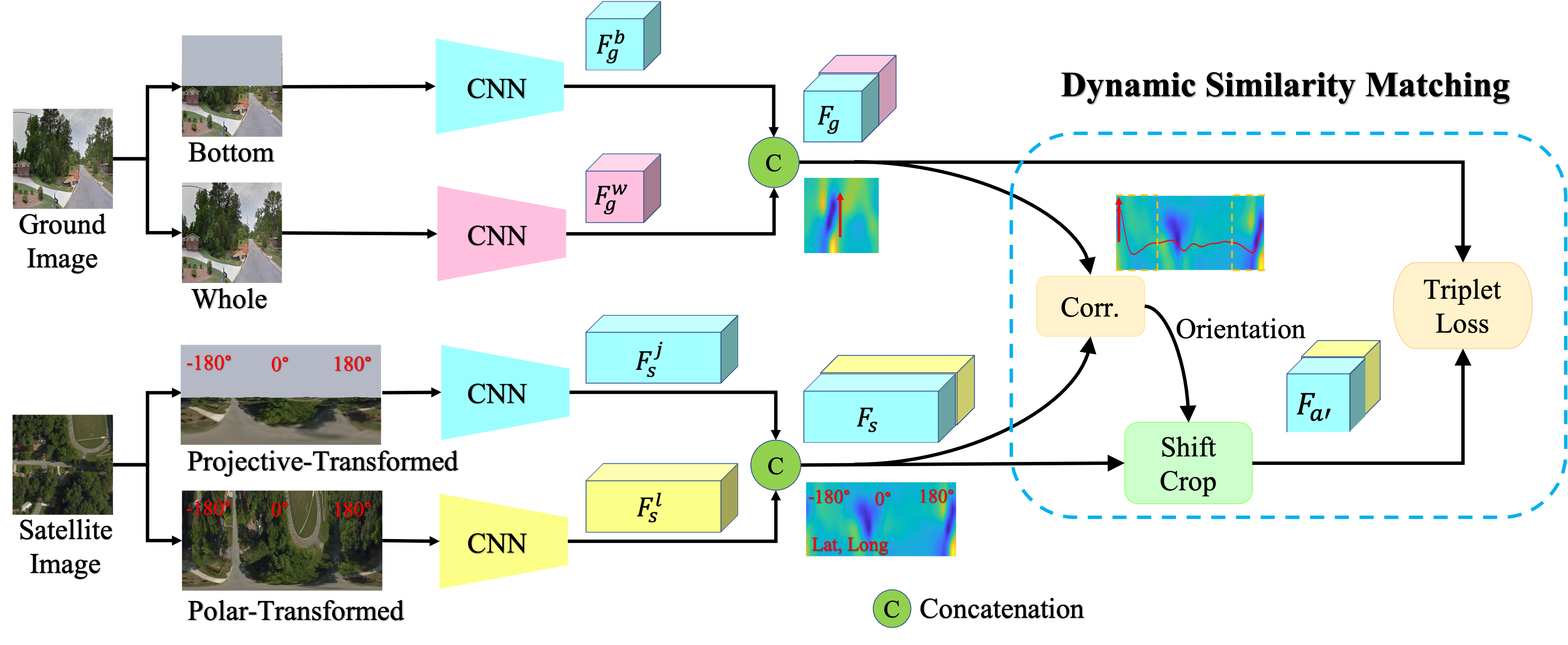}
    \caption{\footnotesize The overall framework of the proposed method. For the satellite image, we use a two-branch network that first applies a polar transform and a projective transform before extracting features with a CNN. For the ground image, we also use a two-branch network that takes the bottom half of the image corresponding to the projective-transformed satellite image and the whole image corresponding to the polar-transformed image before extracting features. Given the concatenated feature tensors, the correlation between the two streams is used for estimation of the orientation of the ground image with respect to the satellite image. Next, the satellite features are shifted and cropped to obtain the section that (potentially) corresponds to the ground features. The similarity of the resulting features is then used for the retrieval of location.}
    \label{fig:framework}
\end{figure*}

We note that for both the polar and projective transforms, the horizontal axis corresponds to the azimuth direction. 
We thus propose a Dynamic Similarity Matching (DSM) module to estimate the orientation of ground images with respect to satellite images.
Specifically, DSM computes the correlation between the ground and satellite features in order to generate a similarity score at each angle, denoted by the red curve in Figure \ref{fig:framework}. The argument of the similarity score maximum corresponds to the latent orientation of the ground image with respect to the satellite image. If the ground image has restricted FoV, we then extract the appropriate local region from the satellite feature representation for the use in the coarse localization stage. 
The output of our coarse localization stage is the satellite image that is the most similar to the query image in the database. 

Considering the displacement between the ground camera and the satellite image center, this article further introduces a fine-grained camera localization stage to localize a fine-grained position of the query ground image.
Specifically, we project the satellite image to the ground viewpoint at a predetermined set of points of projection, as shown in Figure~\ref{fig:intro_coarse_localization}.
The similarity between the projective-transformed satellite images and the query image is then computed in the same way as in the coarse matching stage.
The center of projection of the projective-transformed satellite image that is most similar to the query ground image is taken as the camera location, and the computed relative orientation is taken as the camera azimuth angle.
In particular, the precision of the localization result depends on the density of the sampled centers of projection, whereas the precision of the orientation estimation depends on the resolution of the images and the FoV.

Below, we detail novel contributions of this manuscript which are not explored in our earlier work.
\renewcommand{\labelenumi}{\roman{enumi}.}
\vspace{0.1cm}
\hspace{-1.0cm}
\begin{enumerate}[leftmargin=0.6cm]
\item  
Compared to previous methods which formulate the cross-view image-based geo-localization as a pure image retrieval task, this article introduces a new fine-grained localization method to compute the displacement between the retrieved satellite image center and the query ground camera location. 
\item 
We further extend our conference work~\cite{shi2020looking} for coarse camera localization. 
Apart from the polar transform introduced in our conference work, this article proposes a geometry-constrained projective transform to establish more realistic geometric correspondences of points on the ground plane between satellite and ground-level images. 
This allows our network to focus on informative regions for cross-view image matching.
\item 
We achieve new state-of-the-art cross-view localization performance compared to our conference work. 
Moreover, we analyze and discuss the different task properties for localizing restricted FoV and panorama images and shed some light on how to exploit the proposed method for different localization situations. 
\end{enumerate}

\section{Related Work}

{\color{black}
\noindent \textbf{3D structure-based localization. } 
The task of visual localization is to estimate the 6-DOF camera pose of a query image with respect to a 3D scene model~\cite{svarm2016city, donoser2014discriminative, li2012worldwide, li2010location, lynen2015get, sattler2015hyperpoints, sattler2016efficient, taira2018inloc, zeisl2015camera, sattler2018benchmarking, toft2020long, liu2019stochastic, cao2014minimal, larsson2016outlier, zhou2020da4ad}. 
The key to this problem is to establish efficient, accurate, and robust 2D-3D matches. 
This line of work can predict accurate 6-DoF camera poses.
Nevertheless, 3D models are not available everywhere, and they are usually expensive to obtain. 
This limits the applicability of these methods.  

\smallskip
\noindent \textbf{2D ground-to-ground image-based localization. }
Image-based localization aims to estimate the camera pose of a query image by matching it against a large geo-tagged database. 
It was originally approached as a ground-to-ground image matching task, which is often used for place recognition~\cite{arandjelovic2016netvlad, chen2017deep, lowry2015visual, sattler2016large, torii201524, sattler2017large, kim2017learned, noh2017large, ge2020self}, and loop-closure detection~\cite{cummins2008fab, galvez2012bags, mur2015orb}. 
In the ground-to-ground image-based localization task, 
both the query and database images are captured at ground level. 
The challenges are to address the difficulties of large viewpoint differences, illumination differences (\eg, day and night), and weather differences between query and reference images. 
However, ground-to-ground image matching cannot localize query images where no corresponding reference image is available, since the world is non-uniformly sampled by tourists and ground-level vehicles.

\smallskip
\noindent \textbf{2D ground-to-satellite image-based localization. }
Many recent works~\cite{castaldo2015semantic, lin2013cross, mousavian2016semantic,workman2015location, workman2015wide, vo2016localizing, tian2017cross,  zhai2017predicting, Hu_2018_CVPR, Liu_2019_CVPR, Regmi_2019_ICCV, Cai_2019_ICCV, sun2019geocapsnet, shi2019spatial, shi2020optimal, shi2020looking, zhu2021revisiting, toker2021coming} resort to satellite images as a reference set for image-based camera localization, due to the wide-spread coverage and easy accessibility of satellite imagery. 
Challenges of ground-to-satellite image matching include the significant visual appearance differences, geometric projection differences, and the unknown relative orientation between the two view images, as well as the limited FoV of query ground images. 
Existing works have focused on designing powerful network architectures~\cite{workman2015location, workman2015wide, vo2016localizing, Hu_2018_CVPR, Cai_2019_ICCV, sun2019geocapsnet}, bridging the cross-view domain gaps~\cite{Regmi_2019_ICCV, shi2020optimal, shi2019spatial, shi2020looking, toker2021coming}, and learning orientation invariant or equivariant features~\cite{Hu_2018_CVPR, sun2019geocapsnet, Liu_2019_CVPR, shi2020looking, zhu2021revisiting}. 

Although promising results have been achieved, 
almost all the approaches only estimate the location (latitude and longitude) of a query image but neglect the orientation misalignment.
Our prior work~\cite{shi2020looking} is the first attempt to estimate the 3-DOF camera pose (location and estimation) via ground-to-satellite image matching. 
However, all prior arts (including ours) only retrieve the most similar satellite image for a query image. 
As a result, the estimates of camera location are quite coarse. 
In this article, we provide a new mechanism for estimating the fine-grained location and orientation of the query ground image subsequent to the coarse localization stage.
}

\begin{figure}[t!]
\setlength{\abovecaptionskip}{0pt}
\setlength{\belowcaptionskip}{0pt}
    \centering
    \subfloat[Satellite \protect\\ \centering Image]{
    \begin{minipage}{0.15\linewidth}
    \includegraphics[width=\linewidth, height=\linewidth, height=\linewidth]{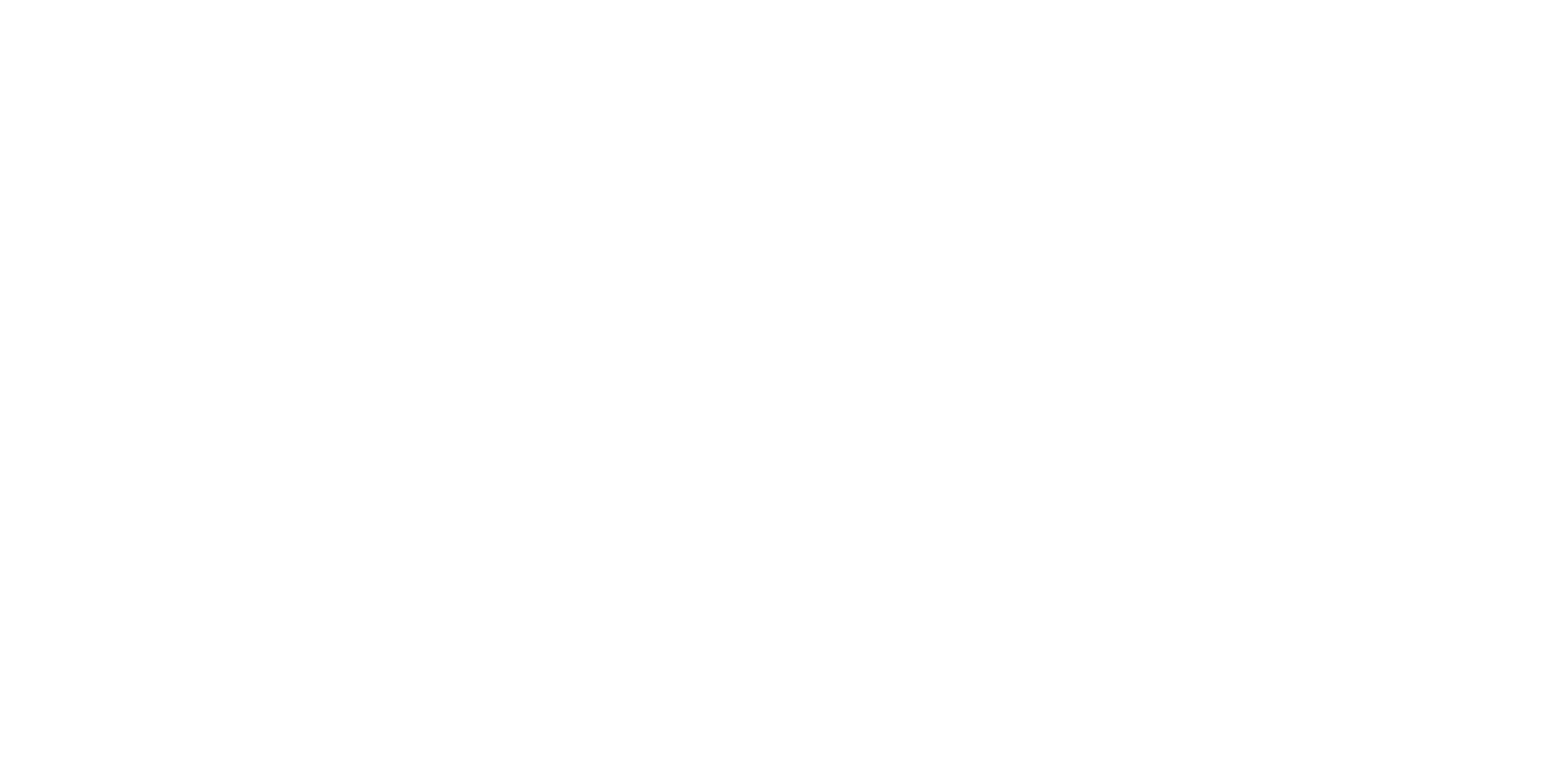}
    \includegraphics[width=\linewidth]{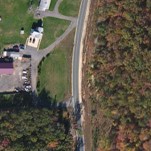}
    \label{fig: challenge_sat}
    \end{minipage}
    }
    \subfloat[Ground Image]{
    \begin{minipage}{0.75\linewidth}
    \includegraphics[width=\linewidth, height=0.2\linewidth]{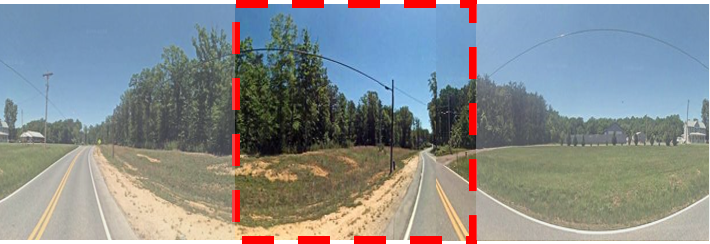}
    \includegraphics[width=\linewidth, height=0.2\linewidth]{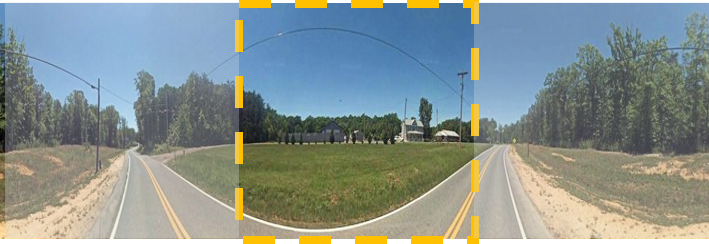}
    \end{minipage}
    \label{fig: challenge_orien_grd}
    }\\
    \begin{minipage}{0.16\linewidth}
    \includegraphics[width=\linewidth, height=0.6\linewidth]{images_frameworks_intro_white_img.jpg}
    \end{minipage}
    \setcounter{subfigure}{2}
    \subfloat[Satellite Image (Polar Transform)]{
    \begin{minipage}{0.75\linewidth}
    \includegraphics[width=\linewidth, height=0.2\linewidth]{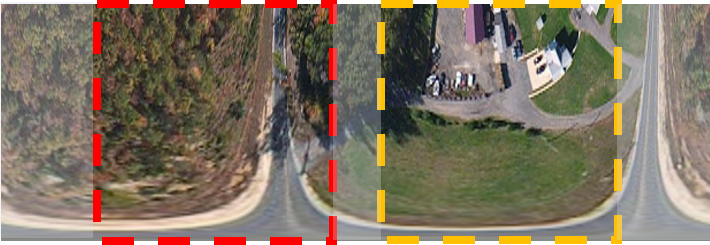}
    \end{minipage}
    \label{fig: challenge_FoV_polar}
    }\\
    \begin{minipage}{0.16\linewidth}
    \includegraphics[width=\linewidth, height=0.6\linewidth]{images_frameworks_intro_white_img.jpg}
    \end{minipage}
    \setcounter{subfigure}{3}
    \subfloat[Satellite Image (Projective Transform)]{
    \begin{minipage}{0.75\linewidth}
    \includegraphics[width=\linewidth, height=0.2\linewidth]{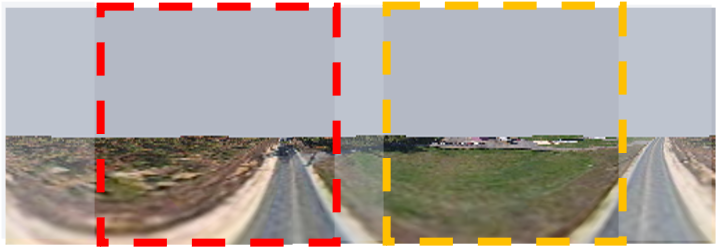}
    \end{minipage}
    \label{fig: challenge_FoV_geo}
    }\\
    \caption{\footnotesize The challenges of cross-view image matching: the orientation of the query ground image is unknown, and its FoV is restricted. The scene content in panoramas captured at the same location but with different azimuth angles is offset. The image content in an image with a restricted FoV can be entirely different from another image captured from the same location, indicated by different boxes in (b).
    {\color{black} The polar-transformed satellite image (c) is an approximation to the ground panorama, which preserves all information from the original satellite image. The projective-transformed satellite image (d) loses some information, but preserves the ground-level geometry.}
    }
    \label{fig:illustration of unknwon orien and limited FoV}
\end{figure}

\section{Image Retrieval for Coarse Camera Geo-Localization}
The approach proposed in this article involves a coarse localization stage in which image retrieval techniques are used to roughly estimate where an image was taken, and a fine-grained localization stage where the displacement between the query ground camera and the center of the retrieved satellite image is computed. 
In this section, 
we outline the extended framework for coarse camera geo-localization in Figure~\ref{fig:framework}. 

For the cross-view geo-localization task, query images are captured at ground level, and satellite images in the database are captured from an overhead view.
Since there are large appearance variations between these two image domains, our strategy is first to reduce the projection differences between the viewpoints and then to extract discriminative features from the two domains.
Furthermore, inspired by how humans localize themselves~\cite{halligan2003spatial, mania2010cognitive}
, we exploit the spatial relationships between objects as a critical cue for inferring location and orientation. To this end, we enable our descriptors to encode the spatial relationship among the features, as indicated by $F_g$ and $F_s$ in Figure~\ref{fig:framework}.

Despite the discriminativeness of the spatially-aware features, they are very sensitive to orientation changes. For instance, when the azimuth angle of a ground camera changes, the scene contents will be shifted in the ground panorama, and the image content may be dramatically different if the camera has a limited FoV, as illustrated in Figure~\ref{fig:illustration of unknwon orien and limited FoV}. Therefore, finding the orientation of the ground images is crucial to make the spatially-aware features meaningful. To this end, we propose a Dynamic Similarity Matching (DSM) module, as illustrated in Figure~\ref{fig:framework}. With this module, we not only estimate the orientation of the ground images but also achieve more accurate feature matching scores, regardless of orientation misalignments and limited FoVs, thus enhancing the performance of geo-localization.

\subsection{Bridging the domain gap by the polar transform}
When a scene is planar, a horizontal line in the ground-level panorama corresponds to a circle in the satellite image, and a vertical line in the ground-level panorama corresponds to a ray starting from the center of the satellite image. 
This layout correspondence motivates us to apply a polar transform to the satellite images.
In this way, the spatial layouts of these two domains can be roughly aligned, as illustrated in Figure \ref{fig: challenge_orien_grd} and \ref{fig: challenge_FoV_polar}.
To be specific, the polar origin is set to the center of each satellite image, corresponding to the geotag location, and the $0^{\circ}$ angle is chosen as the northward direction, corresponding to the upwards direction of an aligned satellite image.
In addition, we constrain the height (\ie, vertical resolution) of the polar-transformed satellite images to be the same as the ground images, and ensure that the angle subtended by each column of the polar transformed satellite images is the same as in the ground images.

We apply a uniform sampling strategy along rays to the satellite image, such that the innermost and outermost circles of the satellite image are mapped to the bottom and top line of the transformed image respectively.
Formally, let $S \times S$ be the size of the satellite image and $H_g \times W_g$ be the target size of polar transform. The polar transform between the original satellite image pixels $(u_i^s, v_i^s)$ and the target polar transformed pixels $(u_i^t, v_i^t)$ is expressed as
\begin{equation}
\small
 \label{PT}
\left\{\begin{matrix}
\begin{aligned}
u_{i}^{s} & = u_0 - r(H_g - v_i^t) \cos (2 \pi u_i^t / W_{g}) /H_g, \\
v_{i}^{s} & = v_0 + r(H_g - v_i^t) \sin (2 \pi u_i^t / W_{g}) /H_g,\\
\end{aligned}
\end{matrix}\right.
\end{equation}
where $(u_0, v_0)$ is the satellite image center, and $r$ is the maximum radius for the polar transform and set to $S/2$.

\subsection{Bridging the domain gap by the projective transform}

\label{subsec:geometry_transform}
\begin{figure}[!t]
\setlength{\abovecaptionskip}{0pt}
\setlength{\belowcaptionskip}{0pt}
    \centering
    \scalebox{0.75}[0.65]{\includegraphics[width=1\linewidth]{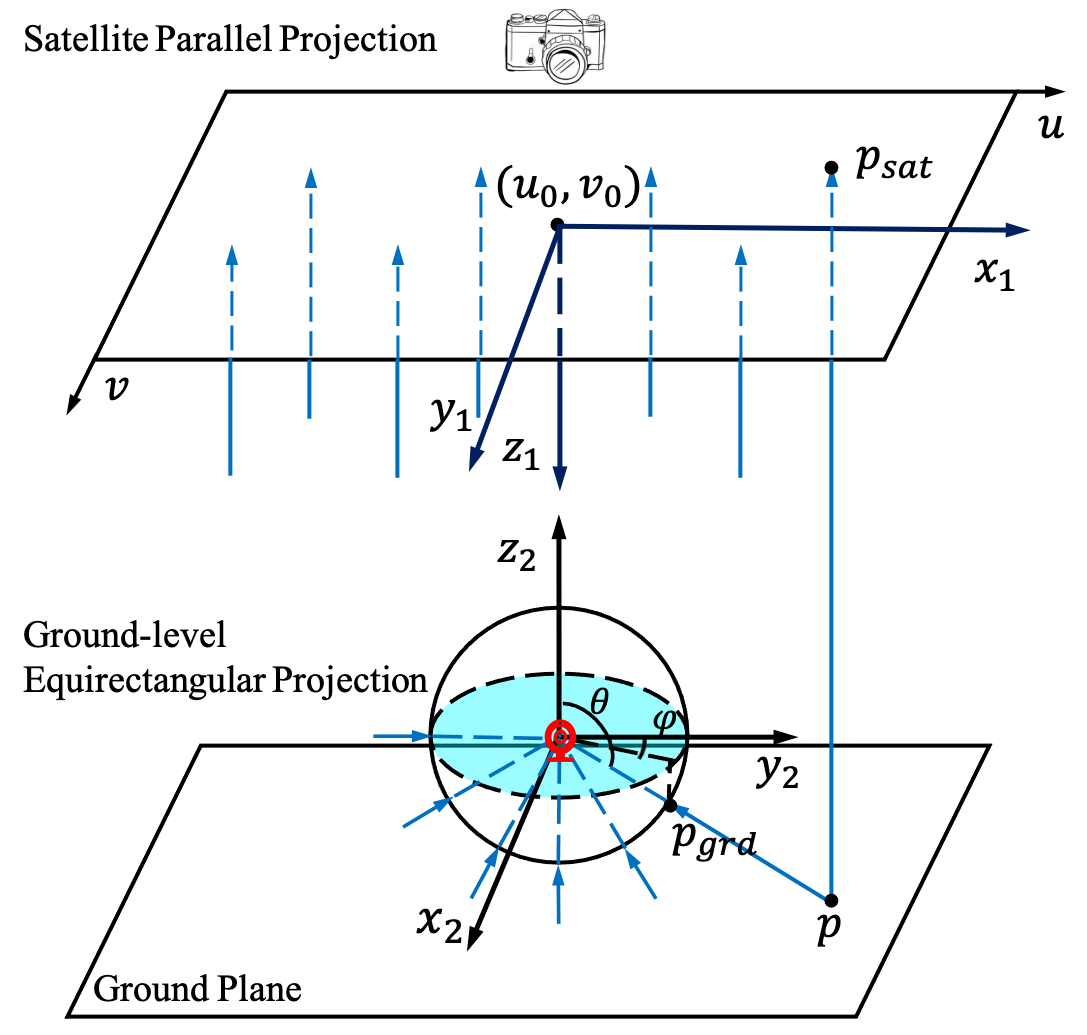}}
    \caption{\footnotesize Illustration of latent geometric correspondences between a satellite image and a ground-level panorama for pixels on the ground plane.}
    \label{fig:geometry_transform}
\end{figure}

The polar transform is a simple approximation for the cross-view image transformation. 
In this article, we further establish more realistic geometric correspondences between the satellite image and the ground-level panorama, especially for scenes that are planar and lie on the ground plane.

\vspace{0.05cm}
\noindent{\textbf{Transformation between satellite and ground-level cameras.}} 
As illustrated in Figure~\ref{fig:geometry_transform}, we use $(x_1, y_1, z_1) $ to represent the satellite camera coordinates and $(x_2, y_2, z_2)$ to denote the ground-level camera coordinates.
The transformation between the two camera coordinate systems is expressed as
\begin{equation}
\small
    \begin{bmatrix}
x_1\\ 
y_1\\ 
z_1\\ 
{\color{black} 1}
\end{bmatrix} 
= 
\begin{bmatrix}
0 & 1 & 0 & 0\\ 
1 & 0 & 0 & 0\\ 
0 & 0 & -1 & \mathcal{H} \\ 
{\color{black} 0} & {\color{black}0} & {\color{black}0} & {\color{black}1}
\end{bmatrix}
\begin{bmatrix}
x_2\\ 
y_2\\ 
z_2\\ 
{\color{black} 1}
\end{bmatrix},
\label{eq:world_transform}
\end{equation}
where $\mathcal{H}$ is the height of the satellite above the ground level.

\vspace{0.05cm}
\noindent{\textbf{Satellite camera coordinate system.}} 
The satellite camera projects a point $(x_1, y_1, z_1)$ to its image coordinates $(u_i^s, v_i^s)$ by a parallel projection

\begin{equation}
\small
    \begin{bmatrix}
u_i^s\\ 
v_i^s
\end{bmatrix} 
= 
\begin{bmatrix}
s & 0 & 0 & u_0\\ 
0 & s & 0 & v_0
\end{bmatrix}
\begin{bmatrix}
x_1\\ 
y_1\\ 
z_1\\ 
1
\end{bmatrix}
\label{eq:sat_transform},
\end{equation}
where $s$ is the resolution of the satellite image and $(u_0, v_0)$ is the satellite image center.

%\smallskip
\vspace{0.05cm}
\noindent{\textbf{Ground-level spherical camera coordinate system.}} 
In the ground-level camera coordinate system, we define $\theta$ as the elevation angle with respect to the $z_2$ axis and $\phi$ as the azimuth angle. 
To be consistent with the polar transform, $\phi=0^\circ$ is the northward direction and it corresponds to the negative direction along $x_2$ axis. 
For both $\theta$ and $\phi$, clockwise is the positive direction.
The mapping between $(\theta, \phi)$ and $(x_2, y_2, z_2)$ is computed as 
\begin{equation}
\small
\left\{\begin{matrix}
x_2 = z_2 \tan \theta \cos \phi,\\ 
y_2 = {\color{black} - } z_2 \tan \theta \sin \phi,\\
\end{matrix}\right.
\label{eq:grd_theta_phi}
\end{equation}
and the projection between $(\theta, \phi)$ and ground-level panorama image coordinates $(u_i^t, v_i^t)$ is expressed as
\begin{equation}
\small
% \left\{\begin{matrix}
% \theta &= \pi v_i^t/H_g\\ 
% \phi &= 2\pi u_i^t/W_g\\
% \end{matrix}\right.
\left\{\begin{matrix}
\begin{aligned}
\theta &= \pi v_i^t/H_g,\\ 
\phi &= 2\pi u_i^t/W_g.\\
\end{aligned}
\end{matrix}\right.
\label{eq: grd_theta_phi_2_img}
\end{equation}

%\smallskip
\vspace{0.05cm}
\noindent{\textbf{Projection between satellite and ground-level images.}} 
Finally, the mapping between the satellite image coordinates $(u_i^s, v_i^s)$ and ground-level panorama image coordinates $(u_i^t, v_i^t)$ is established as
\begin{equation}
\small
\left\{\begin{matrix}
u_i^s = u_0 + s z_2 \tan(\pi v_i^t/H_g) \cos(2 \pi u_i^t/W_g),\\
v_i^s = v_0  {\color{black}-} s z_2 \tan(\pi v_i^t/H_g) \sin(2 \pi u_i^t/W_g),\\
\end{matrix}\right.
\end{equation}
where $z_2$ represent the scene height at pixel $(u_i^t, v_i^t)$. 
% \YS{It seems that I don't need to mention this? $z_2$ is easily observed from Figure 6}
Similar to the polar transform, we choose the satellite image center $(u_0, v_0)$ as the projection point of the projective transform. 

In practice, the height maps of satellite images are hard to obtain and the scene heights are much smaller compared to the distance between the ground and the satellite camera.  
Thus, we assume all the pixels in the satellite image lie on the ground plane.
Therefore, $z_2$ is set to the height of ground plane with respect to the ground camera, and $0.5H_g < v_i^t \le H_g$ because the ground plane is mostly projected to the bottom half of a ground-view image.   
In doing so, we construct the geometric correspondences between satellite and ground images for points that lie on the ground plane. 
% \YS{Newly added}
For clarity, we call this projection as ``projective transform''. 

Note that both the polar transform and the projective transform are implemented in an inverse warping manner. Thus, they are differentiable and applicable in end-to-end training. 
In our implementation, they are applied as a preprocessing step in order to reduce the computation time during both training and testing steps. 

\begin{figure}
\setlength{\abovecaptionskip}{0pt}
\setlength{\belowcaptionskip}{0pt}
    \centering
    \subfloat[Polar Transform]{
     \scalebox{1.0}[1.0]{\includegraphics[trim={35mm 100mm 30mm 90mm}, clip, width=0.4\linewidth]{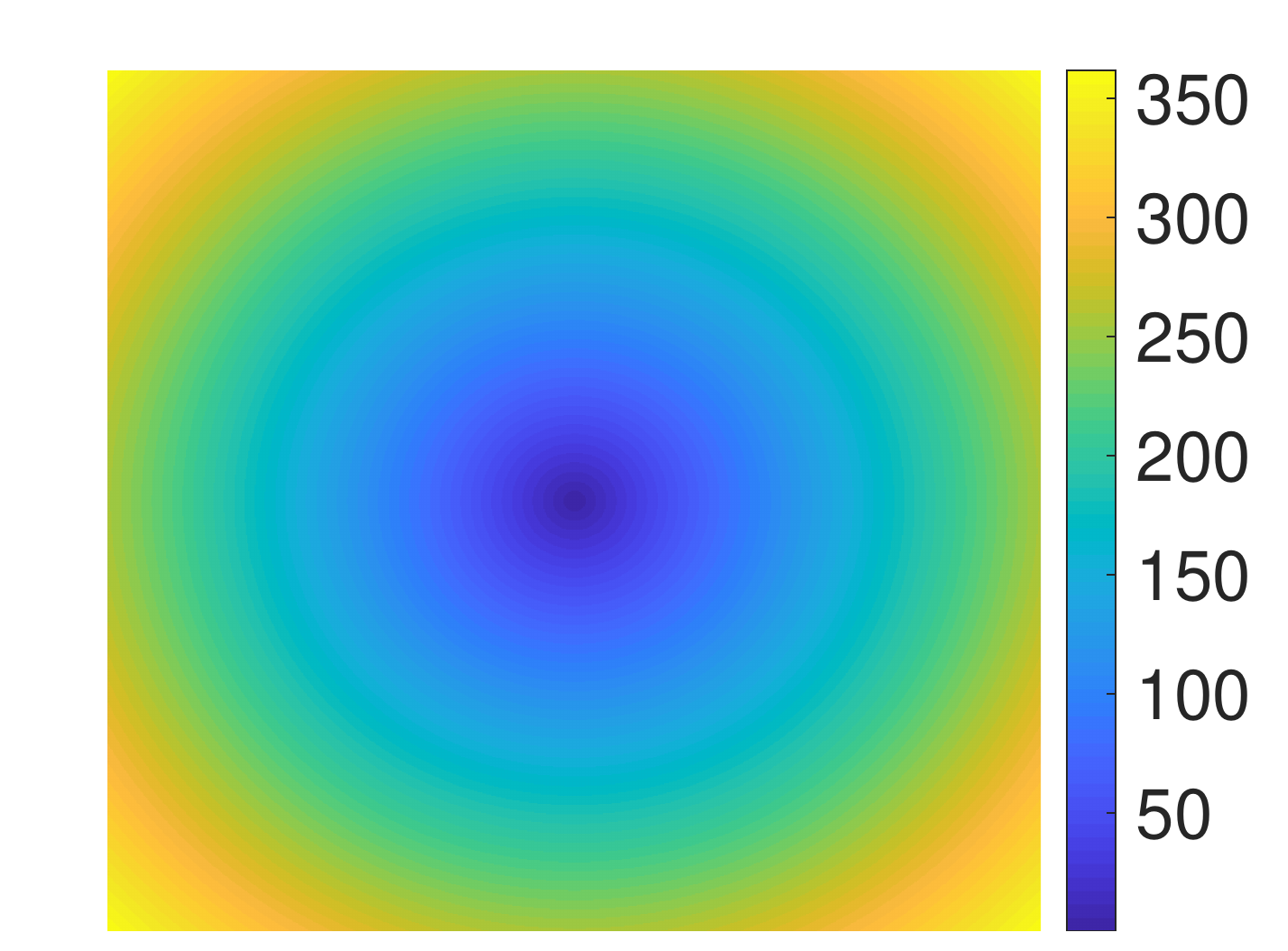}}
    \label{fig:coordinates_mapping_pt}
    }
    \hspace{1em}
    \subfloat[Projective Transform]{
     \scalebox{1.0}[1.0]{\includegraphics[trim={35mm 100mm 30mm 90mm}, clip, width=0.4\linewidth]{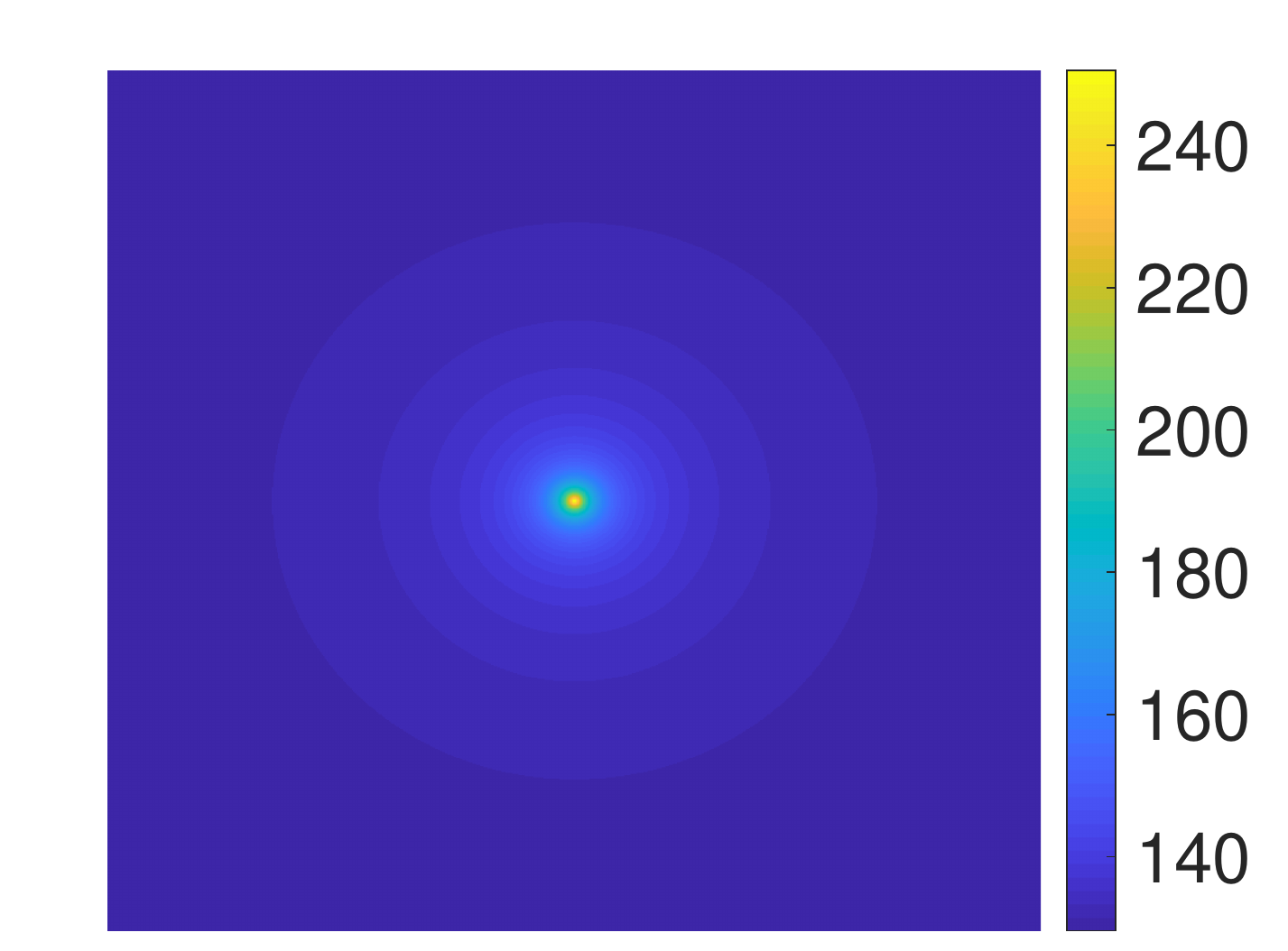}}
    \label{fig:coordinates_mapping_gt}
    }
    \vspace{0.5em}
    \caption{\footnotesize Visualization of the source (satellite) and target (ground-level panorama) coordinate correspondences by (a) the polar transform and (b) the projective transform.
    The spatial positions corresponds to the satellite image pixels. 
    Different colors in the two images indicate the row number in the target coordinates. 
    % The correspondences between different colors and row numbers is illustrated by a color bar at the right of each figure. 
    % The maximum row number in the target image is 256. 
    }
    \label{fig:coordinates_mapping}
\end{figure}

\subsection{Complementary between the transforms}

In Figure~\ref{fig: intro_Polar-transformed} and \ref{fig: intro_Geometry-transformed}, and Figure~\ref{fig: challenge_FoV_polar} and \ref{fig: challenge_FoV_geo}, we present two examples of the polar-transformed and projective-transformed satellite images.
Note that, although the projective transform only retains pixels in a small region of the center of a satellite image, with others pixels being occluded, the scene geometric structure is better preserved by the projective transform than polar transform. This is also reasonable since scene objects far away from the center of a satellite image are unlikely visible in the ground-level panorama.

Furthermore, we illustrate the mapping relationship between the source (satellite) and target (ground-level panorama) coordinates of the two transforms in Figure~\ref{fig:coordinates_mapping}. As the projection method along the azimuth direction (image columns in target coordinates) is the same in both transforms, we compare their projection difference on mapped image row coordinates.
Figure~\ref{fig:coordinates_mapping} manifests that most of the pixels are mapped to a few rows in a target image (same color) by the projective transform, while the polar transform maps satellite image pixels to different rows in a target image (indicated by different colors) uniformly. 
This indicates that the polar transform retains nearly all the information of a satellite image and the projective transform is able to highlight the ground-level scenes and their structure that would be visible in ground-level panoramas. 

As illustrated in Figure~\ref{fig: intro_Polar-transformed} and \ref{fig: intro_Geometry-transformed}, and Figure~\ref{fig: challenge_FoV_polar} and \ref{fig: challenge_FoV_geo}, the projective transformed and polar transformed images provide complementary information for the cross-view matching. Hence, we apply both polar and projective transforms to bridge the cross-view domain gap.

\subsection{Spatially-aware feature representation}

Applying a translation offset along the horizontal axis of a polar-transformed or projective-transformed image is equivalent to rotating a satellite image. 
Hence, the task of learning the rotation equivariant features for satellite images is re-formulated into learning translation equivariant features on polar or projective transformed images. Doing so significantly reduces the learning difficulty of our network since CNNs are inherently translation equivariant \cite{lenc2015understanding, cohen2016group, worrall2017harmonic}. 
Because the horizontal axis corresponds to rotation degrees, we need to ensure that the CNN treats the leftmost and rightmost columns of the transformed image as adjacent neighbours.
Hence, we employ circular convolutions~\cite{esteves2018polar} with periodical padding along the horizontal direction.

Figure~\ref{fig:framework} illustrates the pipeline of our coarse localization framework. 
As the traits 
of representations resulting from polar- and projective-transformed satellite images are different in nature, we adopt separate CNNs to extract features from them. 
For the ground images, we also employ separate CNNs. One focuses on the bottom half of the ground image and it is expected to learn similar feature representations to the projective-transformed satellite image. The other one extracts features from the whole ground image and aims to learn matching features with respect to the polar transformed satellite images.
The four branches in Figure~\ref{fig:framework} have the same architecture. 
Since the projective-transformed satellite images share the same domain as the ground plane images, we employ the same weights in the first and third branch in Figure~\ref{fig:framework} (indicated by the blue color) to extract features and then enforce their similarity. 
The other two branches, marked with different colors, do not share weights so that they can adapt to their individual domains. 

We adopt VGG16~\cite{Simonyan2014VeryDC} as the network backbone. 
In particular, the first ten layers of VGG16 are used for features extraction. 
As the vertical direction of the images may include irrelevant features, such as sky, missing pixels and distortions in the transformed satellite images, we modify the subsequent three layers to decrease the vertical resolution of the feature maps while maintaining their horizontal resolution. 
In this manner, our extracted features are more tolerant to distortions along the vertical direction while retaining information along the horizontal direction. 
We also decrease the number of feature channels to $8$. % by using these three convolutional layers. 
For each branch in Figure~\ref{fig:framework}, the output feature is of size $4\times64\times8$. 
The extracted features in each stream are then concatenated together to form a global feature descriptor.
Hence, for a satellite image or a ground-level panorama, the size of their global descriptors is $4\times64\times16$.
Following our previous work~\cite{shi2019spatial}, the extracted feature volumes are used to preserve the spatial layout of scenes, thus improving the discriminativeness of the descriptors.
For query ground images with limited FoV, the width of the extracted features is decreased proportionally.

\subsection{Dynamic Similarity Matching (DSM)}
\label{subsec:DSM}
When the orientation of ground and transformed satellite features is aligned, their descriptors can be easily compared. 
However, the orientation of the ground images is not always available, and orientation misalignments increase the difficulty of geo-localization significantly, especially when the ground image has a limited FoV. 
When humans use a map to localize themselves, they determine their locations and orientation jointly by matching what they have seen to what a map shows~\cite{halligan2003spatial, mania2010cognitive}.
In order to let the network mimic this process, we compute the correlation between the ground and satellite descriptors along the azimuth angle axis.
To be specific, we use the ground descriptors as a sliding window and then compute the inner product between then ground and satellite descriptors across all possible orientation. 

Let $F_s \in R^{H \times W_s \times C}$ and $F_g \in R^{H \times W_g \times C}$ denote the satellite and ground descriptors, respectively. Let $H$ and $C$ indicate the height and channel number of the descriptors, and $W_s$ and $W_g$ indicate the width of the satellite and ground descriptors. The correlation between $F_s$ and $F_g$ is expressed as
\begin{equation}
\small
    [F_s * F_g](i) \! = \!\!\sum_{c=1}^{C} \sum_{h=1}^{H} \sum_{w=1}^{W_{g}} \! F_s(h, \text{mod}(i+w, W_s), c) F_g(h, w, c),
 \label{Eq: corr}
\end{equation}
where $F(h, w, c)$ is the feature response at index ($h$, $w$, $c$) while operator $\text{mod}$ denotes the modulo operation.
Having computed the correlation, we take the argument of the maximum similarity scores as our estimated orientation misalignment of the ground image with respect to the transformed satellite image. 

We normalize $F_s$ and $F_g$ by the $\ell_2$ norm before calculating the correlation results for panorama images.  %\PK{$\ell_2$-norm normalized?}\YS{Both 'norm' and 'normalized' should be here? Would that be repeated?}. 
When a ground image has a limited FoV, we crop the transformed satellite features corresponding to the FoV of the ground image.
% at the position of the maximum similarity score.
Then we re-normalize the cropped satellite features and calculate the $\ell_2$ distance between the ground and satellite descriptors as the similarity score.
% for matching.
Note that if there are multiple maxima in the similarity scores, the satellite image contains indistinguishable symmetries. Thus, we choose one of these maxima at random.

\subsection{Training DSM}
During the training process, our DSM module is applied to all ground and satellite pairs, regardless of whether they are matching or not. 
For matching pairs, DSM forces the network to learn similar feature embeddings for ground and transformed satellite images as well as discriminative feature representations along the horizontal direction (\ie, azimuth).
In this way, DSM is able to identify the orientation misalignment as well as find the best feature similarity for matching.
For non-matching pairs, we firstly find the orientation with the highest similarity using the DSM module, and then minimize the maximum similarity score of non-matching pairs to make the descriptors more discriminative. 
Following traditional cross-view localization methods \cite{Hu_2018_CVPR, Liu_2019_CVPR, shi2020optimal}, we employ the weighted soft-margin triplet loss \cite{Hu_2018_CVPR} to train our network. 
Our training loss is expressed as

\begin{equation}
\small
\begin{aligned}
    \mathcal{L} & = \log \left( 1+ e^{\alpha \big \| F_g^b - F_{s'}^j \big \|_F - \alpha\big \| F_g^b - F_{s^{\ast'}}^j \big \|_F }\right ) \\
     &+ \log \left( 1+ e^{\alpha \big \| F_g^w - F_{s'}^l \big \|_F - \alpha\big \| F_g^w - F_{s^{\ast'}}^l \big \|_F }\right )\\
     &+ \log \left( 1+ e^{\alpha \big \| F_g - F_{s^{'}} \big \|_F - \alpha\big \| F_g - F_{s^{\ast'}} \big \|_F  }\right ) .
\end{aligned}
\end{equation}
In the above equations, $F_g^b$ and $F_g^w$ denote the extracted features from the bottom half of a query ground image and the whole ground image, 
$F_{s'}^j$ and $F_{s'}^l$ are the cropped features from the non-matching projective-transformed 
satellite features and the polar-transformed satellite features, 
$F_{s^{\ast'}}^j$ and $F_{s^{\ast'}}^l$ denote the cropped features from the matching projective-transformed satellite features and the polar-transformed satellite features.
Recall that we do not require ground view query images to be panoramas, and the transformed satellite images will be cropped automatically to fit the resolution of the ground images by the DSM module. 
$F_g = [F_g^b, F_g^w]$ is the global query ground descriptor, 
$F_{s^{\ast'}} = [F_{s^{\ast'}}^j, F_{s^{\ast'}}^l]$ and $F_{s^{'}}=[F_{s'}^j, F_{s'}^l]$  indicate the cropped satellite descriptors of the matching satellite image and a non-matching satellite image aligned by our DSM module, $[\cdot,\cdot]$ is the concatenation operation, and $\left \|\cdot \right \|_F$ denotes the Frobenius norm. The parameter $\alpha$ controls the convergence speed of training process. Following precedent \cite{Hu_2018_CVPR, Liu_2019_CVPR, shi2020optimal}, we set it to $10$. 

\begin{figure*}[t!]
    \setlength{\abovecaptionskip}{0pt}
\setlength{\belowcaptionskip}{0pt}
    \subfloat[FoV = $360^{\circ}$]{
    \includegraphics[width=0.95\linewidth]{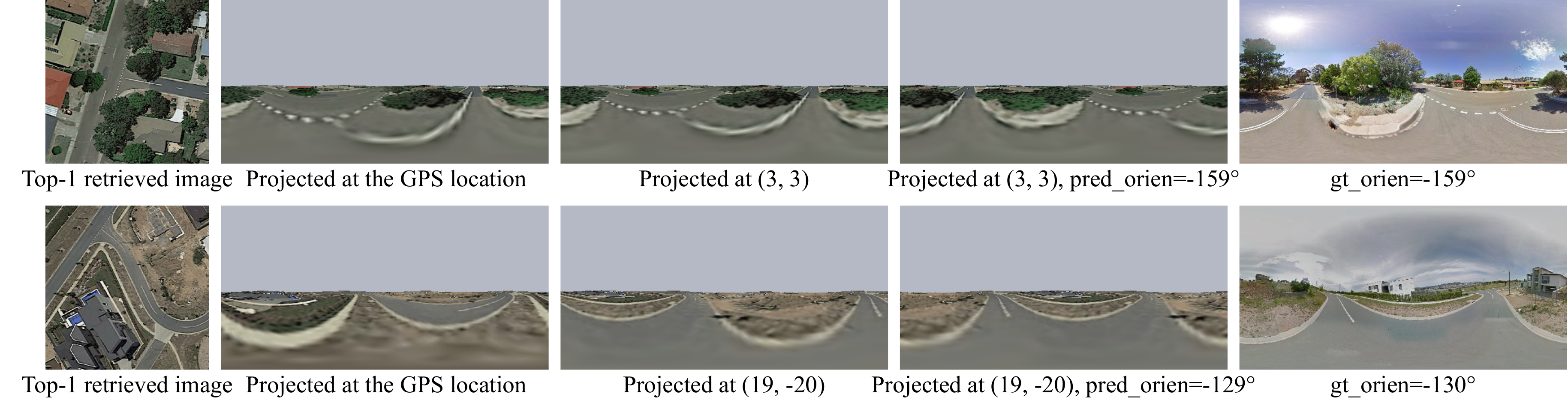}
    }\\
    \subfloat[FoV = $180^{\circ}$]{
    \includegraphics[width=0.91\linewidth]{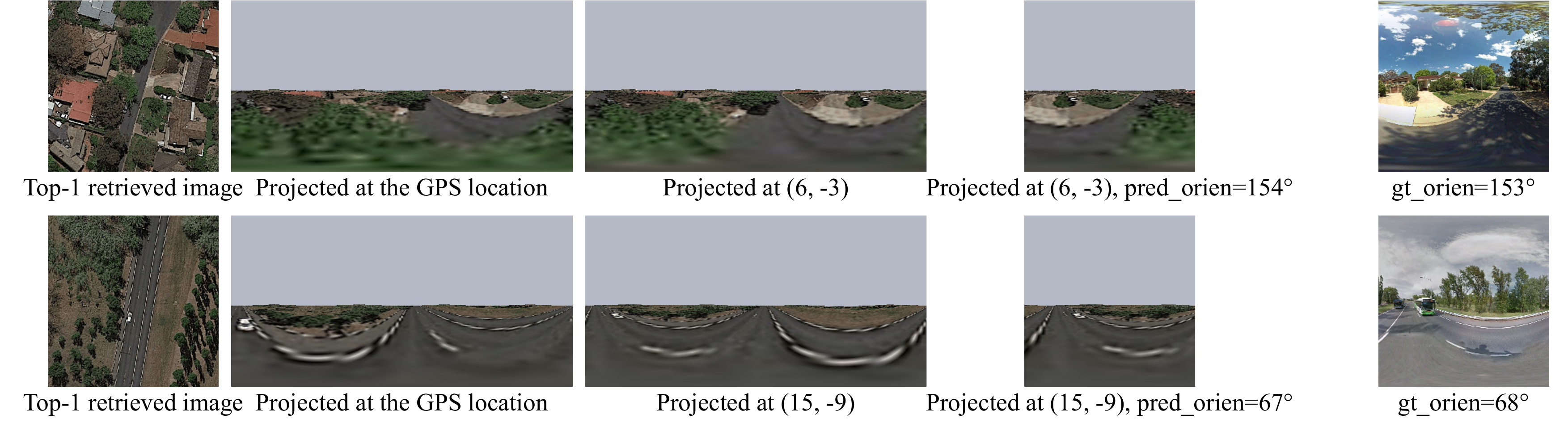} %0.958
    }
    \caption{\footnotesize
    Qualitative illustration of fine-grained 3-DoF camera localization for query images with unknown orientation and varying FoVs. Given query images (the last column), we first retrieve their most similar satellite images (the first column) from the database. The projective-transformed satellite images according to {\color{black}the query camera GPS locations} are presented in the second column. The ground structure of those images (the second column) is significantly different from the query images (the last column), {\color{black}indicating that the GPS locations are not accurate.} 
    According to our fine-grained camera geo-localization method, we exhaustively project the retrieved satellite image to their ground panorama coordinates at points in a central square region of the retrieved satellite image. 
    Among the projected images, the most similar ones to the query ground images are presented in the third column. 
    The fourth column shows the shifted and cropped projective-transformed satellite images that align with query images. 
    }
    \label{fig:accurate_localization}
\end{figure*}

\section{Fine-grained Cross-View Image Matching}
After the coarse geo-localization process, we retrieve an satellite image that is the most similar to the query ground image. 
However, the GPS tag of the satellite image is associated with its center point, whereas the ground image might be captured away from the center of the satellite image. 
Therefore, in this article we further propose a novel mechanism to compute the displacement between the query ground camera and the retrieved satellite image center. 

To achieve this goal, we employ the projective transform explained in Section~\ref{subsec:geometry_transform}.
As indicated in Figure~\ref{fig:coordinates_mapping}, only a small portion of the satellite image is projected to the transformed image. Thus, the projective transform is  sensitive to the projection center.
This phenomenon is useful in aiding fine-grained localization of where a query image was taken. 

In the fine-grained camera geo-localization stage, we firstly select a central square region in the retrieved satellite image, as shown in Figure~\ref{fig: intro_accurate_localization}. 
This central region encloses the possible query camera locations. 
At each pixel in the selected region, we project the satellite image to the corresponding ground panorama coordinates by applying the projective transform. 
Then, a query image with unknown orientation (or even with limited FoV) is compared with each of the projective-transformed satellite images via the dynamic similarity matching module.
Specifically, we circularly shift the projective-transformed satellite image along the azimuth (horizontal) direction. If the query ground image has a restricted FoV, we also crop out a portion of the transformed image according to its FoV. 
The similarity between the transformed image and the query ground image is then computed. 
Note that in this fine-grained camera localization process we use the SSIM as the similarity measure instead of the cross-correlation as SSIM is more suitable to evaluate structural differences between the projective-transformed satellite image and the query ground image. 

For each projective-transformed satellite image, we record the maximum similarity across different orientations as its similarity to the query image. 
Among all the projective-transformed satellite images, the most similar one is selected, as marked by the green box in Figure~\ref{fig: intro_accurate_localization}. 
Its corresponding projection point, indicated by the blue dot in Figure~\ref{fig: intro_accurate_localization}, is taken as the query camera location. 
The computed relative orientation is taken as the camera orientation. 
Under such a scheme, the precision of the location and orientation estimation depends on the real-world distance of a satellite image pixel and the resolution of the query image, respectively.

Figure~\ref{fig:accurate_localization} provides some qualitative examples of fine-grained  localization of orientation-unknown images. 
The query images are presented in the last column and their corresponding retrieved top-1 satellite images are shown in the first column. 
Since the geotags of the database satellite images are associated with the image centers, we firstly apply the projective transform to the retrieved satellite images according to their image centers, which is also the GPS location of query ground images provided by the dataset. 
The projected images are presented in the second column of Figure~\ref{fig:accurate_localization}.
They are significantly different from the query ground images. 
We next apply the proposed fine-grained camera localization method, and visualize the most similar projective-transformed satellite images in the third column. 
The fourth column provides the shifted and cropped projective-transformed satellite images according to the estimated relative orientation and the FoV of ground images. 
The estimated camera location with respect to the satellite image center and the relative orientation are presented under each of the images. 
As can be seen, images presented in the fourth column align with the original query ground images.

Since in the second stage (fine-grained camera localization) we search exhaustively every point in the selected central region, we recommend that the reference satellite images in the first stage (coarse camera localization) should cover the entire region as densely as possible. 
By doing so, the selected central region used for fine-grained camera geo-localization will be small, reducing the computation complexity. 

In our implementation, 
    the satellite image is resized to $256\times 256$ pixels from $1200\times 1200$ pixels. 
    The central square region covers $40\times 40$ pixels in the resized image, 
{\color{black} corresponding to $11.25 \times 11.25$ square meters ($\pm 5.625$ meters to the satellite image center).}
% The original satellite image size is $256\times 256$ pixels. 
{\color{black} 
% The ground-level panorama size is $128\times 512$ pixels.
By searching the $40\times 40$ candidate locations and the $512$ candidate orientations ($40\times 40 \times 512 = 614, 400$ candidate solutions in total), it takes $15$ minutes on average for the fine-grained localization of a query image on an RTX 2080 Ti GPU.}

\section{Experiments}

\subsection{Datasets}

\begin{figure}[t!]
\setlength{\abovecaptionskip}{0pt}
\setlength{\belowcaptionskip}{0pt}
    \centering
    \includegraphics[width=0.17\linewidth, height=0.17\linewidth]{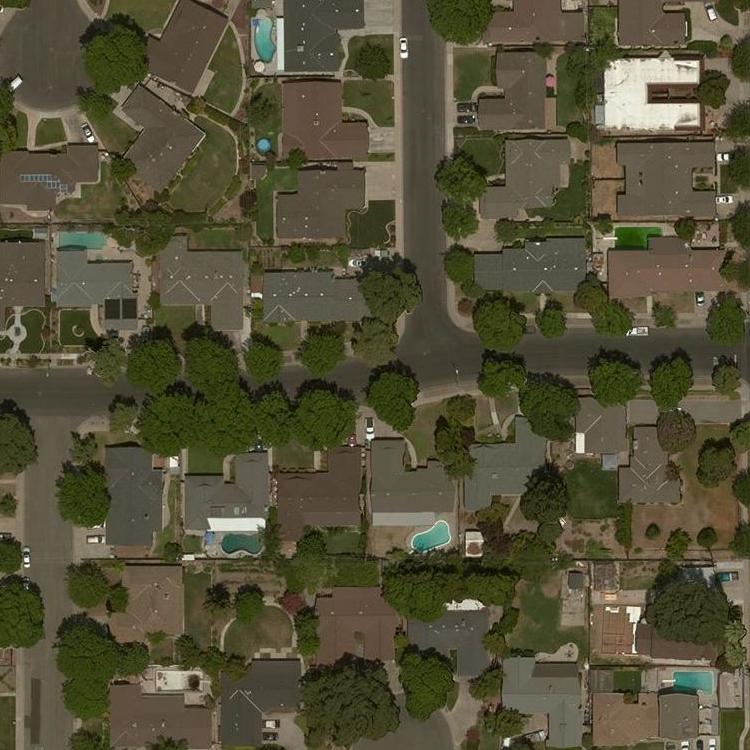}
    \includegraphics[width=0.75\linewidth, height=0.17\linewidth]{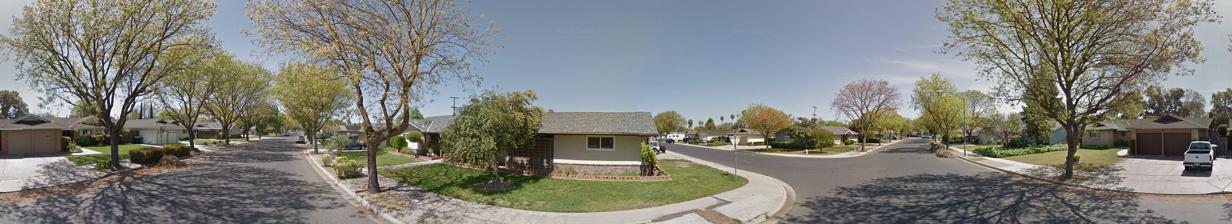}\\ 
    \smallskip
    \includegraphics[width=0.17\linewidth, height=0.17\linewidth]{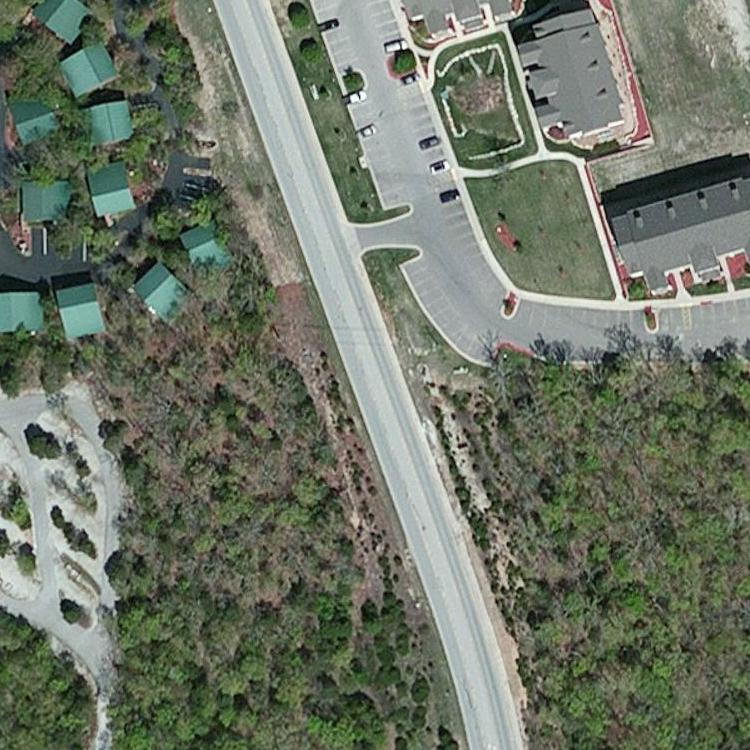}
    \includegraphics[width=0.75\linewidth, height=0.17\linewidth]{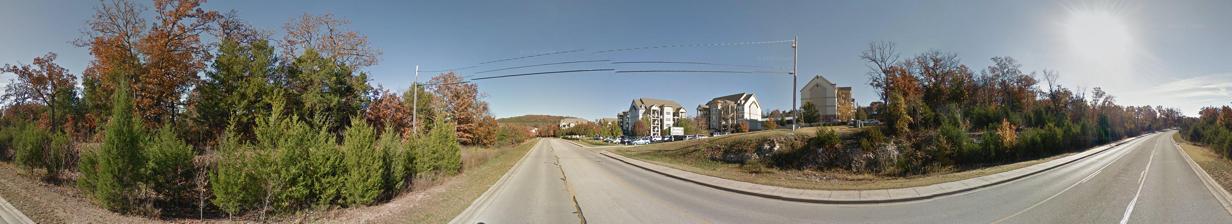}\\
    \smallskip
    \includegraphics[width=0.17\linewidth, height=0.17\linewidth]{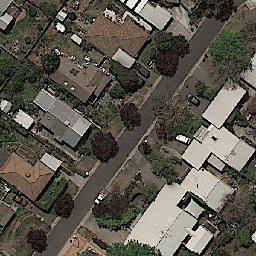}
    \includegraphics[width=0.75\linewidth, height=0.17\linewidth]{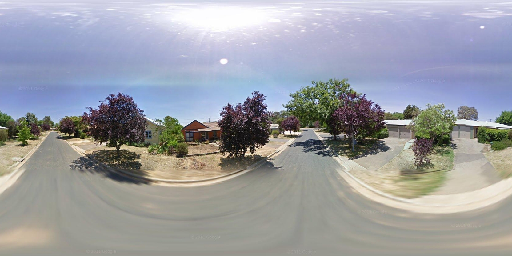}\\
    \smallskip
    \includegraphics[width=0.17\linewidth, height=0.17\linewidth]{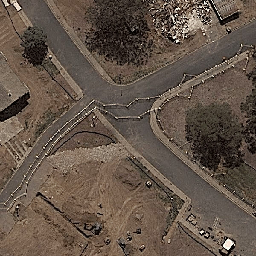}
    \includegraphics[width=0.75\linewidth, height=0.17\linewidth]{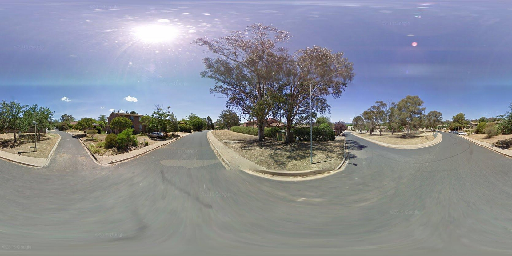}\\
    \caption{\footnotesize Cross-view image pairs from the CVUSA (top two rows) and CVACT (bottom two rows) datasets. The satellite images are on the left and the ground panoramas are on the right.}
    \label{fig:dataset}
\end{figure}

We carry out the experiments on two standard cross-view datasets, CVUSA \cite{zhai2017predicting} and CVACT \cite{Liu_2019_CVPR}. They both contain $35,532$ training ground and satellite pairs and $8,884$ testing pairs. Following an established testing protocol \cite{Liu_2019_CVPR, shi2020optimal}, we denote the test sets in CVUSA and CVACT as CVUSA and CVACT\_val, respectively.
CVACT also provides a larger test set, CVACT\_test, which contains $92,802$ cross-view image pairs for fine-grained city-scale geo-localization.  Note that the ground images in both of the two datasets are panoramas, and all the ground and satellite images are north aligned. 
Figure \ref{fig:dataset} presents samples of cross-view image pairs from the two datasets.

For localizing ground images with unknown orientation and limited ($180^{\circ}$) FoV, we use the image pairs in CVUSA and CVACT\_val and randomly rotate the ground images along the azimuth direction and crop them according to a predetermined FoV. 
The source code of this work is available at \url{https://github.com/shiyujiao/IBL.git}.

\smallskip
\noindent\textbf{Accurate camera geo-localization.}
In the CVUSA dataset, there is no GPS data available for the ground--satellite image pairs. 
The CVACT dataset, introduced in our previous work~\cite{Liu_2019_CVPR}, does provide GPS data for every ground--satellite image pair. 
However, due to GPS drift, the location of a ground camera 
provided by the dataset is not accurate, 
which is visualized in Figure~\ref{fig:accurate_localization}. 
In practice, it is also hard to collect strictly location-aligned satellite--ground image pairs~\cite{lu2020geometry}. 
Hence, we adopt user study for the evaluation of our fine-grained localization method on the CVACT dataset.

\subsection{Coarse camera geo-localization}

\subsubsection{Implementation details}
We use the first ten convolutional layers in VGG16 with pretrained weights on Imagenet \cite{deng2009imagenet} and randomly initialize the parameters in the following three layers for extraction of global feature descriptors. The first seven layers are kept fixed and the subsequent six layers are learnt. The Adam optimizer \cite{kingma2014adam} with a learning rate of $10^{-5}$ is employed for training. Following \cite{vo2016localizing, Hu_2018_CVPR}, we adopt an exhaustive mini-batch strategy \cite{vo2016localizing} with a batch size of $B=32$ to create training triplets. Specifically, for each ground image within a mini-batch, there is one matching satellite image and $B-1$ non-matching images. Thus we construct $B(B-1)$ triplets. Similarly, for each satellite image, there is one matching ground image and $B-1$ non-matching images within a mini-batch, and we create another $B(B-1)$ triplets. In total, we obtain $2B(B-1)$ triplets in total.

In order to obtain a time-efficient approach, we compute the correlation in our DSM module by the fast Fourier transform during inference. Specifically, we store Fourier coefficients of satellite features in the database, and only calculate Fourier coefficients of the ground descriptors in the forward pass. 
In doing so, the computation cost yields $13NHW_s C$ flops (including $4NHW_sC$ flops for coefficient multiplication in the spectral domain, and $1.5NHCW_s\log_2 W$ flops for the inverse Fast Fourier Transform), where $H$, $W_s$ and $C$ are the height, width and channel number of the global feature descriptor of a satellite image, and $N$ is the number of database satellite images.  
In contrast, performing correlation in the spatial domain requires $2NHW^2C$ flops. Thus, the computation time by applying the Fourier transform is reduced by a factor of $10$ ($\frac{13NHWC}{2NHW^2C}\approx \frac{1}{10}$).

{\color{black}
For localizing query images with unknown orientation and a limited FoV, there is a shift-and-crop operation to find the corresponding part of a query image in a satellite image. 
The flops in the shift operation are $NHW_s C$, and the flops in the crop operation are $NHW_g C$, where $N$ is the number of database satellite images, $H$ and $C$ are the height and channel number of satellite (and query ground) features, $W_g$ and $W_s$ are the widths of satellite and query ground features, respectively, and $W_g = \text{FoV} * W_s/360$.  
The flops in computing the similarity between a query image and all satellite images are $3NHW_g C$. 
Hence, the complexity of retrieving a ground image from a database with $N$ satellite images is $O(NHW_sC)$.}

{\color{black}
Using an RTX 2080 Ti GPU, the feature extraction time for a ground image is $0.01$s, and it takes $0.06$s on average to retrieve its satellite counterpart from a database containing 8884 reference images.
On the CVACT\_val dataset, the $8884$ reference images cover approximately $64\text{km}^2$.
}

\subsubsection{Evaluation Metrics}
\smallskip
\noindent\textbf{Location estimation.}
Following the standard evaluation procedure for cross-view image localization \cite{vo2016localizing, Hu_2018_CVPR, Liu_2019_CVPR, shi2020optimal, Cai_2019_ICCV, sun2019geocapsnet, Regmi_2019_ICCV}, we use the top $K$ recall as the location evaluation metric to examine the performance of our method and compare it with the state-of-the-art. 
Specifically, given a ground image, we retrieve the top $K$ satellite images in terms of $\ell_2$ distance between their global descriptors. The ground image is regarded as successfully localized if its corresponding satellite image is retrieved within the top $K$ list. The percentage of correctly localized ground images is recorded as recall at $K$ (r@$K$).

\smallskip
\noindent\textbf{Orientation estimation.} 
The predicted orientation of a query ground image is meaningful only when the ground image is localized correctly. 
Hence, we evaluate the orientation estimation accuracy of our DSM only on ground images that have been correctly localized by the top-1 recall. 
In this experiment, when the differences between the predicted orientation of a ground image and its ground-truth orientation is within $\pm 10\%$ of its FoV, the orientation estimation of this ground image is deemed as a success. We record the percentage of ground images of which the orientation is correctly predicted as the orientation estimation accuracy (Orien\_acc).

\smallskip
\noindent \textbf{Combined measure.}
Denote the top-1 recall rate for location estimation as Loc\_acc. The overall 3-DOF camera localization performance is computed as $\text{Loc\_acc}\times \text{Orien\_acc}$, which we denote as Overall.

\subsubsection{Localizing Orientation-aligned Panoramas}
\label{exp:orien-align}

\begin{figure*}[ht!]
\setlength{\abovecaptionskip}{0pt}
\setlength{\belowcaptionskip}{0pt}
    \centering
    \subfloat[ CVUSA]{
    \centering
    \scalebox{1.0}[1.0]{\includegraphics[trim={8mm 65mm 10mm 70mm}, clip, width=0.27\linewidth]{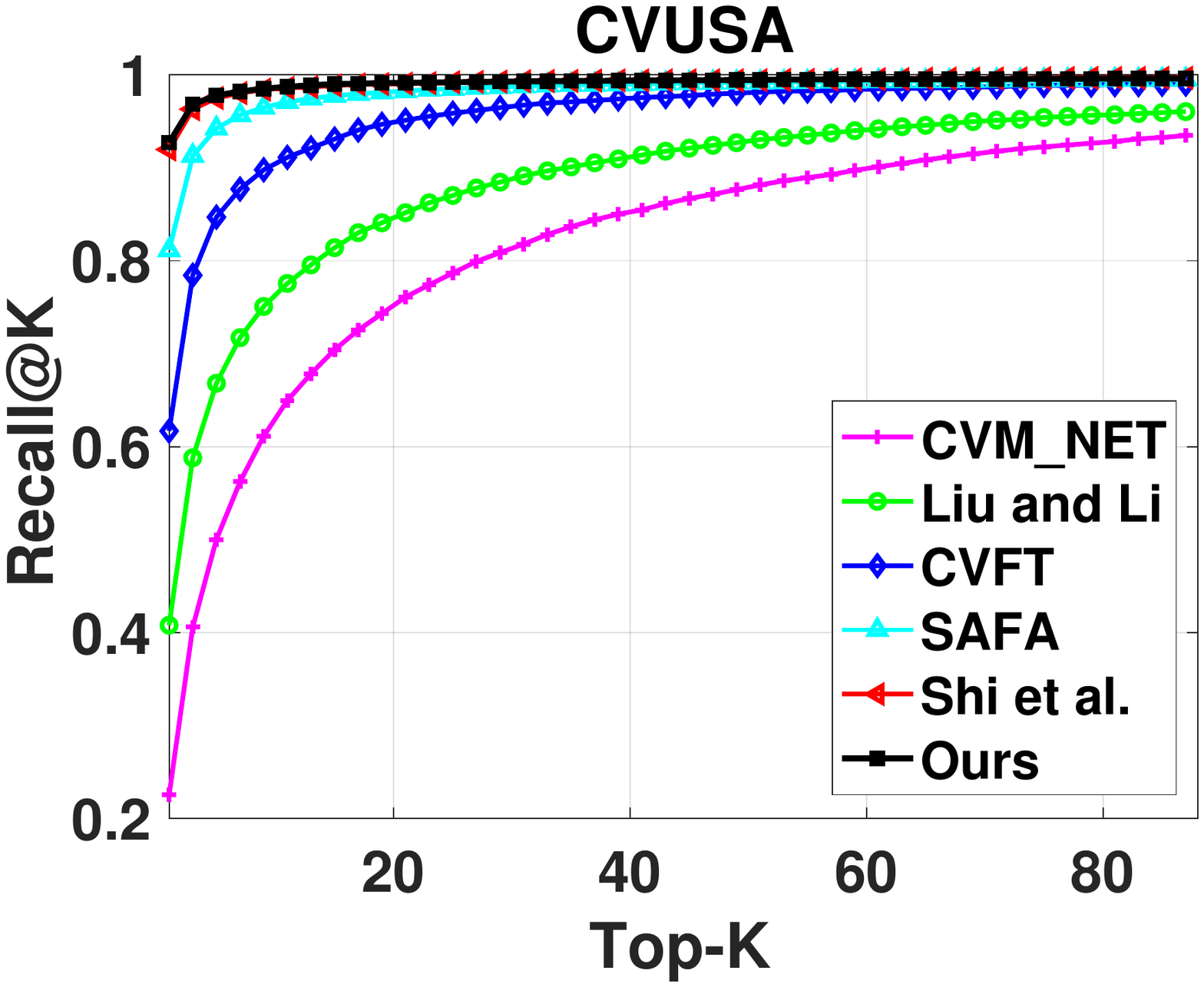}}
    \label{fig: recall_CVUSA}
    }
    \hspace{-2mm}
    \subfloat[ CVACT\_val]{
    \centering
     \scalebox{1.0}[1.0]{\includegraphics[trim={8mm 65mm 10mm 70mm}, clip, width=0.27\linewidth]{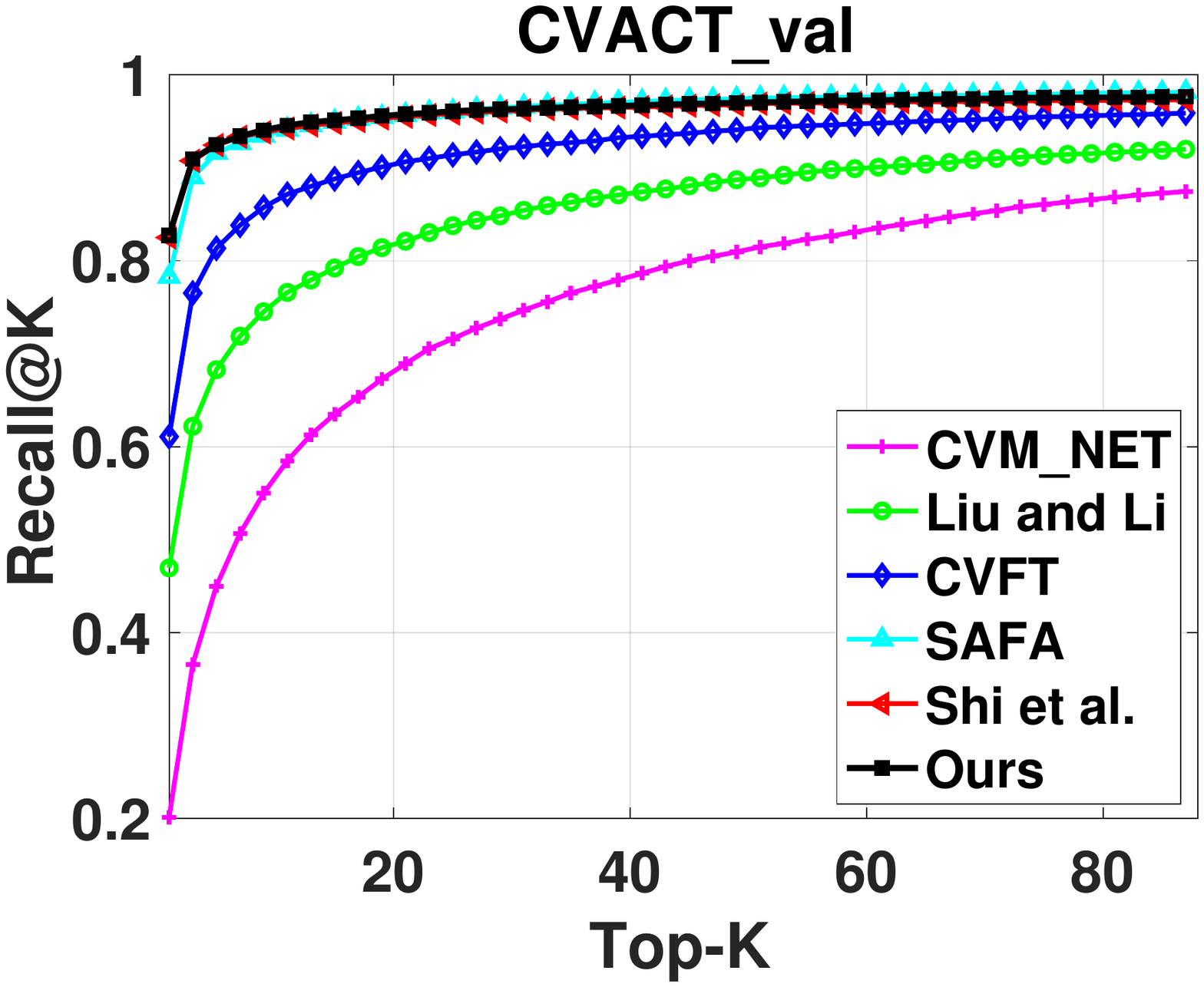}}
    \label{fig: recall_CVACT_val}
    }
     \hspace{-2mm}
    \subfloat[ CVACT\_test]{
    \centering
    \scalebox{1.0}[1.0]{\includegraphics[trim={8mm 65mm 10mm 70mm}, clip, width=0.27\linewidth]{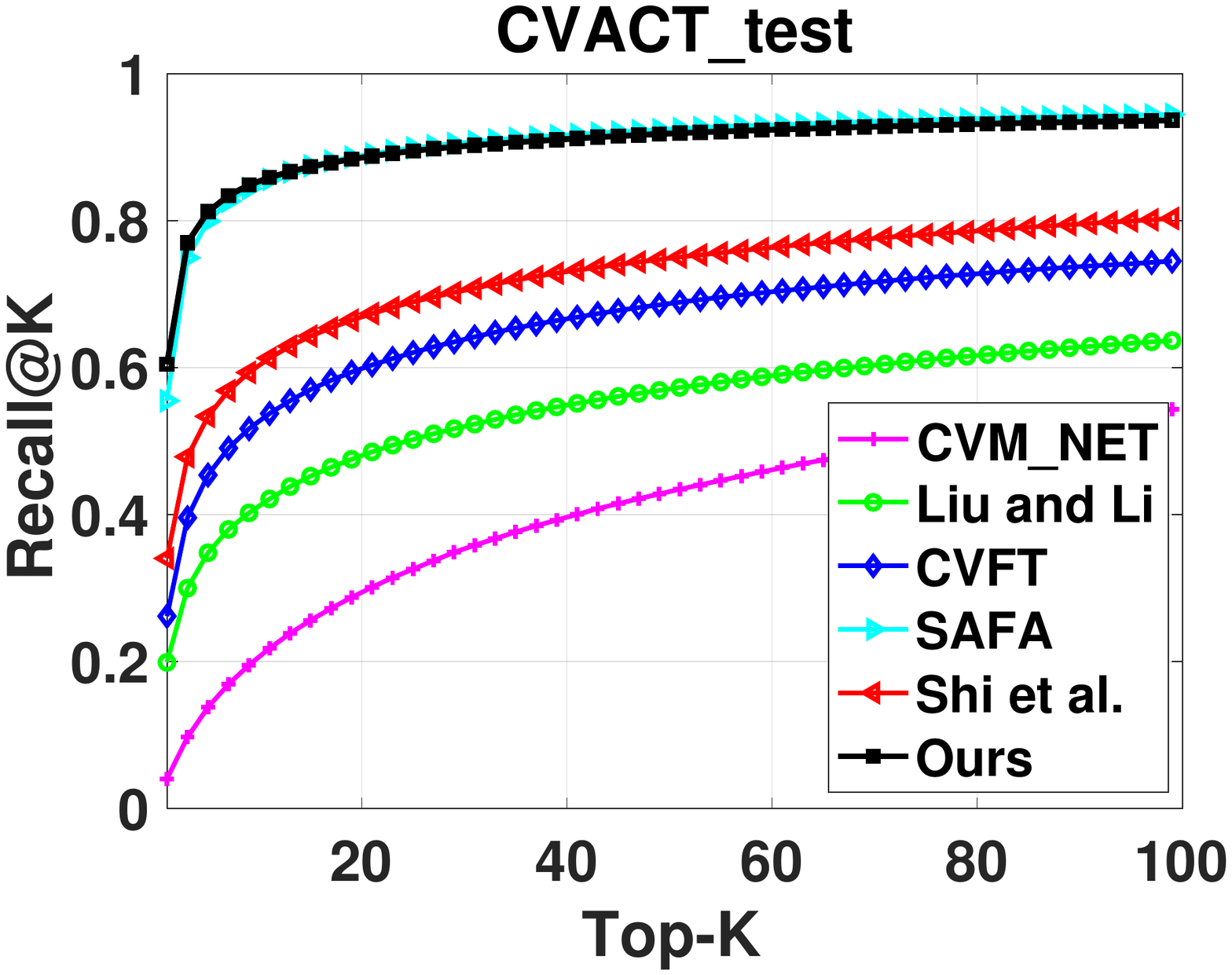}}
    \label{fig: recall_CVACT_test}
    }
    \caption{\footnotesize Evaluations of recall at different values of $K$ on the CVUSA, CVACT\_val and CVACT\_test datasets.}
    \label{fig:recall_CVUSA_CVACT}
\end{figure*}

\begin{table}[tbp]
\setlength{\abovecaptionskip}{0pt}
\setlength{\belowcaptionskip}{0pt}
% \footnotesize
% \setlength{\tabcolsep}{2pt}
\footnotesize
\centering
\caption{\footnotesize Comparison of our approach with existing methods on the CVUSA \cite{zhai2017predicting} dataset. 
% Here, ``--'' denotes that the results on the corresponding evaluation metric are not available as some of the works only use r@1\% as the evaluation metric. 
}
\begin{tabular}{c|c c c c}
\toprule
 \multirow{2}{*}{Methods} & \multicolumn{4}{c}{CVUSA}      \\ \cline{2-5} 
                                      & r@1   & r@5   & r@10  & r@1\%   \\\hline \hline
 Workman \etal \cite{workman2015wide}  & --     & --     & --     & 34.3 \\
 Zhai \etal \cite{zhai2017predicting}  & --     & --     & --     & 43.2   \\  
 Vo and Hays \cite{vo2016localizing}   & --     & --     & --    & 63.7   \\  
 CVM-NET \cite{Hu_2018_CVPR}        & 22.47 & 49.98 & 63.18 & 93.62  \\  
 Liu and Li \cite{Liu_2019_CVPR}        & 40.79 & 66.82 & 76.36 & 96.12        \\ 
 Regmi and Shah \cite{Regmi_2019_ICCV}  & 48.76 & 73.64  & 81.27 & 95.94        \\ 
 GeocapsNet-II~\cite{Cai_2019_ICCV}        & --    & --     & --     & 98.07        \\ 
 Siam-FCANet34 \cite{Cai_2019_ICCV}          & --     & --     & --     & 98.3        \\ 
 CVFT \cite{shi2020optimal}            & 61.43 & 84.69 & 90.49 & 99.02       \\ 
 SAFA \cite{shi2019spatial}            & 89.84 & 96.93 & 98.14 & 99.64 \\
 Shi~\etal\cite{shi2020looking}     & {91.96} & {97.50} & {98.54} &\textbf{99.67}   \\
 Ours                  & \textbf{92.69} & \textbf{97.78} & \textbf{98.60} & {99.61}   \\
\bottomrule
\end{tabular}
\label{tab: compare_stoa_CVUSA}
\end{table}

\begin{table}[tbp]
\setlength{\abovecaptionskip}{0pt}
\setlength{\belowcaptionskip}{0pt}
\footnotesize
% \footnotesize
% \setlength{\tabcolsep}{8pt}
\centering
\caption{\footnotesize Comparison of our approach with existing methods on the CVACT\_val \cite{Liu_2019_CVPR} dataset by re-training existing networks. %using the code provided by the authors.
}
\begin{tabular}{c | c c c c}
\toprule
 \multirow{2}{*}{Methods} & \multicolumn{4}{c}{CVACT\_val} \\ \cline{2-5} 
                                       & r@1   & r@5   & r@10  & r@1\% \\\hline \hline
 CVM-NET  \cite{Hu_2018_CVPR}          & 20.15 & 45.00 & 56.87 & 87.57    \\  
 Liu and Li \cite{Liu_2019_CVPR}        & 46.96 & 68.28 & 75.48 & 92.01      \\ 
 Regmi and Shah \cite{Regmi_2019_ICCV}  & 48.62 & 72.48  & 79.65 & 93.16        \\ 
 CVFT \cite{shi2020optimal}            & 61.05 & 81.33 & 86.52 & 95.93      \\
 SAFA \cite{shi2019spatial}            & 81.03 & \textbf{92.80} & \textbf{94.84} & \textbf{98.17}  \\
 Shi~\etal~\cite{shi2020looking}                    & {82.49}  & {92.44}     & {93.99}     & {97.32}       \\ 
 Ours                     & \textbf{82.70}  & {92.50}     & {94.24}     & {97.65}       \\ 
\bottomrule
\end{tabular}
\label{tab: compare_stoa_CVACT}
\end{table}

Firstly, we investigate the performance of location estimation of our method and compare it with the state-of-the-art~\cite{workman2015wide, zhai2017predicting, vo2016localizing, Hu_2018_CVPR, Liu_2019_CVPR, Regmi_2019_ICCV, Cai_2019_ICCV, shi2020optimal, shi2019spatial, shi2020looking} on the standard CVUSA and CVACT datasets, where ground images are orientation-aligned panoramas. 
The recall results  at top-1, top-5, top-10 and top-1\% on the CVUSA and CVACT\_val datasets are presented in Table \ref{tab: compare_stoa_CVUSA} and Table \ref{tab: compare_stoa_CVACT}, respectively. 
They are reported from other works or produced by the re-trained models using source codes provided by the authors. 
The complete r@$K$ performance curves on CVUSA and CVACT\_val are illustrated in Figure \ref{fig: recall_CVUSA} and Figure \ref{fig: recall_CVACT_val}, respectively.

Among those baseline methods, \cite{workman2015wide,zhai2017predicting, vo2016localizing} are the first approaches that utilize deep learning for cross-view related tasks. 
CVM-NET \cite{Hu_2018_CVPR}, GeocaosNet-II~\cite{Cai_2019_ICCV} and Siam-FCANet34 \cite{Cai_2019_ICCV} focus on designing powerful feature extraction networks. 
Liu and Li \cite{Liu_2019_CVPR} introduce the orientation information to networks so as to facilitate geo-localization performance. 
These works do not explicitly address the domain gap between ground and satellite images which leads to their inferior performance. 

Regmi and Shah \cite{Regmi_2019_ICCV} adopt a conditional GAN to generate satellite images from ground panoramas. Although it helps to bridge the cross-view domain gap, undesired scene contents are also introduced in this process. 
CVFT \cite{shi2020optimal} proposes a Cross-view Feature Transport (CVFT) module to better align ground and satellite features. However, it is hard for networks to learn geometric and feature response correspondences simultaneously. 
SAFA \cite{shi2019spatial} explores a parameter-free polar transform to bridge the geometric domain gap, and proposes a Spatially-aware Position Embedding (SPE) module to further construct geometric and feature correspondences between the two view images. 
In~\cite{shi2020looking}, we found that further increasing the spatial size of the global image descriptors can increase the localization performance. 
Similar to SAFA, the polar transform is also employed in~\cite{shi2020looking} to bridge the cross-view domain gap. 
However, in this journal paper, we further establish the geometric correspondences between a satellite image and its corresponding ground-level panorama by a projective transform. 
The projective transform and the polar transform cooperate with each other to align the matching ground and satellite image pairs.
As shown in Table \ref{tab: compare_stoa_CVUSA} and Table \ref{tab: compare_stoa_CVACT}, the top-1 recall rate has been further boosted by the newly proposed method compared to our conferece work~\cite{shi2020looking}. 
Although SAFA achieves slightly better performance on the CVACT\_val, its performance degrades significantly when orientation of query images is unknown, which will be investigated in later experiments.

\begin{figure*}[!t] 
\setlength{\abovecaptionskip}{0pt}
\setlength{\belowcaptionskip}{0pt}
\centering
\begin{minipage}{0.7\linewidth}
\begin{minipage}{0.27\linewidth}
\centering
\includegraphics[width=\textwidth]{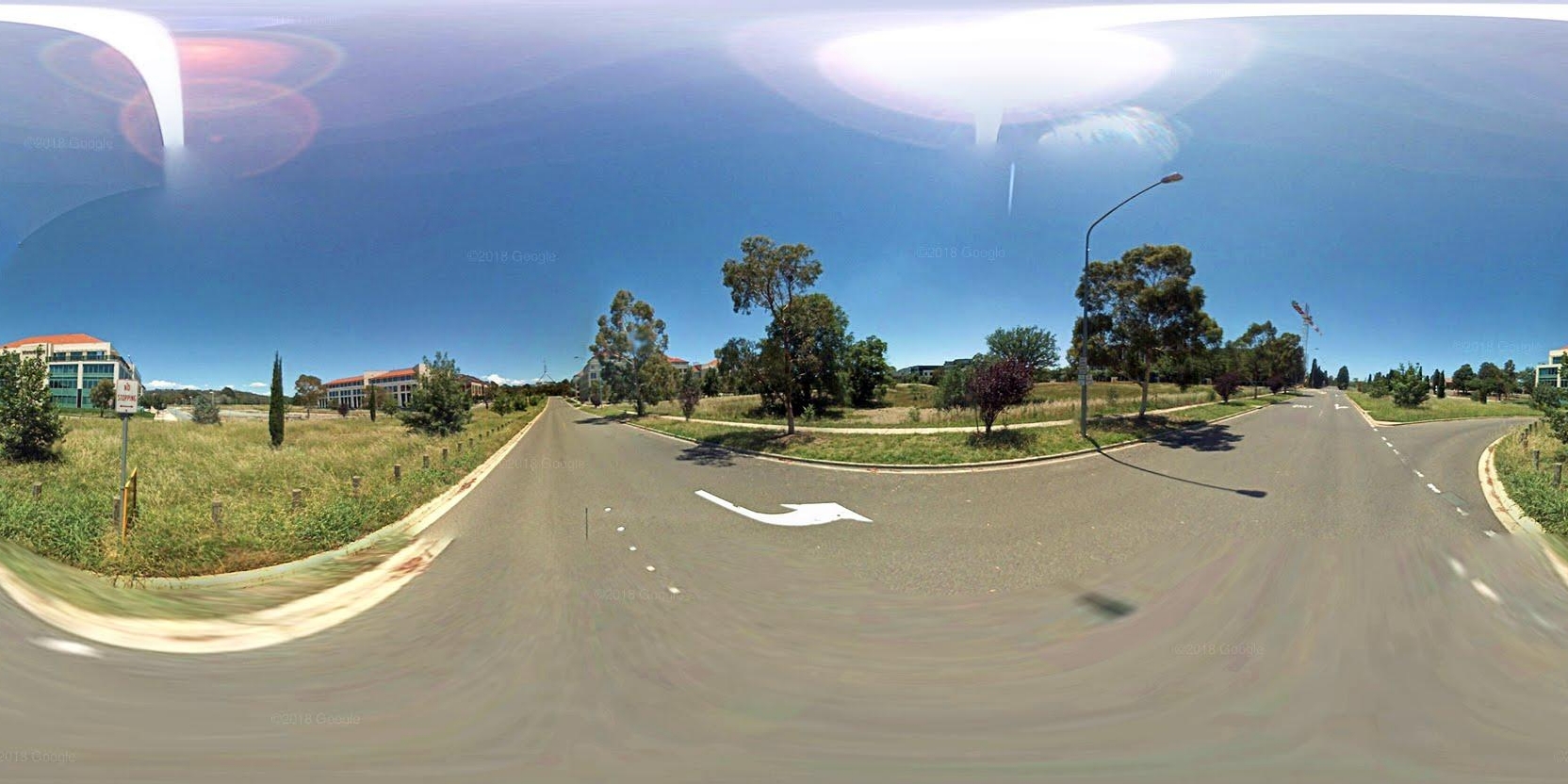}
\end{minipage}
% \hfill
\begin{minipage}{0.135\linewidth}
\centering
\includegraphics[width=\textwidth]{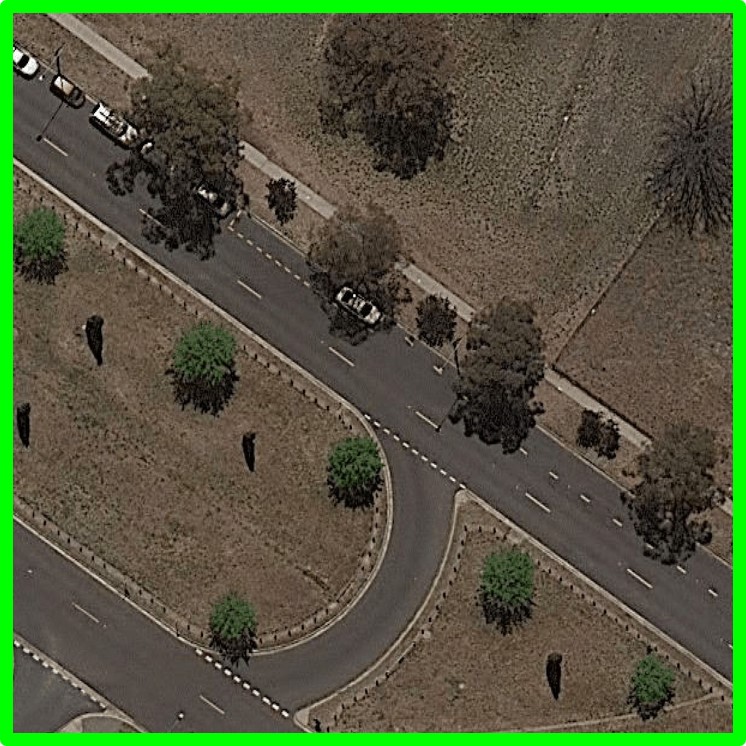}
\end{minipage}
% \hfill
\begin{minipage}{0.135\linewidth}
\centering
\includegraphics[width=\textwidth]{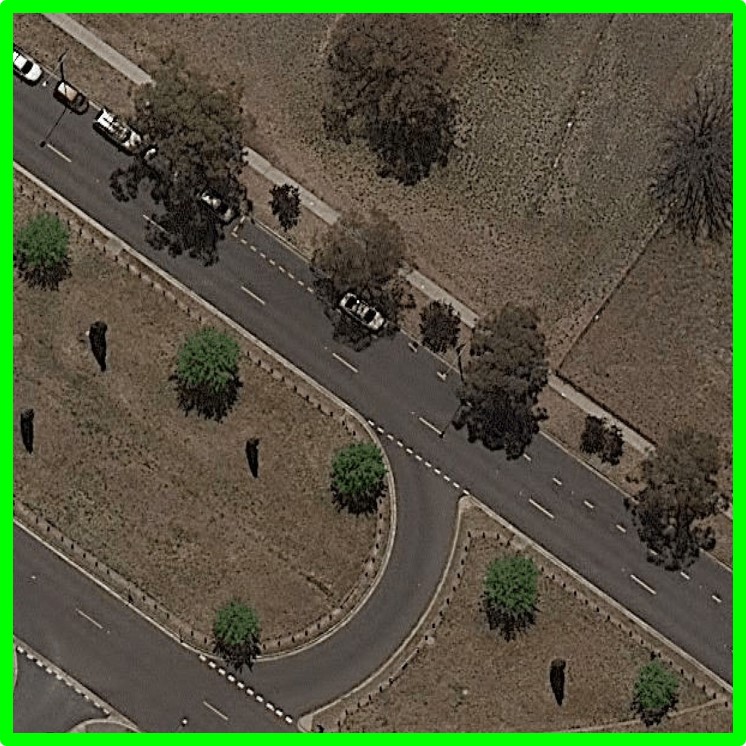}
\end{minipage}
% \hfill
\begin{minipage}{0.135\linewidth}
\centering
\includegraphics[width=\textwidth]{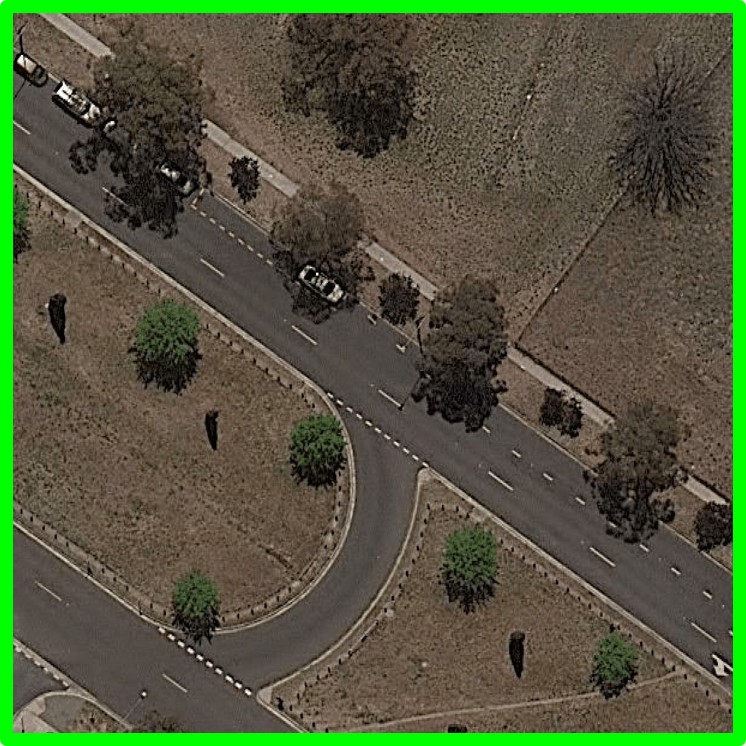}
\end{minipage}
% \hfill
\begin{minipage}{0.135\linewidth}
\centering
\includegraphics[width=\textwidth]{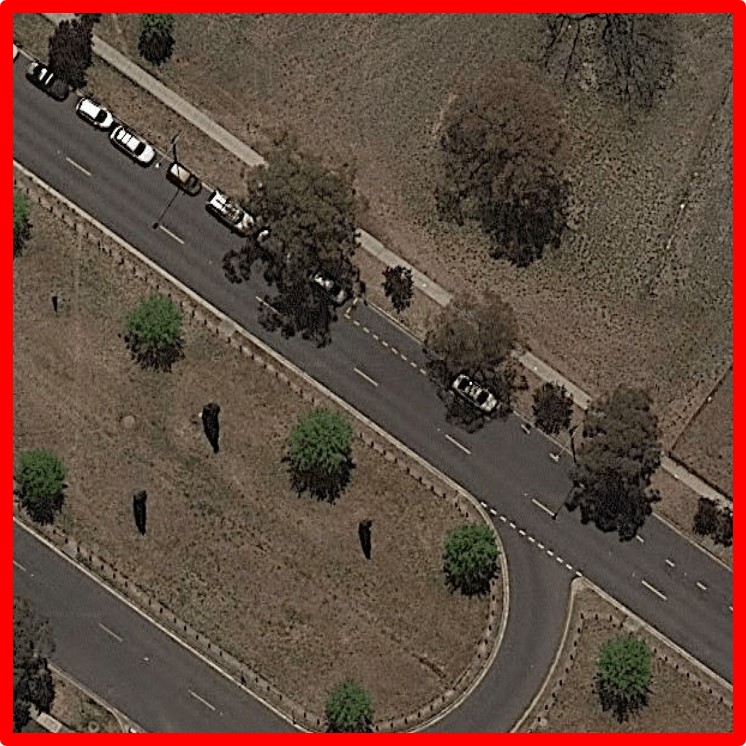}
\end{minipage}
% \hfill
\begin{minipage}{0.135\linewidth}
\centering
\includegraphics[width=\textwidth]{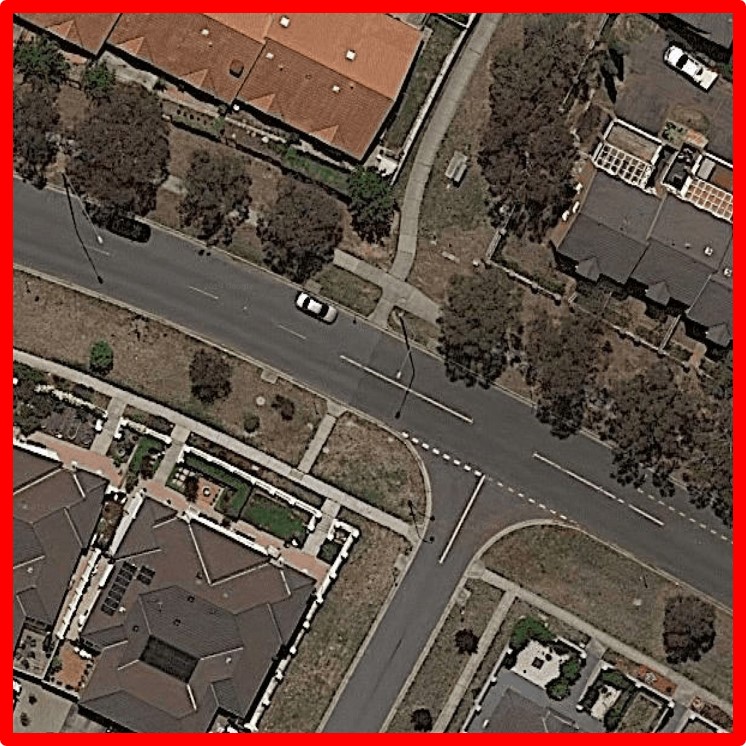}
\end{minipage}
% \hfill
\\\smallskip
\begin{minipage}{0.27\linewidth}
\centering
\includegraphics[width=\textwidth]{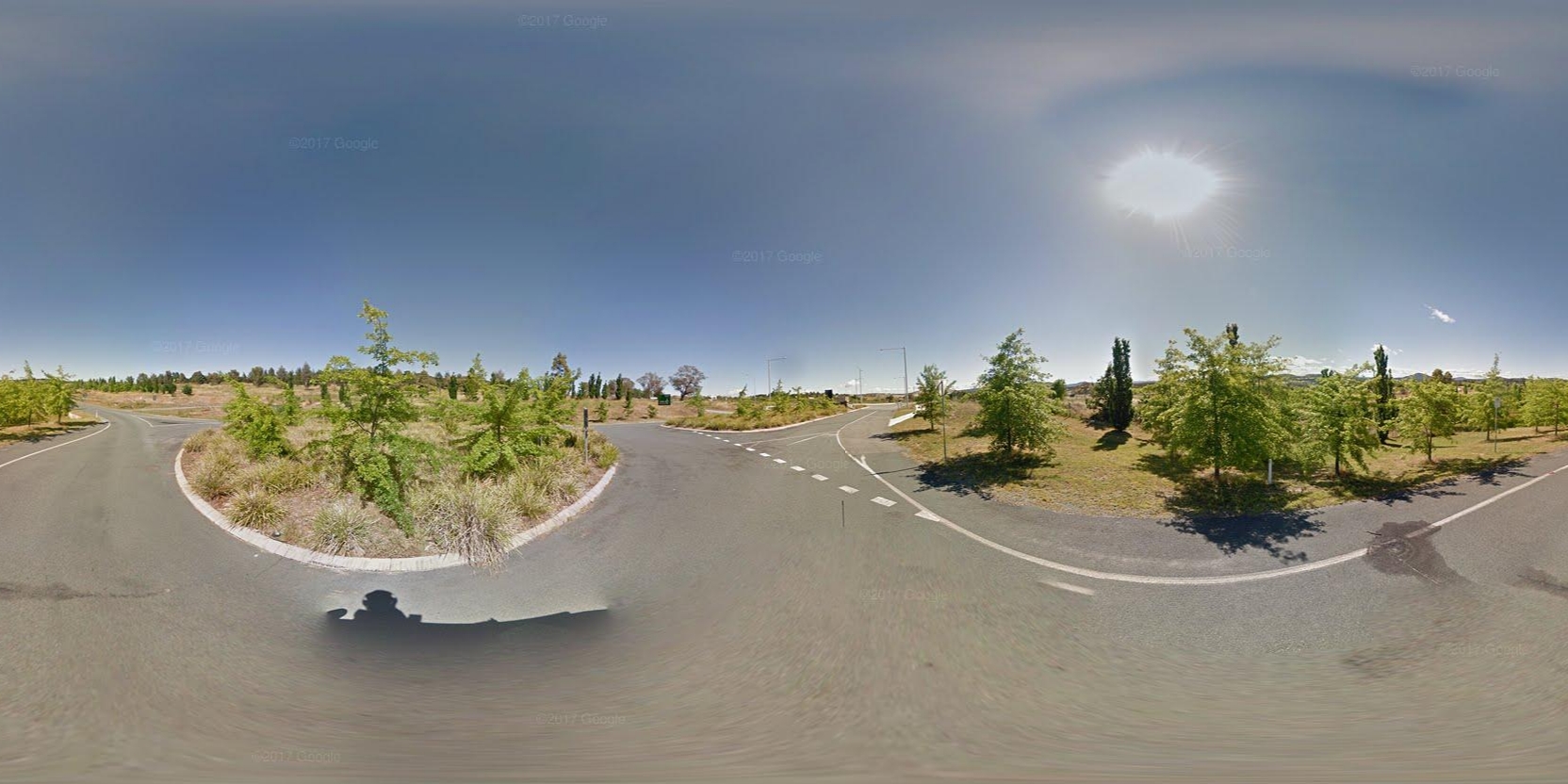}
\end{minipage}
% \hfill
% \hspace{0 mm}
\begin{minipage}{0.135\linewidth}
\centering
\includegraphics[width=\textwidth]{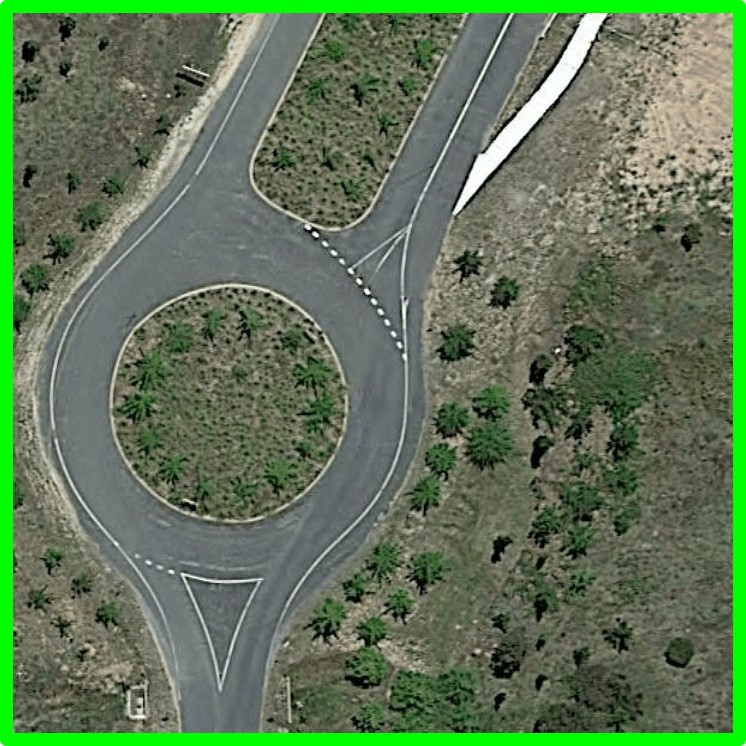}
\end{minipage}
% \hfill
\begin{minipage}{0.135\linewidth}
\centering
\includegraphics[width=\textwidth]{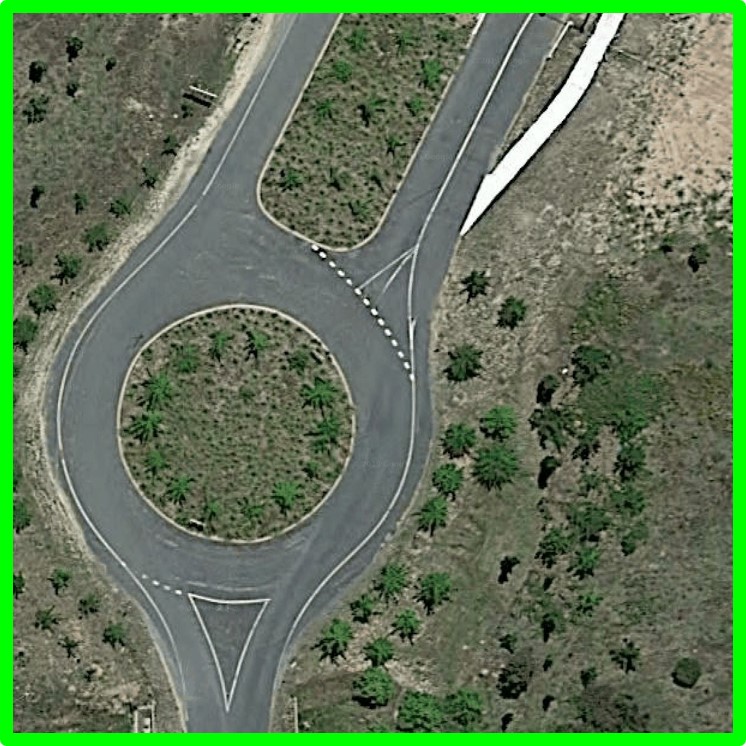}
\end{minipage}
% \hfill
\begin{minipage}{0.135\linewidth}
\centering
\includegraphics[width=\textwidth]{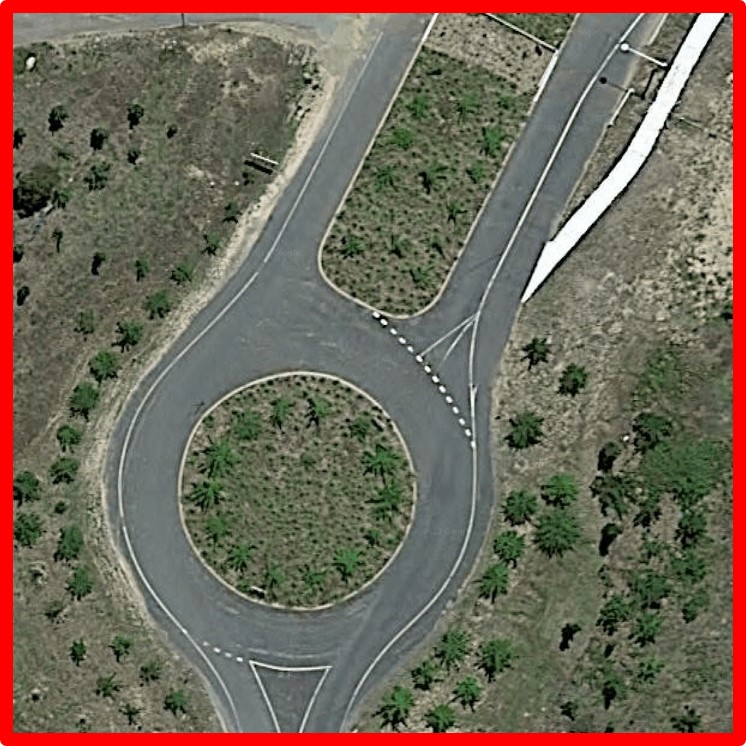}
\end{minipage}
% \hfill
\begin{minipage}{0.135\linewidth}
\centering
\includegraphics[width=\textwidth]{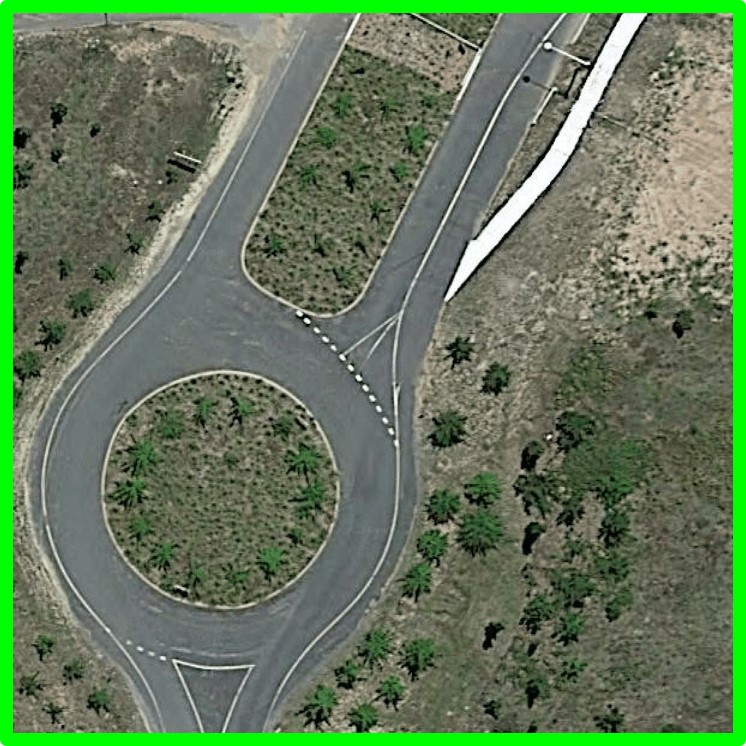}
\end{minipage}
% \hfill
\begin{minipage}{0.135\linewidth}
\centering
\includegraphics[width=\textwidth]{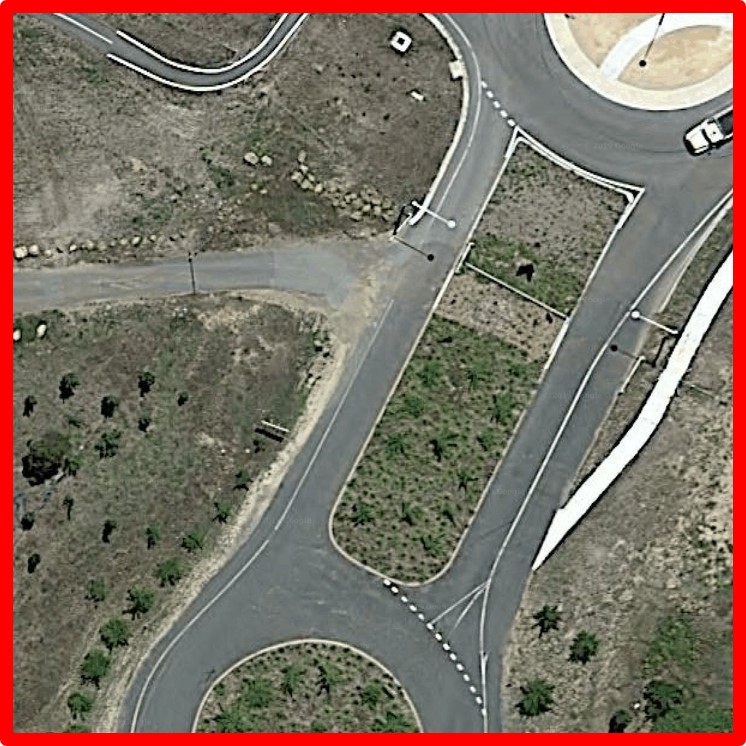}
\end{minipage}
% \hfill
\\\smallskip
\begin{minipage}{0.27\linewidth}
\centering
\includegraphics[width=\textwidth]{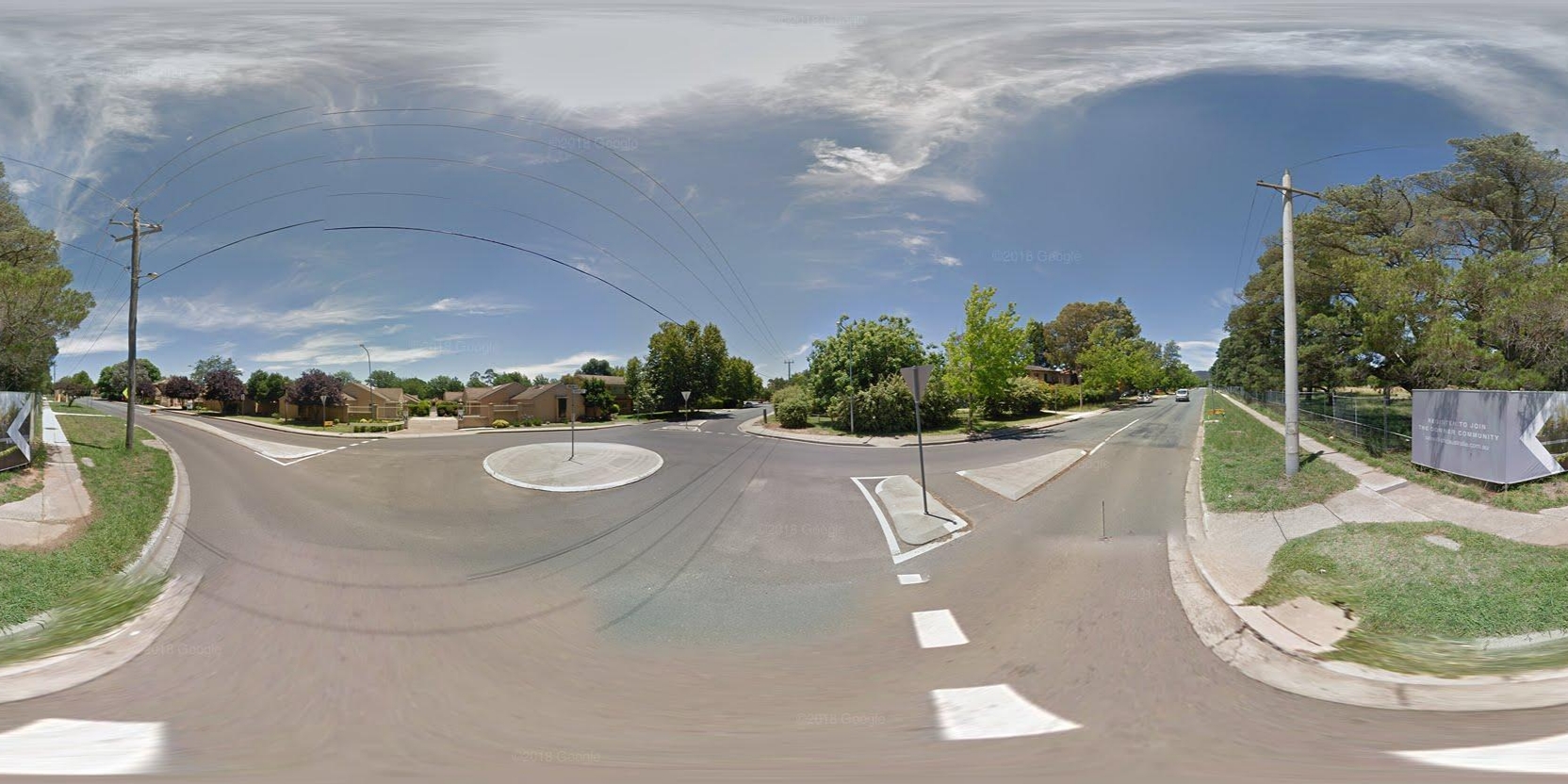}
\end{minipage}
% \hfill
\begin{minipage}{0.135\linewidth}
\centering
\includegraphics[width=\textwidth]{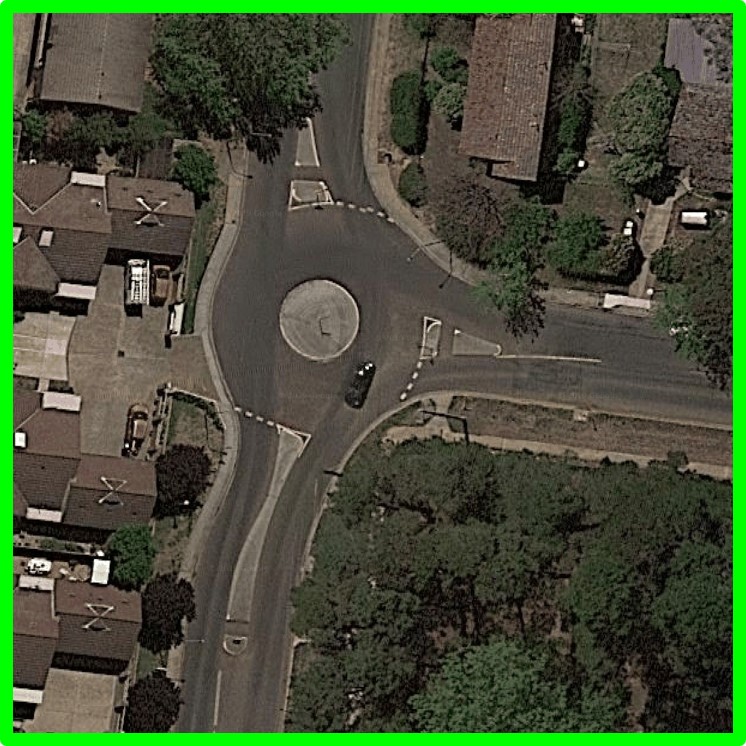}
\end{minipage}
% \hfill
\begin{minipage}{0.135\linewidth}
\centering
\includegraphics[width=\textwidth]{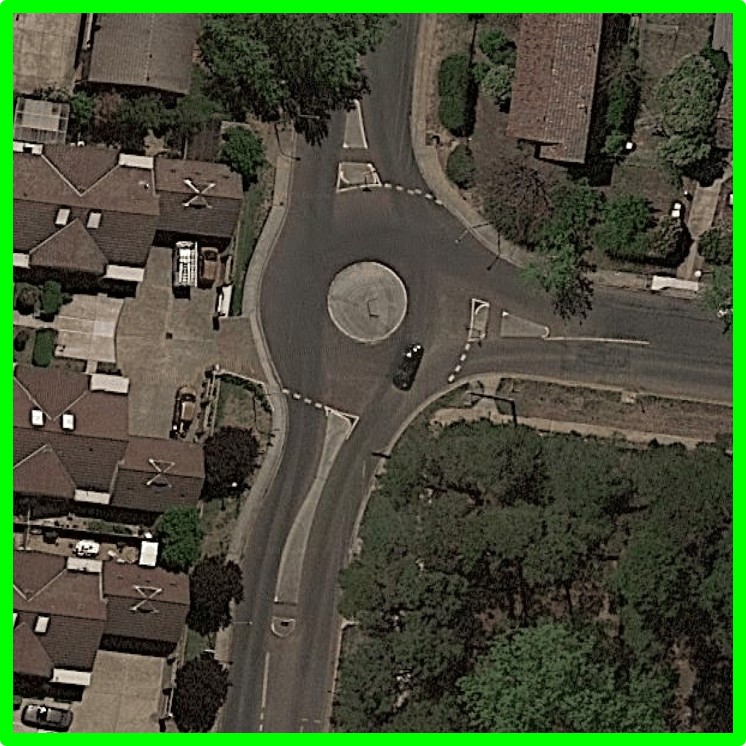}
\end{minipage}
% \hfill
\begin{minipage}{0.135\linewidth}
\centering
\includegraphics[width=\textwidth]{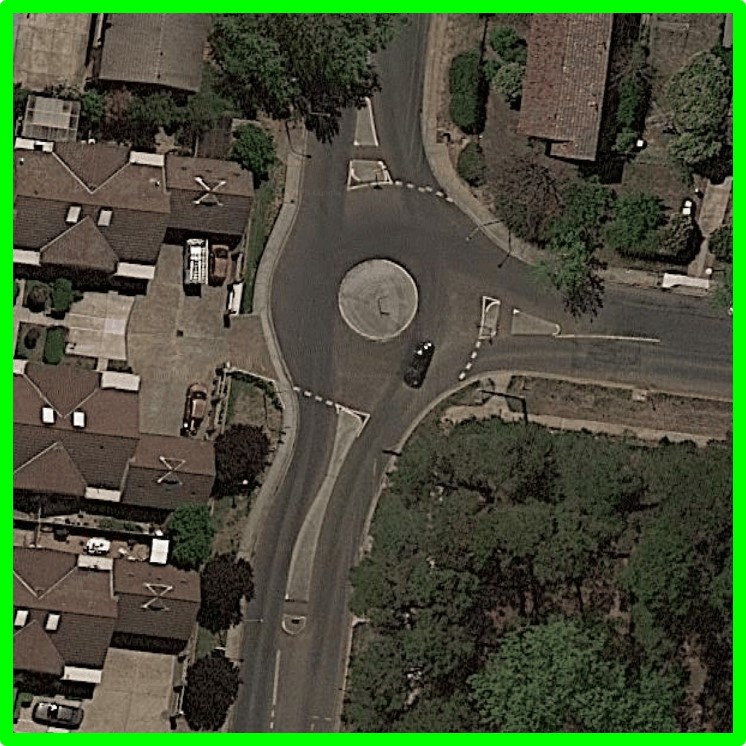}
\end{minipage}
% \hfill
\begin{minipage}{0.135\linewidth}
\centering
\includegraphics[width=\textwidth]{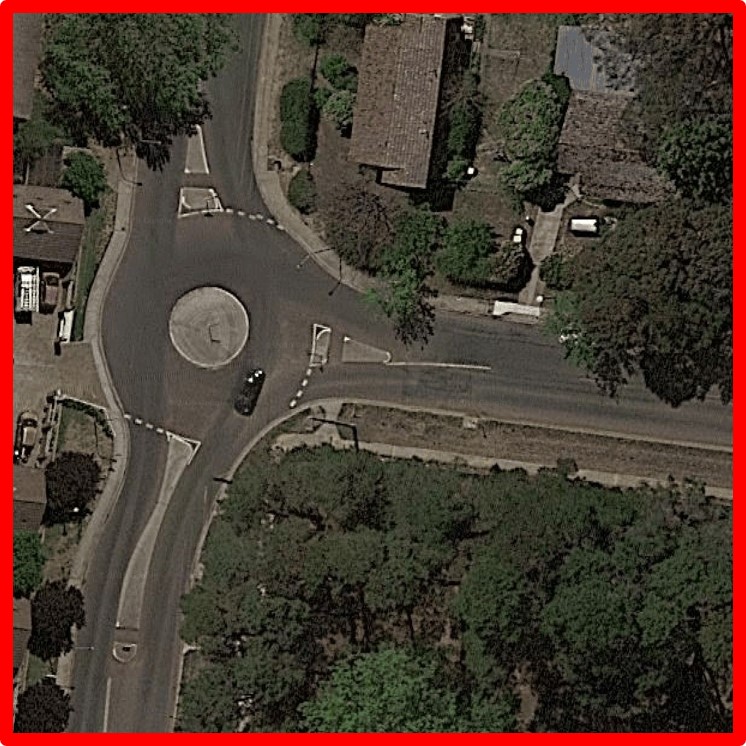}
\end{minipage}
% \hfill
\begin{minipage}{0.135\linewidth}
\centering
\includegraphics[width=\textwidth]{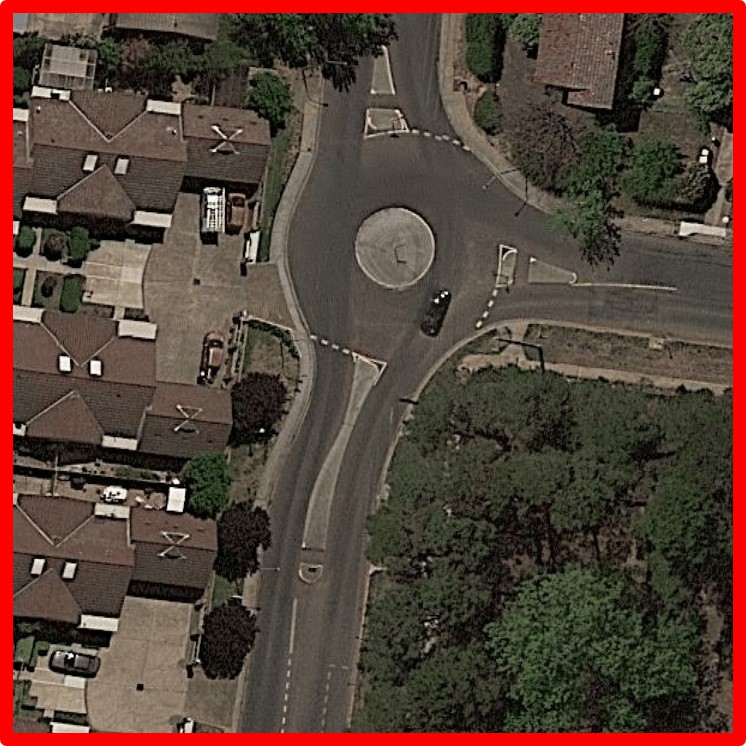}
\end{minipage}
% \hfill
\\\smallskip
\begin{minipage}{0.27\linewidth}
\centering
\includegraphics[width=\textwidth]{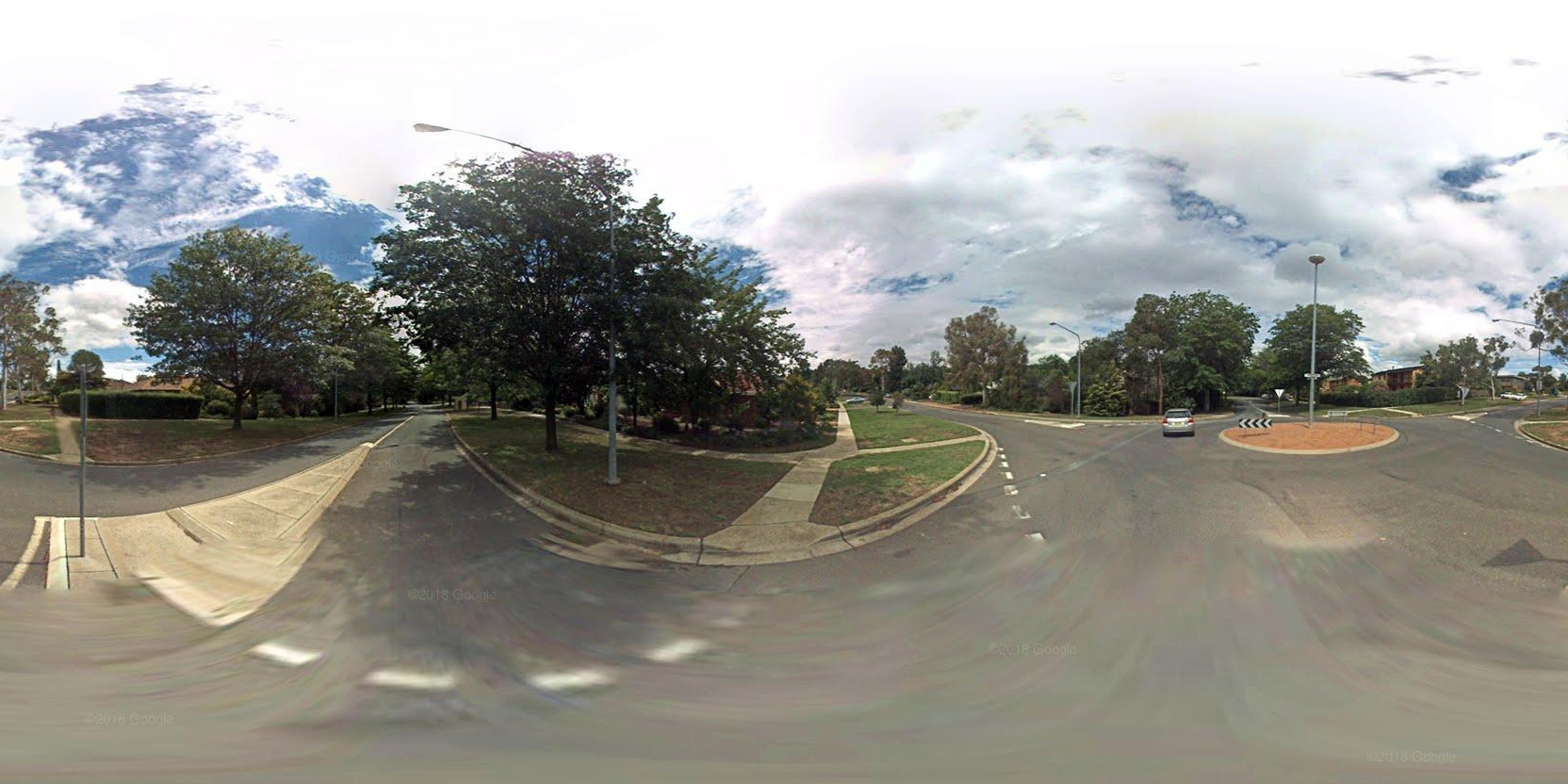}
\end{minipage}
% \hfill
\begin{minipage}{0.135\linewidth}
\centering
\includegraphics[width=\textwidth]{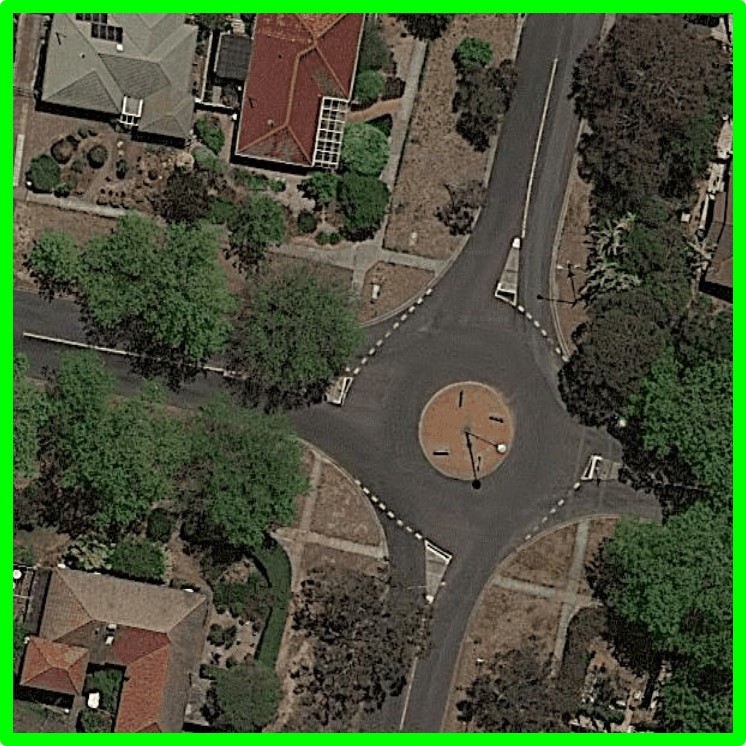}
\end{minipage}
% \hfill
\begin{minipage}{0.135\linewidth}
\centering
\includegraphics[width=\textwidth]{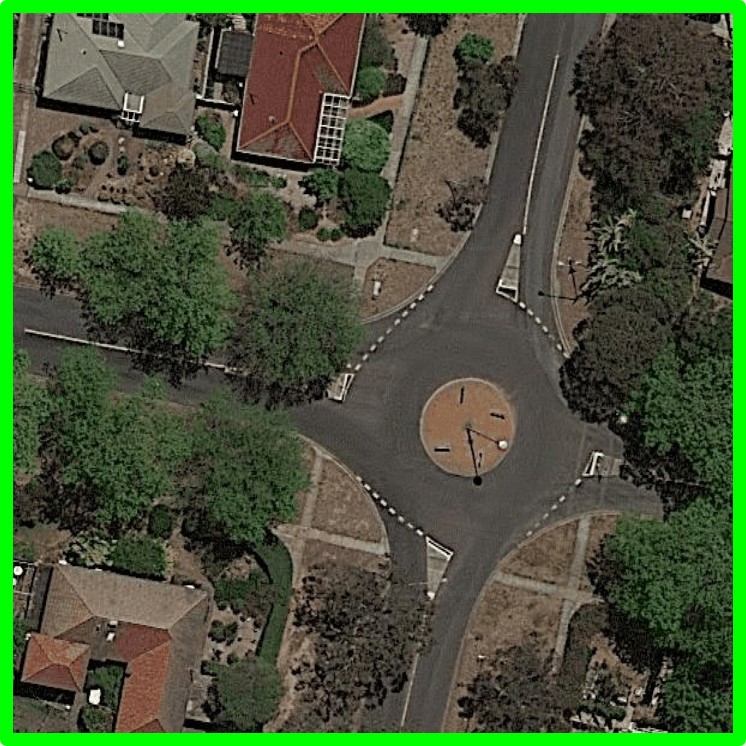}
\end{minipage}
% \hfill
\begin{minipage}{0.135\linewidth}
\centering
\includegraphics[width=\textwidth]{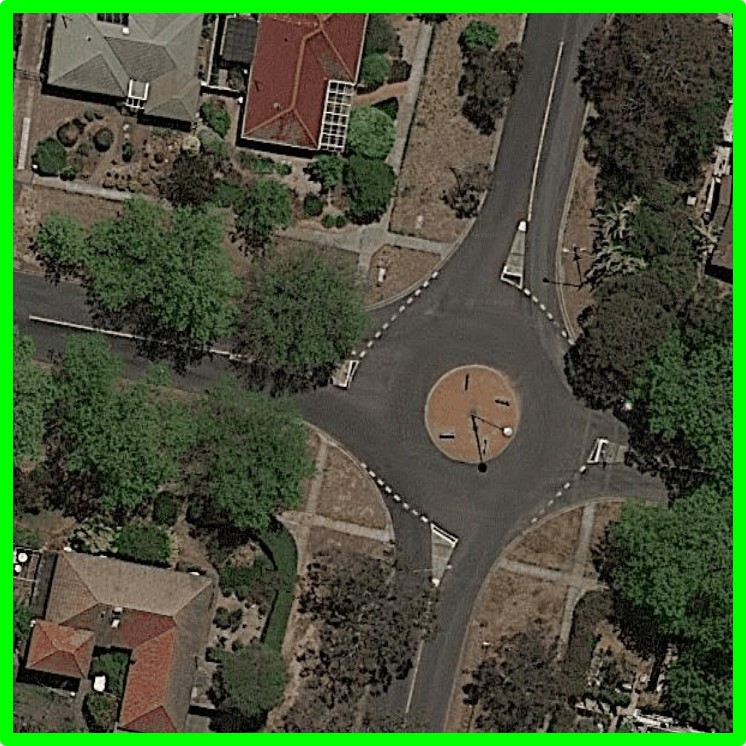}
\end{minipage}
% \hfill
\begin{minipage}{0.135\linewidth}
\centering
\includegraphics[width=\textwidth]{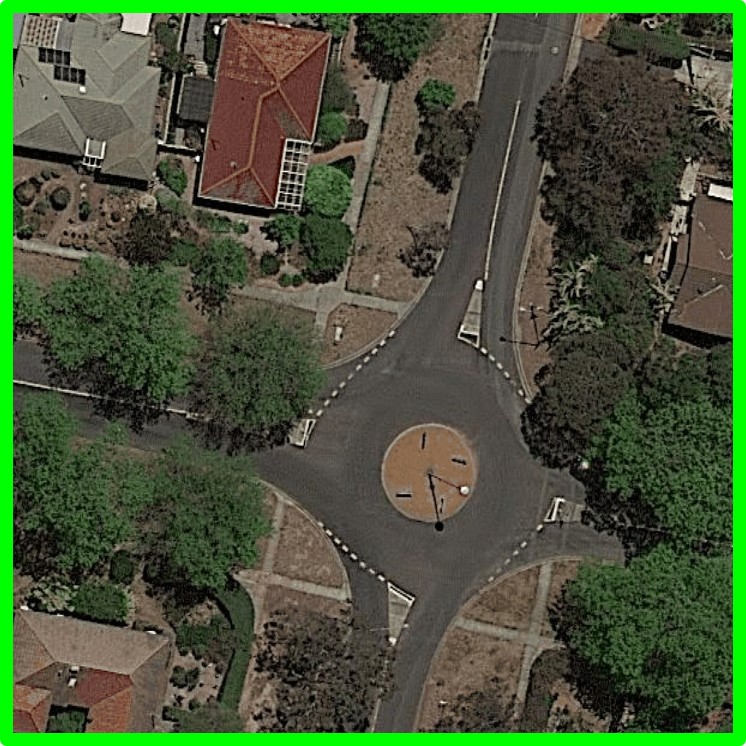}
\end{minipage}
% \hfill
\begin{minipage}{0.135\linewidth}
\centering
\includegraphics[width=\textwidth]{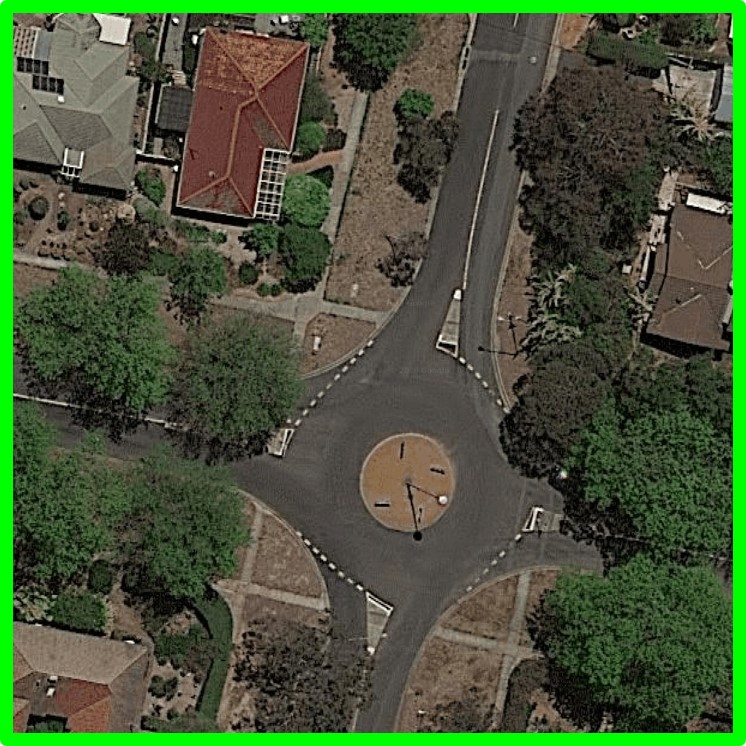}
\end{minipage}
\end{minipage}
% \hfill
\\\smallskip
\centering
\caption{\footnotesize{
Visualization of localization results attained by our method on the CVACT\_test set. From left to right: ground-level query image and the top 1-5 retrieved satellite candidates. Green and red borders indicate correctly and incorrectly retrieved results, respectively.
}}
\label{fig:ACT_test_local_examp}
\end{figure*}

\begin{table*}[!ht]
\setlength{\abovecaptionskip}{0pt}
\setlength{\belowcaptionskip}{0pt}
\footnotesize
\setlength{\tabcolsep}{3pt}
\centering
\caption{\footnotesize Comparison of recall rates for localizing ground images with unknown orientations and varying FoVs (models trained with random orientation augmentation).}
\begin{tabular}{c|c|c c c c| c c c c|c c c c |c c c c}
\toprule
 \multirow{3}{*}{Dataset} & \multirow{3}{*}{Methods} & \multicolumn{8}{c|}{FoV=$360^{\circ}$}     & \multicolumn{8}{c}{FoV=$180^{\circ}$}     \\ \cline{3-18}
 & & \multicolumn{4}{c|}{Orientation Aligned}     & \multicolumn{4}{c|}{Orientation Unknown} &\multicolumn{4}{c|}{Orientation Aligned}     & \multicolumn{4}{c}{Orientation Unknown} \\
                      &                                       & r@1   & r@5   & r@10  & r@1\%  & r@1   & r@5   & r@10  & r@1\% & r@1   & r@5   & r@10  & r@1\%  & r@1   & r@5   & r@10  & r@1\% \\\hline \hline 
 \multirow{4}{*}{CVUSA} 
% & CVM-NET  \cite{Hu_2018_CVPR}   & 16.07    &39.07    &51.58    &88.06      & 16.25 & 38.86 & 49.41 & 88.11 & 7.42    &22.28    &32.37   & 74.30 & 7.38  & 22.51 & 32.63 & 75.38    \\  
& CVFT \cite{shi2020optimal}     &  23.13    &45.01    &55.94    &86.93     & 23.38 & 44.42 & 55.20 & 86.64  & 7.97   &2354    &33.32    &74.40     & 8.10 & 24.25 & 34.47 & 75.15  \\ 
& SAFA \cite{shi2019spatial}  & 53.33  &  76.82   & 84.29 &   97.48    & 52.85  &  76.27  &  83.78   &97.50     & 27.43   & 53.85    &65.70  &  92.27     & 28.71 &   55.27 &   66.65  &  92.78    \\
& Shi~\etal\cite{shi2020looking}  &{91.96} & {97.50} & {98.54} &\textbf{99.67}      & {78.11} & {89.46} & {92.90} & {98.50} &75.11    &\textbf{89.72}    &\textbf{93.48}    &98.71     & {48.53} & {68.47} & {75.63} & {93.02}  \\ 
& Ours &\textbf{92.69} & \textbf{97.78} & \textbf{98.60} & {99.61}     & \textbf{78.94}    &\textbf{90.31}    &\textbf{93.42}    &\textbf{98.67} & \textbf{75.65}    &89.17   & 93.44    &\textbf{98.90}
     & \textbf{54.27}    &\textbf{72.78}    &\textbf{79.54}    &\textbf{94.73} \\\hline
\multirow{4}{*}{CVACT\_val}  
% & CVM-NET  \cite{Hu_2018_CVPR}  & & & &       & 13.09 & 33.85 & 45.69 & 81.80  & & & &      & 3.94  & 13.69 & 21.23 & 59.22   \\ 
& CVFT \cite{shi2020optimal}    & 26.19    & 46.09    &54.52   & 80.53        & 26.79 & 46.89 & 55.09 & 81.03  & 6.56    &18.10    &26.24    &62.78     & 7.13 & 18.47 & 26.83 & 63.87    \\ 
& SAFA \cite{shi2019spatial}  & 44.15   & 68.02    & 75.83    & 93.14    & 43.54    &67.65    &75.64    &93.05     &20.98    & 43.65   &54.58    &86.09     &21.15    &44.60    &55.66    &86.40   \\
& Shi~\etal\cite{shi2020looking}  & {82.49}  & {92.44}     & {93.99}     & {97.32}      & {72.91}  & {85.70}     & \textbf{88.88}     & {95.28} & \textbf{67.26}    &\textbf{83.84}    &87.57    &\textbf{95.36}     & {49.12}  & {67.83}     & {74.18}     & {89.93}  \\  
&Ours & \textbf{82.70}  & \textbf{92.50}     & \textbf{94.24}     & \textbf{97.65}       & \textbf{73.06} & \textbf{85.73} & {88.76}    &\textbf{95.44} & 67.23    &83.57    &\textbf{87.81}    &95.25     & \textbf{52.98}    &\textbf{71.18}    &\textbf{77.36}    &\textbf{91.61}
\\\bottomrule
\end{tabular}
\label{tab: compare_stoa_unknown_orien_limited_FoV}
\end{table*}

\smallskip
\noindent\textbf{Distance-based localization.}
In CVUSA and CVACT\_val, there is only one matching satellite image in the database for a query image in the test set. 
In~\cite{Liu_2019_CVPR}, we introduced CVACT\_test to evaluate the performance of different methods on real-world localization scenarios. 
This test set provides geotagged (GPS) satellite images that densely cover a city, and the localization performance is measured in terms of distance (meters). Specifically, a ground image is considered as successfully localized if one of the retrieved top $K$ satellite images is within 5 meters of the ground-truth location of the query ground image. 
That is to say, in this test set, there might be several matching satellite images in the database for a query ground image. 
Following the evaluation protocol in \cite{Liu_2019_CVPR}, we plot the percentage of correctly localized ground images (recall) at different values of $K$ in Figure \ref{fig: recall_CVACT_test}.

Compared to the work~\cite{shi2020looking}, SAFA~\cite{shi2019spatial} has eight additional Spatially-aware Position Embedding (SPE) modules to construct cross-view correspondences. 
Thus, the performance of SAFA is significantly better than that of Shi~\etal~\cite{shi2020looking} in this challenging test set. 
As a contribution of this paper, we establish more realistic geometric correspondences between satellite and ground-level images compared to our conference version~\cite{shi2020looking} which makes the extracted feature descriptors more informative. 
Hence, the localization performance on the CVACT\_test is further boosted. 
The top-1 recall rate of SAFA and the newly proposed algorithm on the CVACT\_test is $55.50\%$ and $60.46\%$, respectively. 
Note that having the SPE module in the framework will prohibit the application of our DSM which is designed to address orientation-unknown and limited FoV query images. Thus we do not retain it in the proposed framework in this article. 
In the following section, we will demonstrate the superiority of our DSM module on localizing query images with unknown-orientation and limited FoV.

\begin{table*}[!t]
\setlength{\abovecaptionskip}{0pt}
\setlength{\belowcaptionskip}{0pt}
\footnotesize
\setlength{\tabcolsep}{3pt}
\centering
\caption{\footnotesize
The overall performance of 3-DoF coarse camera localization. 
% Orientation prediction performance on correctly localized ground images.
}
\label{tab: overall}
\begin{tabular}{c |cc c|ccc|cc c|ccc }
\toprule
Dataset     & \multicolumn{6}{c|}{CVUSA}                                        & \multicolumn{6}{c}{CVACT\_val}  \\ \midrule
\multirow{2}{*}{FoV}         & \multicolumn{3}{c|}{$360^{\circ}$} & \multicolumn{3}{c|}{$180^{\circ}$}         & \multicolumn{3}{c|}{$360^{\circ}$} & \multicolumn{3}{c}{$180^{\circ}$}     \\
&Loc\_acc&Orien\_acc  & Overall &Loc\_acc&Orien\_acc  & Overall&Loc\_acc&Orien\_acc  & Overall &Loc\_acc&Orien\_acc  & Overall\\\midrule \midrule
Shi~\etal\cite{shi2020looking}  & 78.11         & 99.41          & 77.65          & 48.53         &  \textbf{98.54} & 47.72          &72.91          &\textbf{99.84}& 72.79         &49.12         &\textbf{99.10} & 48.68 \\
Ours                       &\textbf{78.94} & \textbf{99.45} & \textbf{78.51} & \textbf{54.27}& 96.87           & \textbf{52.57} &\textbf{73.06} &99.75         &\textbf{72.88} &\textbf{52.98}& 98.72        & \textbf{52.30}    \\
\bottomrule

\end{tabular}
\end{table*}

\subsubsection{Practical Localization: Localizing With Unknown Orientation and Limited FoV}

We compare the performance of our newly proposed method with 
our three previous works \cite{shi2020optimal, shi2019spatial, shi2020looking}, on the CVUSA and CVACT\_val datasets in a more realistic localization scenario, where the ground images do not have a known orientation and have a limited FoV. Recall that Liu and Li \cite{Liu_2019_CVPR} require the orientation information as an input, so we cannot compare our approach with such a method. 

\smallskip
\noindent\textbf{Location estimation.} 
In order to evaluate the impact of orientation misalignments and limited FoVs on localization performance, we randomly shift and crop the ground panoramas along the azimuth direction for the CVUSA and CVACT\_val datasets. In this manner, we mimic the procedure of localizing images with limited FoV and unknown orientation. 
For completeness, we also report the performance of the models trained without known orientations on orientation-aligned query images. 
Results are presented in Table \ref{tab: compare_stoa_unknown_orien_limited_FoV}. 

For CVFT~\cite{shi2020optimal} and SAFA~\cite{shi2019spatial}, 
the difference between the models from Sec.~\ref{exp:orien-align} and this section is that we apply the random orientation augmentation on training image pairs. 
By doing so, we expect that the networks can tolerate some orientation misalignments between satellite and ground image pairs.  
Those two methods do not have the ability to explicitly align the orientation of ground and satellite images.
Thus, their performance drops significantly when localizing ground images with unknown orientation, which can be seen by comparing Table \ref{tab: compare_stoa_CVUSA}, \ref{tab: compare_stoa_CVACT} and \ref{tab: compare_stoa_unknown_orien_limited_FoV}. 
Furthermore, since the data augmentation on orientation is applied during training, these networks need to spend more capacity on tolerating orientation misalignments. Their performance on localizing orientation-aligned query images drops. 

For the method~\cite{shi2020looking} and the newly proposed method, the models from Sec.~\ref{exp:orien-align} and this section are the same, where random orientation augmentation is applied during training.
In these two methods, an explicit parameter-free orientation alignment mechanism (the Dynamic Similarity Matching module) is applied, and thus they outperform CVFT and SAFA by a large margin on localizing orientation-unknown query images. 
The performance degradation of the two methods between orientation-aligned and orientation-unknown images is mainly due to the orientation ambiguity.
Compared to the work~\cite{shi2020looking} that we extend from, the newly proposed method establishes more sensible geometric correspondences between satellite and ground-level images by a projective transform, leading to a better performance. 

Finally, as FoVs increase, one can observe that all of the methods perform better. That is mainly because a larger FoV image provides richer scene contents, making a global descriptor more discriminative.

\begin{figure}
\setlength{\abovecaptionskip}{0pt}
\setlength{\belowcaptionskip}{0pt}
    \centering
    \includegraphics[width=0.2\linewidth]{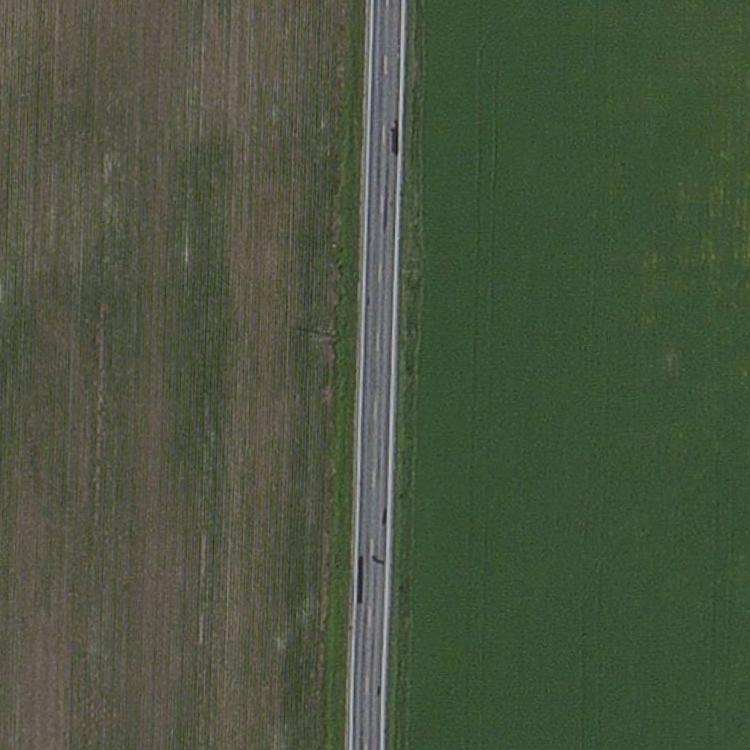}
    \includegraphics[width=0.2\linewidth]{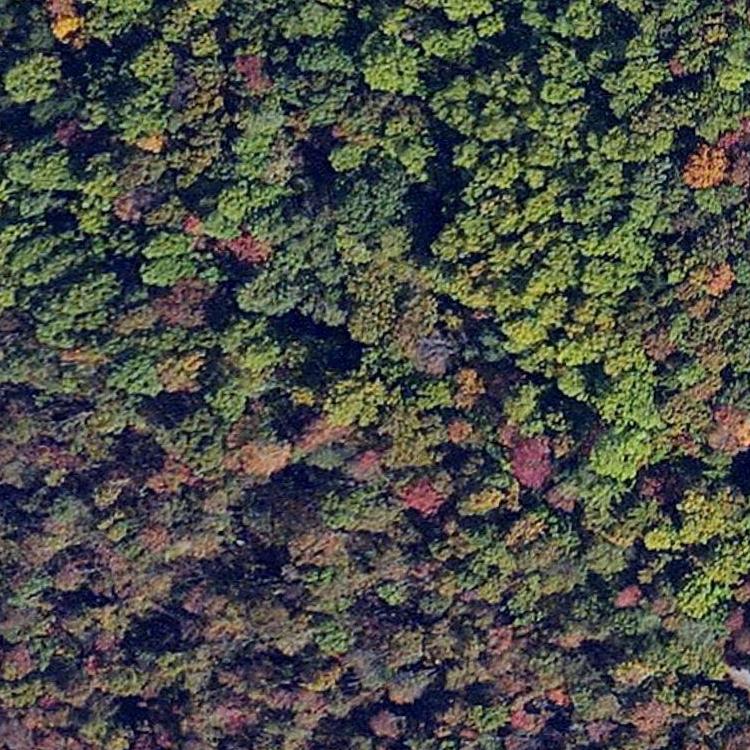}
    \includegraphics[width=0.2\linewidth]{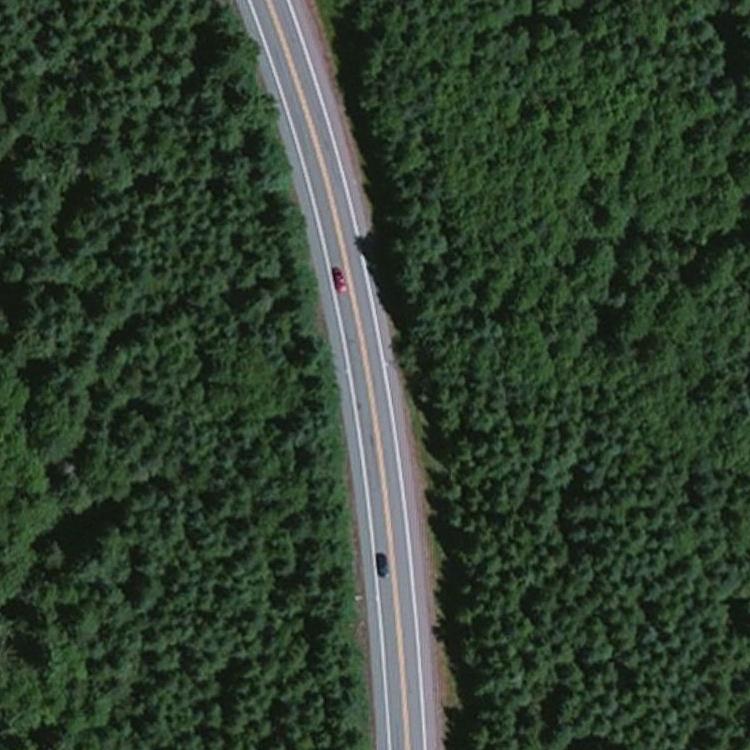}
    \includegraphics[width=0.2\linewidth]{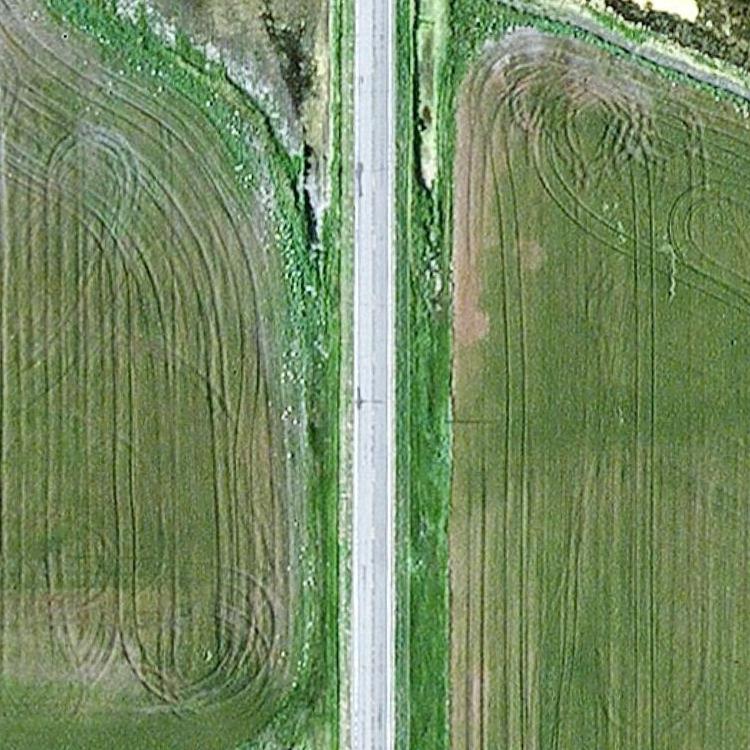}
    \caption{ \footnotesize Examples of symmetric scenes (satellite images). At these locations, it is hard to determine the orientation (azimuth angle) of a ground image.
    % if it only contains a small sector of the satellite image.
    }
    \label{fig:symmetric_scenes}
\end{figure}

\smallskip
\noindent\textbf{Orientation estimation.} 
As aforementioned, the estimated orientation of a query image is meaningful only if its location has been correctly determined. 
Thus, the experiment on the orientation estimation is conducted on ground images which are correctly localized in terms of top-1 retrieved candidates. 
For the fair comparison between different algorithms, we combine the location estimation accuracy and the orientation estimation accuracy together to present the overall performance score in Table~\ref{tab: overall}. 
As indicated by Table~\ref{tab: overall}, our proposed method achieves higher location estimation accuracy than our previous work~\cite{shi2020looking}, whereas the orientation estimation accuracy is slightly lower. This implies that scenes may look similar in multiple directions, even when the location is estimated correctly.
For instance, a person standing on a road may be able to localize their position but will find it difficult to determine the orientation if the views are similar along the road in both directions. Figure~\ref{fig:symmetric_scenes} contains examples of such symmetric scenes. 
According to the combined score, our newly proposed method achieves the best performance on 3-DoF camera localization. 
Figure~\ref{fig:visual_orien} visualizes the orientation estimation process.

\begin{figure*}[!h!t!]
\setlength{\abovecaptionskip}{0pt}
\setlength{\belowcaptionskip}{0pt}
    \centering
    \subfloat[]{
    \centering
    \includegraphics[width=0.11\linewidth]{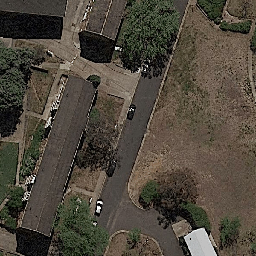}   
    \label{subfig:satellite}
    }
    \subfloat[]{
    \centering
    \includegraphics[width=0.22\linewidth, height=0.11\linewidth]{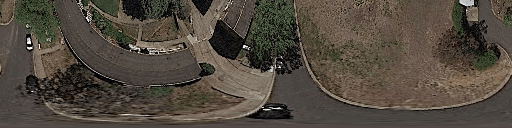}
    \label{subfig:polar-transformed}
    }
    \subfloat[]{
    \includegraphics[width=0.22\linewidth, height=0.11\linewidth]{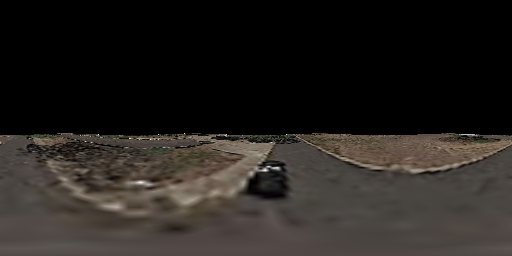}
    \label{subfig:projective-transformed}
    }
    \subfloat[]{
    \includegraphics[width=0.22\linewidth, height=0.11\linewidth]{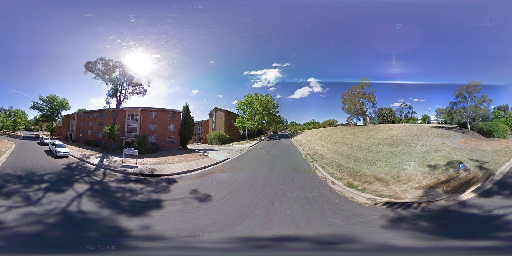}
    \label{subfig:grd_Panorama}
    }
    \subfloat[]{
    \includegraphics[width=0.22\linewidth, height=0.11\linewidth]{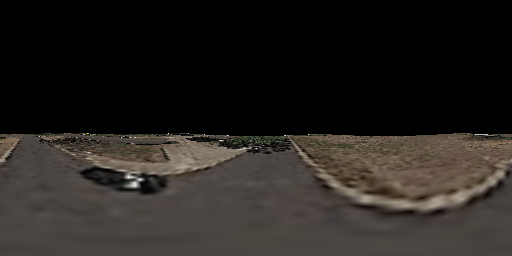}
    \label{subfig:projective_correct}
    }
    \caption{\footnotesize (a) Satellite image; (b) polar-transformed satellite image at the satellite image center; (c) projective-transformed satellite image at the satellite image center; (d) query ground-level panorama; (e) projective-transformed satellite image at the ground camera location. The ground camera location is determined by our fine-grained camera localization method. All of the images are orientation aligned. }
    \label{fig: abla_PT_HT}
\end{figure*}

\begin{table*}[!t]
\setlength{\abovecaptionskip}{0pt}
\setlength{\belowcaptionskip}{0pt}
\footnotesize
\setlength{\tabcolsep}{4pt}
\centering
\caption{\footnotesize
% Comparison of recall rates by our method with (\cmark) or without (\xmark) aligning query images to non-matching satellite images during inference. 
Comparison of recall rates when the query image is orientation-aligned to every element in the database (standard case) or just the matching image.
}
\begin{tabular}{c|c|cccc|cccc|cccc|cccc}
\toprule
\multirow{3}{*}{Orien} & \multirow{3}{*}{\begin{tabular}[c]{@{}c@{}}Align\\ Negative\end{tabular} }     & \multicolumn{8}{c|}{CVUSA}                                               & \multicolumn{8}{c}{CVACT\_val}  \\ %\cmidrule{3-18}
                            &                                                                        & \multicolumn{4}{c|}{FoV=$360^{\circ}$} & \multicolumn{4}{c|}{FoV=$180^{\circ}$}  & \multicolumn{4}{c|}{FoV=$360^{\circ}$} &\multicolumn{4}{c}{FoV=$180^{\circ}$} \\
                            &                                                                        & r@1   & r@5   & r@10  & r@1\%      & r@1   & r@5   & r@10  & r@1\%       & r@1   & r@5   & r@10  & r@1\%      & r@1   & r@5   & r@10  & r@1\% \\ \hline \hline
Aligned                     & --                                                                & 92.69 & 97.78 & 98.60 & 99.61      & 75.65 &89.17  & 93.44 & 98.90       & 82.70 & 92.50 & 94.24 & 97.65      & 67.23 & 83.57 & 87.81 &95.25 \\
Unknown                   & \cmark                                                                 & 78.94    &90.31   &93.42   &98.67      &54.27  &72.78  &79.54  &94.73        & 73.06 & 85.73 & 88.76 & 95.44      & 52.98 &71.18  &77.36  &91.61 \\
Unknown                   & \xmark                                                                 & 93.22 &{98.22}    &{98.94}    &{99.74} &79.44    & 92.64    & 95.78    & 99.46  & 84.30 & 93.93 &95.52 & 98.49& 73.06    & 88.61   & 92.19   & 97.90 \\\bottomrule
\end{tabular}
\label{tab: rotation_equivariance}
\end{table*}

{\color{black}
\smallskip
% \noindent\textbf{Rotational equivariance.} 
\noindent\textbf{Why does performance decrease when the orientation is unknown?} 
When the orientation of the camera is unknown, the search space becomes significantly larger. The alignment between the query image with non-matching satellite images increases the number of local maxima in the similarity objective function. 
Due to occlusions and the definition of the location alignment between the ground and satellite images in two transforms, the matching satellite image projected by the transforms might be not very similar to the query image. 
When the similarity score of an incorrect location-orientation pair is higher than the true pair, our method will estimate the wrong location and orientation for a query image.

\begin{figure}[!t]
\setlength{\abovecaptionskip}{0pt}
\setlength{\belowcaptionskip}{0pt}
    \centering
    \subfloat[FoV=$360^{\circ}$]{
    \begin{minipage}[b]{\linewidth}
    \begin{minipage}[b]{0.136\linewidth}
    \centering
    \includegraphics[width=\linewidth]{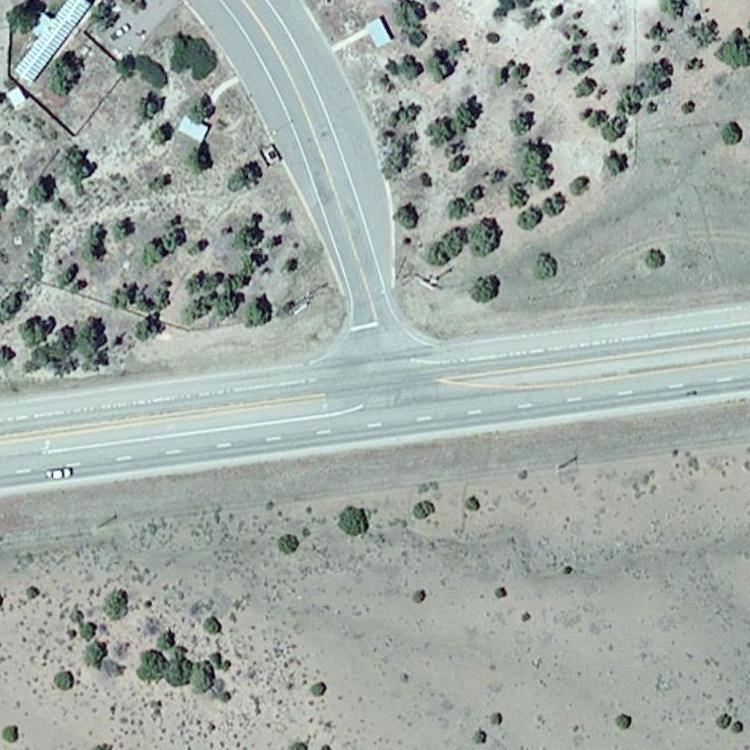}
    \end{minipage}
    \begin{minipage}[b]{0.42\linewidth}
    \centering
    \includegraphics[width=\linewidth,height=0.3333\linewidth]{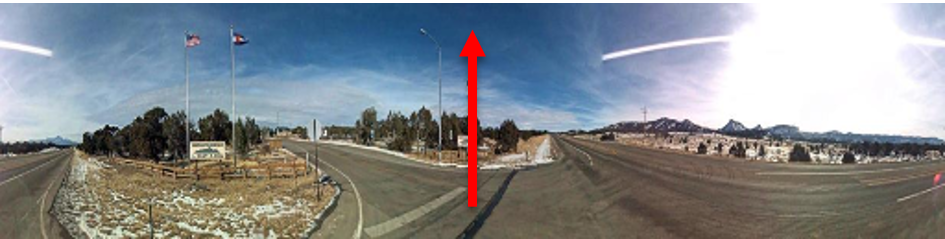}
    \end{minipage}
    \begin{minipage}[b]{0.42\linewidth}
    \centering
    \scalebox{1.0}[1.0]{\includegraphics[trim={2mm 0mm 0mm 3mm}, clip, width=\linewidth,height=0.3333\linewidth]{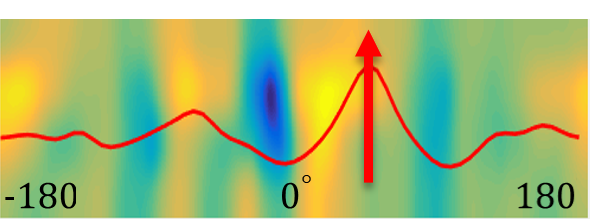}}
    \end{minipage}\\
     \begin{minipage}[b]{0.136\linewidth}
    \centering
    \includegraphics[width=\linewidth]{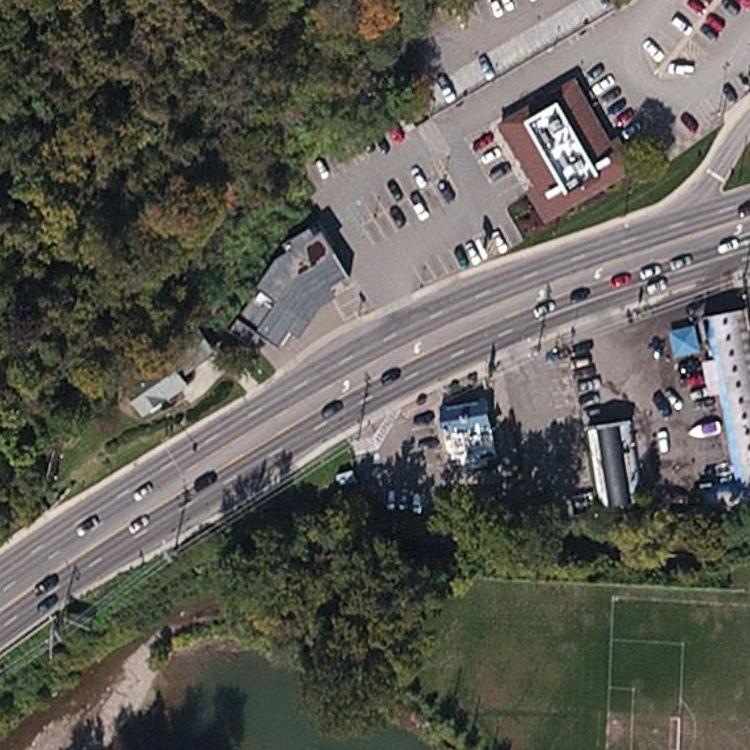}
    \end{minipage}
    \hspace{-1mm}
    \begin{minipage}[b]{0.42\linewidth}
    \centering
    \includegraphics[width=\linewidth,height=0.3333\linewidth]{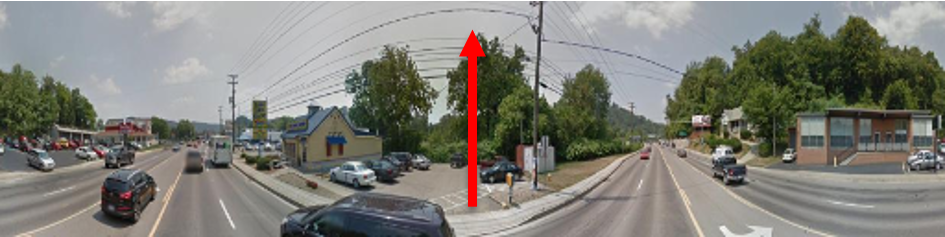}
    \end{minipage}
    \hspace{-1.5mm}
    \begin{minipage}[b]{0.42\linewidth}
    \centering
    \scalebox{1.0}[1.0]{\includegraphics[trim={2mm 0mm 0mm 3mm}, clip, width=\linewidth,height=0.3333\linewidth]{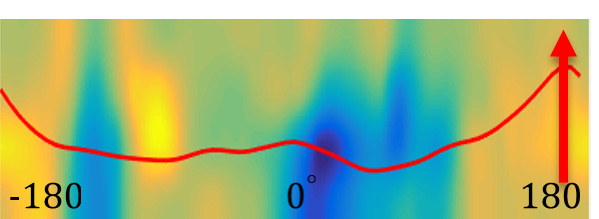}}
    \end{minipage}\\
    \vspace{-1em}
     \end{minipage}
     \label{fig: visual_orien_360}
     }\\
     \subfloat[FoV=$180^{\circ}$]{
     \vspace{-2mm}
    \begin{minipage}[b]{\linewidth}
    \begin{minipage}[b]{0.136\linewidth}
    \centering
    \includegraphics[width=\linewidth]{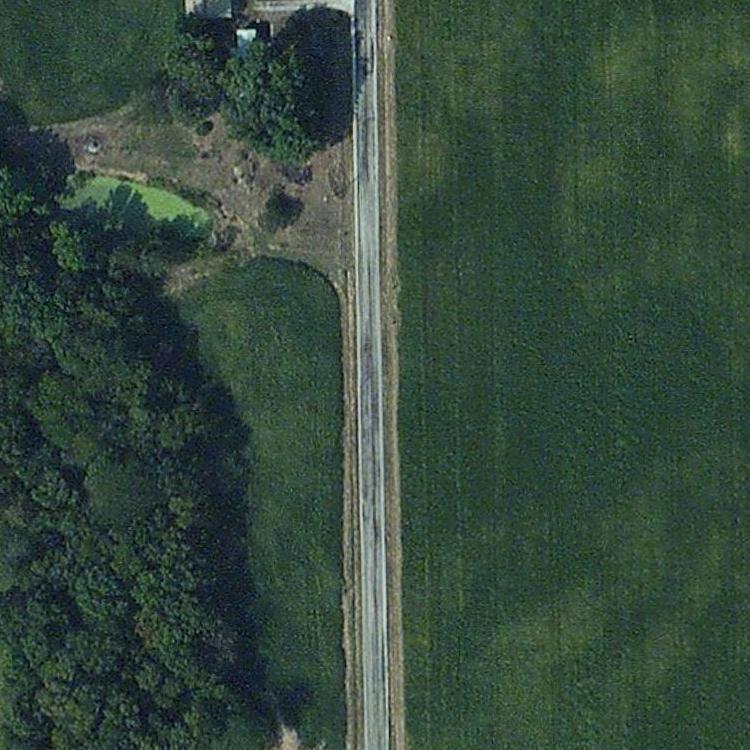}
    \end{minipage}
    \hspace{-1mm}
    \begin{minipage}[b]{0.42\linewidth}
    \centering
    \includegraphics[height=0.3333\linewidth]{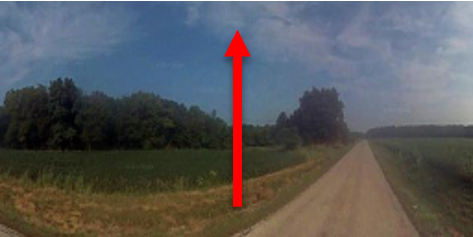}
    \end{minipage}
    \hspace{-1.5mm}
    \begin{minipage}[b]{0.42\linewidth}
    \centering
    \scalebox{1.0}[1.0]{\includegraphics[trim={2mm 0mm 0mm 3mm}, clip, width=\linewidth,height=0.3333\linewidth]{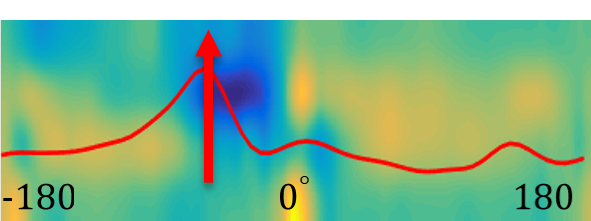}}
    \end{minipage}\\
     \begin{minipage}[b]{0.136\linewidth}
    \centering
    \includegraphics[width=\linewidth]{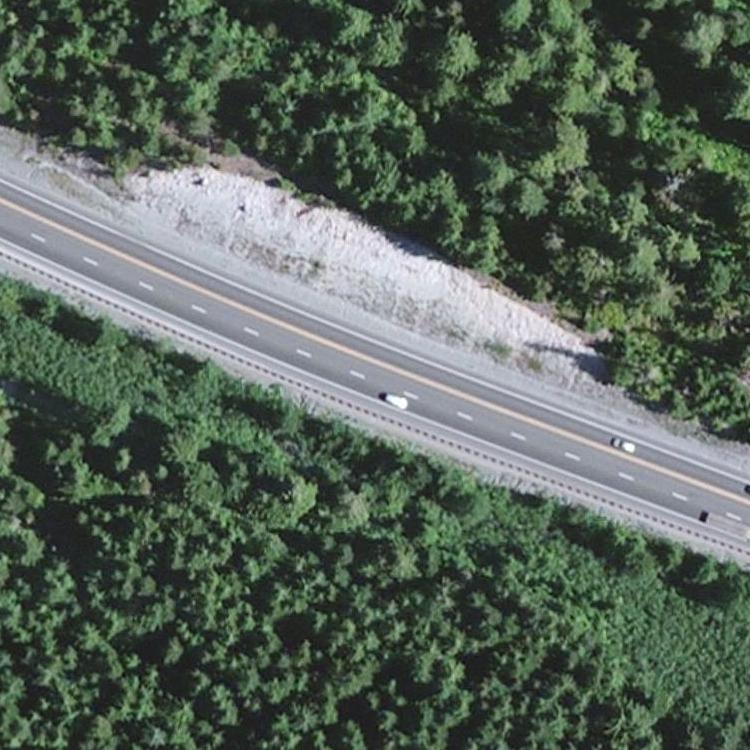}
    \end{minipage}
    \hspace{-1mm}
    \begin{minipage}[b]{0.42\linewidth}
    \centering
    \includegraphics[height=0.3333\linewidth]{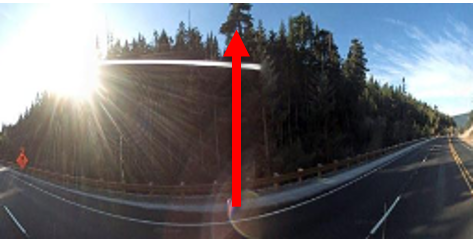}
    \end{minipage}
    \hspace{-1.5mm}
    \begin{minipage}[b]{0.42\linewidth}
    \centering
    \scalebox{1.0}[1.0]{\includegraphics[trim={2mm 0mm 0mm 3mm}, clip, width=\linewidth,height=0.3333\linewidth]{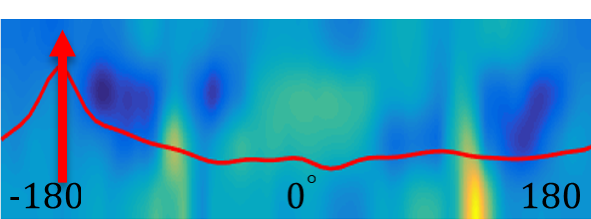}}
    \end{minipage}
    \end{minipage}
    \label{fig: visual_orien_70}}\\
    \caption{\footnotesize Visualization of estimated orientation for ground images with FoV = $360^{\circ}$ and $180^{\circ}$. In each of the subfigures, the satellite images are on the left and the ground images are in the middle. We visualize the transformed satellite features and the correlation results (red curves) in the right column. The positions of the correlation maxima in the curves corresponds to the orientation of the ground images.}
    \label{fig:visual_orien}
\end{figure}

\begin{table*}[!h!t!]
\setlength{\abovecaptionskip}{0pt}
\setlength{\belowcaptionskip}{0pt}
\footnotesize
\setlength{\tabcolsep}{3pt}
\centering
\caption{\footnotesize Comparison of recall rates for localizing ground images with unknown orientation and varying FoVs.}
\begin{tabular}{c|l|c c c c| c c c c|c c c c |c c c c}
\toprule
 \multirow{3}{*}{Dataset} & \multirow{3}{*}{Methods} & \multicolumn{8}{c|}{FoV=$360^{\circ}$}     & \multicolumn{8}{c}{FoV=$180^{\circ}$}     \\ \cline{3-18}
 & & \multicolumn{4}{c|}{Orientation Align}     & \multicolumn{4}{c|}{Orientation Unknown} &\multicolumn{4}{c|}{Orientation Align}     & \multicolumn{4}{c}{Orientation Unknown} \\
                      &                                       & r@1   & r@5   & r@10  & r@1\%  & r@1   & r@5   & r@10  & r@1\% & r@1   & r@5   & r@10  & r@1\%  & r@1   & r@5   & r@10  & r@1\% \\\hline \hline
 \multirow{3}{*}{CVUSA} 
& Ours w/o polar   & 84.78  &  93.99   & 95.95 &  99.13    & 65.33    &80.46    &85.61   & 0.9612     & 60.48    &79.78    &85.40    &96.49     & 37.27    &57.36    &65.93    &88.15   \\
& Ours w/o proj~\cite{shi2020looking}  &{91.96} & {97.50} & {98.54} &\textbf{99.67}      & {78.11} & {89.46} & {92.90} & {98.50} &75.11    &\textbf{89.72}    &\textbf{93.48}    &98.71     & {48.53} & {68.47} & {75.63} & {93.02}  \\ 
& Ours &\textbf{92.69} & \textbf{97.78} & \textbf{98.60} & {99.61}     & \textbf{78.94}    &\textbf{90.31}    &\textbf{93.42}    &\textbf{98.67} & \textbf{75.65}    &89.17   & 93.44    &\textbf{98.90}
     & \textbf{54.27}    &\textbf{72.78}    &\textbf{79.54}    &\textbf{94.73} \\\hline
\multirow{3}{*}{CVACT\_val}  
& Ours w/o polar  & 68.71    &84.38    &87.94    &94.70    & 50.50    &68.83    &74.75    &89.45     &43.88    &65.80    &73.05    &89.46     &27.95    &46.26    &54.10    &79.45   \\
& Ours w/o proj~\cite{shi2020looking}  & {82.49}  & {92.44}     & {93.99}     & {97.32}      & {72.91}  & {85.70}     & {88.88}     & {95.28} & \textbf{67.26}    &\textbf{83.84}    &87.57    &\textbf{95.36}     & {49.12}  & {67.83}     & {74.18}     & {89.93}  \\  
&Ours & \textbf{82.70}  & \textbf{92.50}     & \textbf{94.24}     & \textbf{97.65}       & \textbf{73.06} & \textbf{85.73} & \textbf{88.76}    &\textbf{95.44} & 67.23    &83.57    &\textbf{87.81}    &95.25     & \textbf{52.98}    &\textbf{71.18}    &\textbf{77.36}    &\textbf{91.61}
\\
\end{tabular}
\begin{tabular}{c|l|c c c c| c c c c|c c c c |c c c c}
\toprule
 \multirow{3}{*}{Dataset} & \multirow{3}{*}{Methods} & \multicolumn{8}{c|}{FoV=$90^{\circ}$}     & \multicolumn{8}{c}{FoV=$70^{\circ}$}     \\ \cline{3-18}
 & & \multicolumn{4}{c|}{Orientation Align}     & \multicolumn{4}{c|}{Orientation Unknown} &\multicolumn{4}{c|}{Orientation Align}     & \multicolumn{4}{c}{Orientation Unknown} \\
                      &                                       & r@1   & r@5   & r@10  & r@1\%  & r@1   & r@5   & r@10  & r@1\% & r@1   & r@5   & r@10  & r@1\%  & r@1   & r@5   & r@10  & r@1\% \\\hline \hline
 \multirow{3}{*}{CVUSA} 
& Ours w/o polar   & 11.65    &23.73    &31.09    &59.87     &   7.02    &17.95    &25.01    &59.43      & 6.27    &14.55    &19.57    &43.91   &   4.09    &11.81    &18.13    &50.98  \\
& Ours w/o proj~\cite{shi2020looking}  &\textbf{33.66}    &\textbf{51.70}    &\textbf{59.68}    &\textbf{82.35}     & \textbf{16.19} & \textbf{31.44} & \textbf{39.85} & \textbf{71.13}
&  \textbf{20.88}    &\textbf{36.99}    &\textbf{44.70}    &\textbf{70.95}  & \textbf{8.78} & \textbf{19.90} & \textbf{27.30} & \textbf{61.20}   \\ 
& Ours  & 20.28    &36.41    &44.47    &72.83     &    11.30    &26.00    &34.23    &69.03    &    10.50    &21.83    &28.79    &56.07    &   5.85    &15.71    &23.13    &59.34   \\\hline
\multirow{3}{*}{CVACT\_val}  
& Ours w/o polar  & 8.13    &17.54    &23.06    &48.50     &    4.19    &11.52    &16.34    &42.75
      &   3.33    &9.31    &13.09    &33.86    &  2.04    &5.83    &9.16    &31.89  \\
& Ours w/o proj~\cite{shi2020looking}  & \textbf{31.17}    &\textbf{51.44}    &\textbf{60.05}    &\textbf{82.73}     & \textbf{18.11}  & \textbf{33.34}     & \textbf{40.94}     & \textbf{68.65}
&   \textbf{18.45}    &\textbf{35.87}    &\textbf{44.39}    &\textbf{71.85}
    & \textbf{8.29}  & \textbf{20.72}     & \textbf{27.13}     & \textbf{57.08}  \\  
&Ours  &  21.42    &38.43    &47.06    &73.86      &   14.34    &27.90    &35.18    &64.86    &    9.68    &20.60    &27.68    &56.22     &   4.60    &12.73    &17.99    &46.86 
\\\bottomrule
\end{tabular}
\label{tab:ablation}
\end{table*}

\begin{table*}[!h!t!]
\setlength{\abovecaptionskip}{0pt}
\setlength{\belowcaptionskip}{0pt}
\footnotesize
\centering
\setlength{\tabcolsep}{4pt}
\caption{\footnotesize {Comparison of localization performance when ablating the transforms for orientation estimation. } }
\begin{tabular}{c|c|cccc|cccc|cccc|cccc}
\toprule
\multirow{2}{*}{}           & \multirow{2}{*}{\begin{tabular}[c]{@{}c@{}}Orien. \\ Est.\end{tabular}} & \multicolumn{4}{c|}{FoV=$360^\circ$}                               & \multicolumn{4}{c|}{FoV=$180^\circ$}                               & \multicolumn{4}{c|}{FoV=$90^\circ$}                                & \multicolumn{4}{c}{FoV=$70^\circ$}                               \\
                            &                              & r@1            & r@5            & r@10           & r@1\%          & r@1            & r@5            & r@10           & r@1\%          & r@1            & r@5            & r@10           & r@1\%          & r@1           & r@5            & r@10           & r@1\%          \\ \hline \hline
\multirow{3}{*}{CVUSA}      & Polar                        & 78.61          & \textbf{90.32} & 93.34          & 98.56          & 53.88          & 72.43          & 79.10          & 94.48          & 10.85          & 24.16          & 32.11          & 66.06          & 5.49          & 14.61          & 21.15          & 55.47          \\
                            & Proj                         & 78.57          & 89.90          & 93.00          & 98.37          & 52.00          & 69.82          & 76.28          & 92.28          & 10.74          & 23.69          & 31.37          & 65.49          & 5.61          & 15.31          & 22.20          & 56.89          \\
                            & Combined                        & \textbf{78.94} & 90.31          & \textbf{93.42} & \textbf{98.67} & \textbf{54.27} & \textbf{72.78} & \textbf{79.54} & \textbf{94.73} & \textbf{11.30} & \textbf{26.00} & \textbf{34.23} & \textbf{69.03} & \textbf{5.85} & \textbf{15.71} & \textbf{23.13} & \textbf{59.34} \\ \midrule
\multirow{3}{*}{CVACT\_val} & Polar                        & \textbf{73.11}          & 85.56          & 88.74          & 95.41          & 52.71          & 70.45          & 76.83          & 91.11          & 13.45          & 26.31          & 33.26          & 61.91          & 3.80          & 10.65          & 15.26          & 40.54          \\
                            & Proj                         & 72.73          & 85.02          & 88.12          & 94.65          & 50.54          & 67.83          & 73.67          & 87.97          & 12.13          & 23.12          & 29.19          & 55.31          & 4.12          & 11.19          & 15.62          & 41.90      \\
                            & Combined                        & {73.06} & \textbf{85.73} & \textbf{88.76} & \textbf{95.44} & \textbf{52.98} & \textbf{71.18} & \textbf{77.36} & \textbf{91.61} & \textbf{14.34} & \textbf{27.90} & \textbf{35.18} & \textbf{64.86} & \textbf{4.60} & \textbf{12.73} & \textbf{17.99} & \textbf{46.86} \\ \bottomrule
\end{tabular}
\label{tab:abla_corr}
\end{table*}

To demonstrate this, we conduct experiments for localizing orientation-unknown query images where the orientation alignment is only performed with respect to the matching satellite image, not the entire database as is standard.
Note that this is for illustration only, and is not a practical setting.
The results are shown in the third row of Table~\ref{tab: rotation_equivariance}. 
It can be seen that the performance is significantly better than that of aligning query images with non-matching ones by the DSM (the second row of Table~\ref{tab: rotation_equivariance}). 
It can be seen that the performance is significantly better than when aligning with respect to the entire database (the second row of Table~\ref{tab: rotation_equivariance}). 
In fact, the results are even better than when using orientation-aligned query images, since minor orientation misalignments between the query and matching satellite images can be rectified by our DSM module.

}

% \subsubsection{Ablation study}
\subsubsection{Discussion on the necessity of the polar transform and the projective transform}
In this section, we conduct an ablation study to demonstrate the effectiveness and necessity of the polar transform (polar) and the projective transform (proj). 
To this end, we remove the branch of polar-transformed satellite images or projective-transformed satellite images from our whole pipeline. We denote such settings as ``Ours w/o polar'' and ``Ours w/o proj'', respectively. 
Accordingly, the corresponding ground branch is also removed from the whole pipeline. 
In particular, ``Ours w/o proj'' becomes the method proposed in our conference version~\cite{shi2020looking}. 
Table~\ref{tab:ablation} reports the localization performance of the three baselines.

When the ground images have a $360^{\circ}$ or $180^{\circ}$ FoV, the newly proposed method with both polar and projective transform achieves the best performance. 
That is because the polar transform preserves all of the scene content information and the projective transform recovers the ground structure from a satellite image. 
As seen in Figure~\ref{fig: abla_PT_HT}, there is a building in the ground-level panorama (Figure~\ref{subfig:grd_Panorama}), and it is also visible from the satellite image (\ref{subfig:satellite}). The polar transformed satellite image (Figure~\ref{subfig:polar-transformed}) successfully preserves this scene information. 
However, the building is hard to be recognized from the projective-transformed satellite image (\ref{subfig:projective-transformed}). In contrast, the ground structure is better recovered by the projective transform. 
The two types of transformed images complement with each other to make the satellite image descriptor informative and discriminative. 
Moreover, we also present the projective-transformed satellite image at the ground camera location in Figure~\ref{subfig:projective_correct}. It aligns better with the ground-level panorama. The ground camera location is estimated by our fine-grained camera localization method. 

When the ground images have a more restricted FoV, \ie{ $90^{\circ}$ and $70^{\circ}$}, the results in Table~\ref{tab:ablation} show that the newly proposed method achieves inferior performance than ``Ours w/o proj''. 
The reason is that the projective-transformed satellite image only preserves a small portion information from the original satellite image. %  and it is more sensitive to the projection center. 
When the query image has a small FoV, \eg, only the building part in Figure~\ref{subfig:grd_Panorama}, it will not be matched to the projective-transformed satellite image. 
Thus its localization result would be incorrect. 
Not surprisingly, when the polar transform is removed from our pipeline, the performance of `Ours w/o polar'' decreases significantly. This demonstrate the necessity of the polar transform. 

To summarize, our newly proposed method is more suitable for localizing larger FoV images, and when the FoVs of query images are small, removing the projective transform branch and its corresponding ground image branch will be more recommended.
% and the method introduced in our conference version~\cite{shi2020looking} performs better when localizing smaller FoV query images. 
In real-world applications, it would be better to employ different branches in our proposed pipeline according to different deployment situations.  
% More illustrations will be introduced next. 

{\color{black}
\smallskip
\noindent \textbf{Orientation estimation w.r.t. the two transforms. }
We also study the influence on localization performance when only using the polar transform or the projective transform in orientation estimation, denoted as ``polar'' and ``proj'', respectively. 
The results are given in Table~\ref{tab:abla_corr}. 
The ablation labelled as ``combined'' is our whole method where both of the transforms are applied to orientation estimation. 
It can be seen that the polar transform contributes more to the final performance, but both help boost performance.

}

\begin{figure}[t!]
\setlength{\abovecaptionskip}{0pt}
    \setlength{\belowcaptionskip}{0pt}
    \centering
    \includegraphics[width=0.7\linewidth]{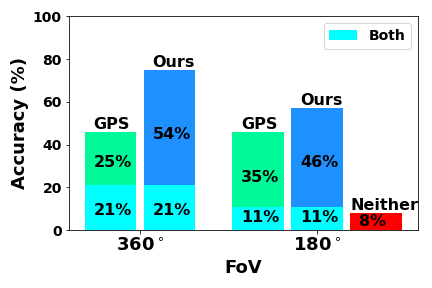}
    \caption{\footnotesize User study results for fine-grained camera localization (orientation aligned). 
    In this evaluation, users are asked to determine whether a location is correct or not. 
    The color cyan indicates the portion of data where both GPS and our estimated locations are correct. 
    The red bar ``Neither'' indicates the portion of data that cannot be localized when the FoV decreases from $360^\circ$ to $180^\circ$.
    }
    \label{acc:loc}
\end{figure}

\subsection{Fine-grained camera geo-localization} 
{\color{black}

Here, we conduct three groups of experiments on fine-grained camera geo-localization: 
(1) location estimation with known orientation; (2) orientation estimation with known location; and (3) joint location and orientation estimation.

\smallskip
\noindent \textbf{Location estimation.}
We first evaluate the location estimation accuracy of our method when the orientation of a query image is given. 
Since ground-truth labels of the precise camera locations are not available, we conduct a user study with five participants for quantitative evaluation. 
As shown in Figure~\ref{fig:accurate_localization}, given a query ground image (the last column) and its corresponding satellite image (the first column), it is very difficult for a human to localize which pixel in the satellite image corresponds to the query camera location. 
However, when the projected satellite image at a given location (the second or third column) is provided, it is much easier to determine whether this location is correct or not by comparing the projected satellite image with the query image. 
Therefore, the participants are provided with a query image, a projected satellite image at the GPS location, and a projected satellite image at the location estimated by our method. They are required to decide whether the camera location of each projected image is correct or not. 
Moreover, when the query image is different from both projected images, we mark this case as unknown. 
This case occurs when there is a severe occlusion in the vertical dimension. For example, tree canopies entirely cover the road underneath, which makes the ground-level localization extremely difficult.

We randomly select $150$ image pairs from the CVACT\_val set for this user study. 
The user study is first conducted on $360^\circ$ query images. 
We exclude the rare (\#) `unknown' samples and use the remaining images for the evaluation. 
From the results in Figure~\ref{acc:loc}, it can be seen that our method helps to correct the GPS drifting problem. 
The experiments on localizing $180^\circ$ FoV images are conducted on the same image set. 
We found that around $8\%$ of the query images cannot be localized when their FoV decreases from $360^\circ$ to $180^\circ$, marked as ``Neither'' in Fig.~\ref{acc:loc}. 
This is due to the increasing ambiguity when FoV decreases. 

\begin{table}[t!]
\setlength{\abovecaptionskip}{0pt}
    \setlength{\belowcaptionskip}{0pt}
    \setlength{\tabcolsep}{2pt}
    \centering
    \footnotesize
    \caption{\footnotesize The performance of our method for fine-grained camera localization. 
    }
    \begin{tabular}{cc|cc|cc}
    \toprule
\multirow{2}{*}{Loc} &\multirow{2}{*}{Orien} & \multicolumn{2}{c|}{FoV=$360^\circ$}  & \multicolumn{2}{c}{FoV=$180^\circ$}  \\
    &  & Loc\_acc (\%)   &  Orien\_acc(\%)  & Loc\_acc(\%)  &  Orien\_acc(\%) \\ \midrule
\xmark & \cmark  & 75.29 & -- & 56.46  & -- \\ 
\cmark & \xmark  & -- & 95.31 & --  & 70.83 \\ 
\xmark & \xmark  & 50.59 &  65.88 & 24.53 & 66.04   \\\bottomrule
    \end{tabular}
    \label{tab:acc_orien}
\end{table}

We present a failure case of our method in Figure~\ref{fig:failure}. As seen in the middle of the projected satellite image at the GPS location, the pixels from the tree canopies occlude the road pixels underneath. 
Those pixels are not similar to the pixels in the query image at corresponding positions. 
However, humans still can infer that the two images are at the same location, by leveraging the visible surrounding content and the geometric correspondences. 
In contrast, the SSIM metric only measures the overall pixel-wise similarity between the two images. 
It cannot handle pixels that are visually different but geometrically consistent. 
Thus, it yields a low similarity value between the image projected at the GPS location and the query image. 
Instead, it regards the image projected at a different coordinate (-20, 8) as the most similar one to the query image, since the road pixels in the two images align well and they account for a significant fraction of the image. 
As a result, our method fails to localize the correct location in this case.

\begin{figure*}[ht!]
    \setlength{\abovecaptionskip}{0pt}
    \setlength{\belowcaptionskip}{0pt}
    \centering
    \begin{minipage}{0.108\linewidth}
    \includegraphics[width=\linewidth]{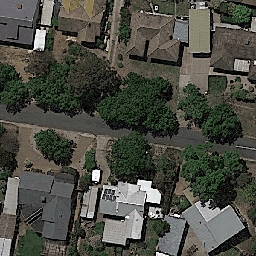}
    \centerline{\footnotesize Top-1 retrieved image}
    \end{minipage}
    \begin{minipage}{0.27\linewidth}
    \includegraphics[width=\linewidth, height=0.4\linewidth]{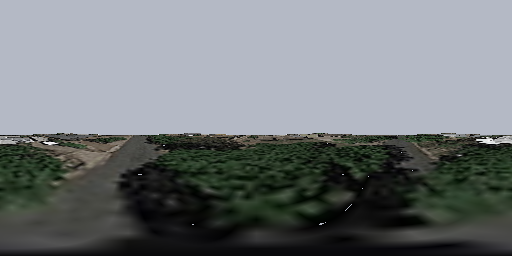}
    \centerline{\footnotesize Projected at the GPS location}
    \end{minipage}
    \begin{minipage}{0.27\linewidth}
    \includegraphics[width=\linewidth, height=0.4\linewidth]{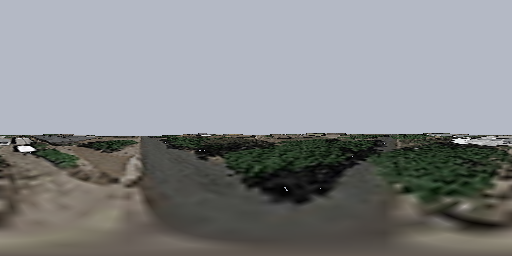}
    \centerline{\footnotesize Projected at (-20, 8)}
    \end{minipage}
    \begin{minipage}{0.27\linewidth}
    \includegraphics[width=\linewidth, height=0.4\linewidth]{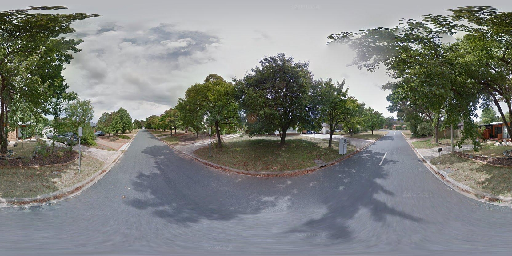}
    \centerline{\footnotesize Query ground image}
    \end{minipage}
    \caption{\footnotesize A failure case of the location estimation produced by our method in fine-grained $360^\circ$-FoV camera geo-localization (given orientation). 
    In this case, our method mistakenly regards the satellite image projected at the coordinate (-20, 8) rather than the GPS location as the most similar to the query image.}
    \label{fig:failure}
\end{figure*}

\begin{figure}[t!]
\setlength{\abovecaptionskip}{0pt}
    \setlength{\belowcaptionskip}{0pt}
    \centering
    \subfloat[Satellite image]{
    \begin{minipage}{0.31\linewidth}
    \centering
    \includegraphics[width=0.7\linewidth]{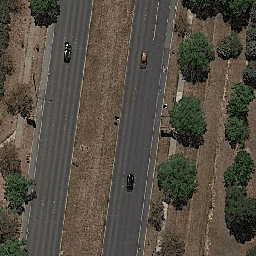}
    \end{minipage}
    \label{fig:180_sat}
    }
    \subfloat[Query image]{
    \begin{minipage}{0.31\linewidth}
    \centering
    \includegraphics[width=0.7\linewidth]{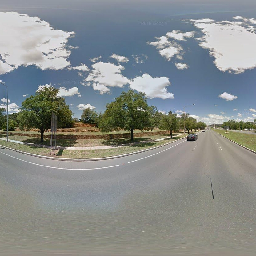}
    \end{minipage}
    \label{fig:gt_orien}
    }
    \subfloat[Scene contents at the predicted orientation]{
    \begin{minipage}{0.31\linewidth}
    \centering
    \includegraphics[width=0.7\linewidth]{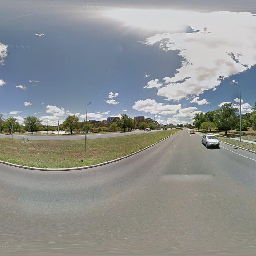}
    \end{minipage}
    \label{fig:pred_orien}
    }
   \caption{\footnotesize A failure case of the orientation estimation produced by our method in fine-grained $180^\circ$-FoV camera localization (given location). 
The scene structure at the predicted orientation is very similar to that of the query image.}
   \label{fig:failure_orien_180}
\end{figure}

\smallskip
\noindent \textbf{Orientation estimation. }
We also investigate the accuracy of the orientation estimation of our method when the camera location is known.
Here, the ground-truth location information is obtained from the previous stage of the user study, rather than the noisy GPS provided by the dataset. 
We generate query images by rotating the aligned images randomly along the azimuth direction and storing the rotation as the ground-truth orientation.
The results in the second row of Table~\ref{tab:acc_orien} indicate that the accuracy of orientation estimation is satisfactory when the query images have a $360^\circ$ FoV. However, the accuracy decreases significantly when the FoV of the query images decreases to $180^\circ$. 
Figure~\ref{fig:failure_orien_180} presents a failure case of the orientation estimation for $180^\circ$ FoV images. 
As shown in the figure, the scene contents at the estimated orientation are very similar to those at the ground-truth orientation (\ie, the query image), 
making the orientation estimation rather ambiguous. 

\smallskip
\noindent \textbf{Joint location and orientation estimation.} 
Here we evaluate the overall performance of our method on joint location and orientation estimation for fine-grained camera localization. 
The experiments are conducted on the images of which the ground-truth is provided by the user study participants. 
For location estimation, we consider localization within $\pm 5$ pixels ($\pm 1.4$ meters) as a success.
The results are presented in the third row of Table~\ref{tab:acc_orien}. 
The first row of Table~\ref{tab:acc_orien} shows the localization performance of our method when orientation is given. 
It can be seen that the joint estimation task is much more challenging, since the ambiguity of the problem becomes even severe.
}

{\color{black}
\section{Limitations.}
Our method assumes that the image plane of the query camera is perpendicular to the ground plane.
The tilt (elevation) and roll angles of the ground cameras in the current datasets are approximately zero. Thus, the sensitivity of our method to these angles is not investigated. 

For the fine-grained 3-DoF camera localization from satellite images,
there is considerable space for our method to improve.
The pixel-wise SSIM similarity used by our method is not robust to severe occlusions between the ground and satellite images. 
It could be replaced by a higher-level content similarity metric. 
Improving the runtime is also an important aspect. 
For example, a more sophisticated searching strategy instead of the exhaustive searching is required to find the camera location within a selected region. 
}
\section{Conclusions}
% In this paper we have proposed an effective algorithm for image-based geolocalization, which can hand

In this paper, we proposed an effective two-stage algorithm for ground-to-satellite image geo-localization, which can handle complex cases where neither location nor orientation are known.  
% Armed with this algorithm, and take a camera and a satellite map with you, you will never get lost or disoriented on hiking trails. 
In contrast to many existing methods, our algorithm provides accurate 3-DoF (location and orientation) camera localization results. 
Specifically, our method includes a coarse localization step which retrieves the most similar satellite image from the database given a query image, and a fine-grained localization step which computes the 
%distance
displacement between the query ground camera location and the retrieved satellite image center. 
The orientation alignment between satellite and ground images in both steps is evaluated. 
% Key contributions of this paper include a polar-transformation to bring different domains closer and a novel Dynamic Similarity Matching module (DSM) to regress on relative orientation. 
% Key components of our framework include a polar-transformation and a geometry-transform to bring different domains closer, and a novel Dynamic Similarity Matching module (DSM) to regress on relative orientation. 
In contrast to our previous works, we established more authentic geometric correspondences between satellite and ground images using a projective transform. 
By exploring the geometric correspondences, we successfully boost the coarse localization performance in terms of higher location recalls, and provide a novel method for accurate 3-DOF camera localization. 

\ifCLASSOPTIONcaptionsoff
  \newpage
\fi

\bibliographystyle{IEEEtran}
\bibliography{egbib}

\begin{IEEEbiography}[{\includegraphics[width=1in,height=1.25in,clip,keepaspectratio]{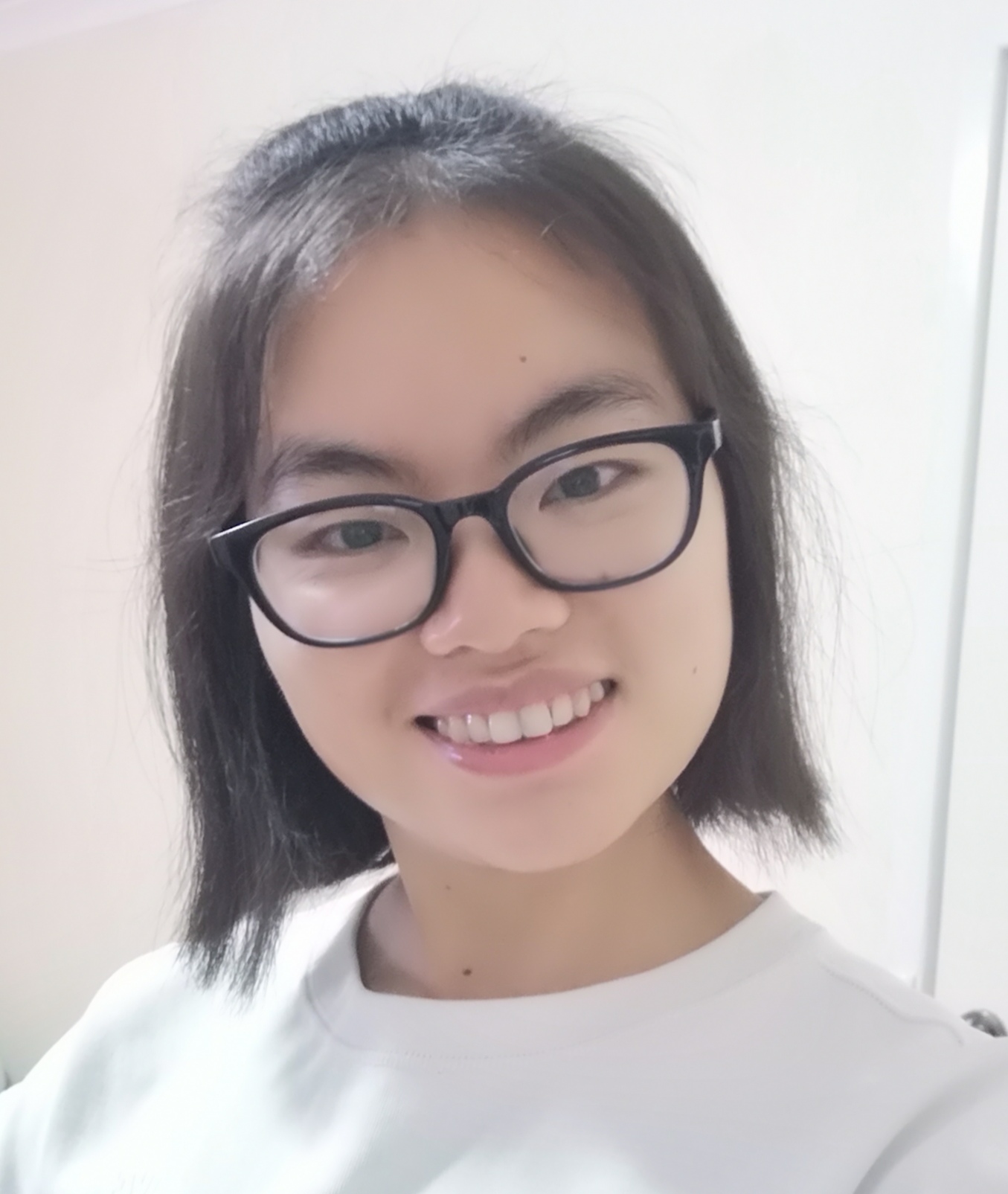}}]{Yujiao Shi} is a PhD student at the College of Engineering and Computer Science (CECS), Australian National University. She received her B.E. degree and M.S. degree in automation from Nanjing University of Posts and Telecommunications, Nanjing, China, in 2014 and 2017, respectively. Her research interests include satellite image based geo-localization, novel view synthesis and scene understanding.
\end{IEEEbiography}

\begin{IEEEbiography}[{\includegraphics[width=1in,height=1.25in,clip,keepaspectratio]{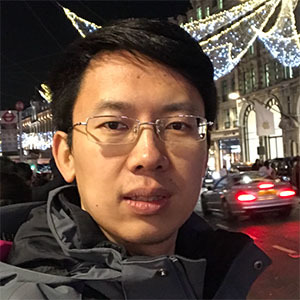}}]{Xin Yu}
received his B.S. degree in Electronic Engineering from University of Electronic Science and Technology of China, Chengdu, China, in 2009, and received his Ph.D. degree in the Department of Electronic Engineering, Tsinghua University, Beijing, China, in 2015. He also received a Ph.D. degree in the College of Engineering and Computer Science, Australian National University, Canberra, Australia, in 2019. He is currently a senior lecturer in University of Technology Sydney. His interests include computer vision and image processing.
\end{IEEEbiography}

\begin{IEEEbiography}[{\includegraphics[width=1in,height=1.25in,clip,keepaspectratio]{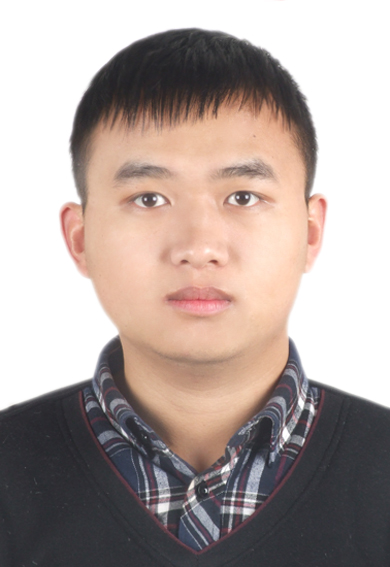}}]{Liu Liu}
is a Ph.D. student at the College of Engineering and Computer Science (CECS), Australian National University. He is also a member of the Australian Centre for Robotic Vision (ACRV) group.  He is from the honors college, and received the B.E. degree from the School of Automation, Northwestern Polytechnical University in 2012. His research
interests include robotic vision, multi-view geometry, and visual localization.
\end{IEEEbiography}

\begin{IEEEbiography}[{\includegraphics[width=1in,height=1.25in,clip,keepaspectratio]{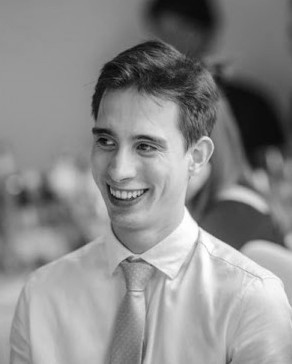}}]{Dylan Campbell}
is a Research Fellow with the Visual Geometry Group (VGG) at the University of Oxford. Prior to that he was a Research Fellow in the Research School of Computer Science at the Australian National University and the Australian Research Council Centre of Excellence in Robotic Vision. He received his PhD degree in Engineering from the Australian National University in 2018 while concurrently working as a research assistant in the Cyber-Physical Systems group at Data61–CSIRO. Prior to that, Dylan received a BE in Mechatronic Engineering from the University of New South Wales. His research interests cover a range of computer vision and machine learning topics, including visual geometry and differentiable optimization.
\end{IEEEbiography}

\begin{IEEEbiography}[{\includegraphics[width=1in,height=1.25in,clip,keepaspectratio]{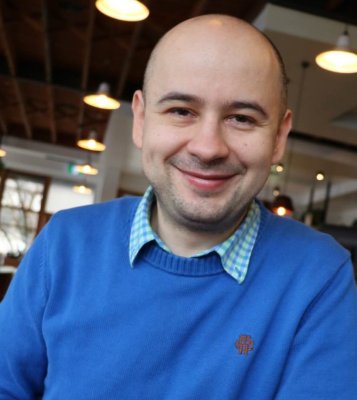}}]{Piotr Koniusz}
is a Senior Researcher in Machine Learning Research Group at Data61/CSIRO, formerly known as NICTA, and a Senior Honorary Lecturer at Australian National University (ANU). Previously, he worked as a postdoctoral researcher in the team LEAR, INRIA, France. He received his BSc degree in Telecommunications and Software Engineering in 2004 from the Warsaw University of Technology, Poland, and completed his PhD degree in Computer Vision in 2013 at CVSSP, University of Surrey, UK. 
His interests include visual categorization, spectral learning on graphs and tensors, kernel methods and deep learning.
\end{IEEEbiography}

\begin{IEEEbiography}[{\includegraphics[width=1in,height=1.25in,clip,keepaspectratio]{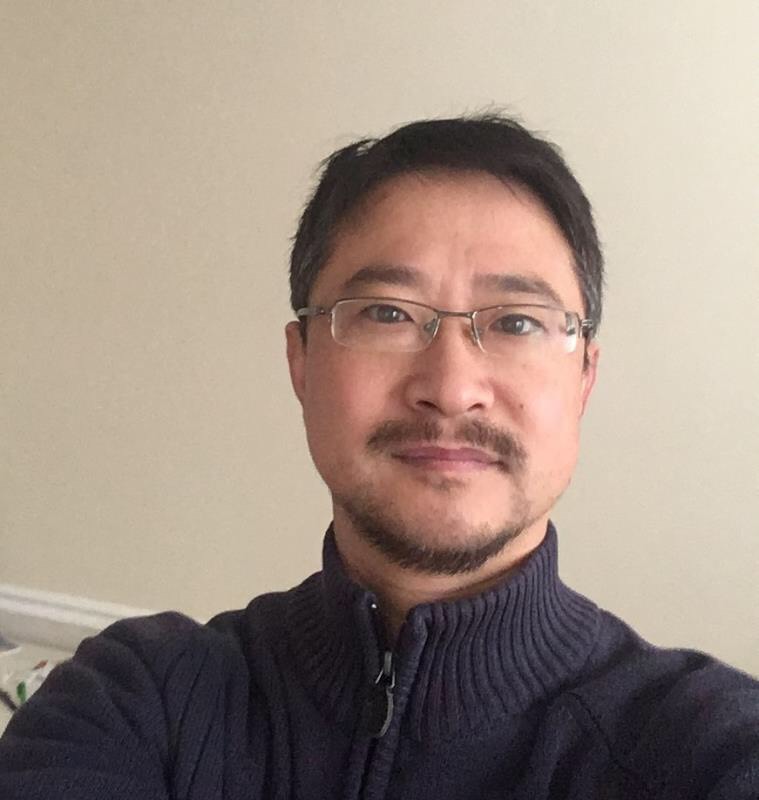}}]{Hongdong Li} is a Professor of ANU.  He is also a founding Chief Investigator for the Australia Centre of Excellence for Robotic Vision (ACRV). His research interests include 3D vision reconstruction, structure from motion, multi-view geometry, as well as applications of optimization methods in computer vision.  Prior to 2010, he was with NICTA working on the “Australia Bionic Eyes” project.  He is an Associate Editor for IEEE T-PAMI, Guest editor for IJCV, and Area Chair in recent year ICCV, ECCV and CVPR conferences.  He was a Program Chair for ACRA 2015 – Australia Conference on Robotics and Automation, and a Program Co-Chair for ACCV 2018 – Asian Conference on Computer Vision.  He won a number of paper awards in computer vision and pattern recognition, and was the receipt for the CVPR 2012 Best Paper Award, the ICCV Marr Prize Honorable Mention in 2017, and a shortlist of the CVPR 2020 best paper award.
\end{IEEEbiography}

%\vfill


\end{document}